\documentclass[11pt]{article}

\usepackage{blindtext}

\usepackage{geometry}
\usepackage{amsthm}
\usepackage{dsfont}
\usepackage{booktabs}
\usepackage{amssymb}
\usepackage{mathtools}
\usepackage{float}
\usepackage{upgreek}
\allowdisplaybreaks
\usepackage{bm}
\usepackage{enumitem}
\usepackage{multirow}
\usepackage{xcolor}
	
\usepackage{color, colortbl}
\usepackage{makecell}
\usepackage{graphicx}
\usepackage{natbib}
\usepackage{bookmark}

\usepackage{stackengine}
\usepackage[linesnumbered,ruled,vlined]{algorithm2e}

\newcommand{\q}{q}

\newcommand{\p}{p}
\newcommand{\D}{\mathbb{D}}

\newcommand{\e}{\mathbb{E}}
\newcommand{\V}{\mathbb{V}}

\newcommand{\ind}{\perp \!\!\! \perp }

\newcommand{\doo}{\textnormal{do}}
\allowdisplaybreaks

\newtheoremstyle{exampstyle}
  {3pt} {0} {} {} {\bfseries} {.} {.5em} {} \theoremstyle{exampstyle} \newtheorem{assumption}{Assumption}
  \newtheorem{example}{Example}
 \newtheorem{remark}{Remark}
\newtheorem{lemma}{Lemma}

\definecolor{Gray}{gray}{0.9}

\usepackage{lastpage}

\title{\textbf{Federated Causal Inference from\\Observational Data}}

\author{Thanh Vinh Vo \qquad Young Lee  \qquad Tze-Yun Leong \\[0.3cm]
	National University of Singapore
}

\date{}

\begin{document}

\maketitle

\begin{abstract}
\noindent Decentralized data sources are prevalent in real-world applications, posing a formidable challenge for causal inference. These sources cannot be consolidated into a single entity owing to privacy constraints. The presence of dissimilar data distributions and missing values within them can potentially introduce bias to the causal estimands. In this article, we propose a framework to estimate causal effects from decentralized data sources. The proposed framework avoid exchanging raw data among the sources, thus contributing towards privacy-preserving causal learning. Three instances of the proposed framework are introduced to estimate causal effects across a wide range of diverse scenarios within a federated setting. (1) FedCI: a Bayesian framework based on Gaussian processes for estimating causal effects from federated observational data sources. It estimates the posterior distributions of the causal effects to compute the higher-order statistics that capture the uncertainty. 
(2) CausalRFF: an adaptive transfer algorithm that learns the similarities among the data sources by utilizing Random Fourier Features to disentangle the loss function into multiple components, each of which is associated with a data source. The data sources may have different distributions; the causal effects are independently and systematically incorporated. It estimates the similarities among the sources through transfer coefficients, and hence requiring no prior information about the similarity measures. 
(3) CausalFI: a new approach for federated causal inference from incomplete data, enabling the estimation of causal effects from multiple decentralized and incomplete data sources. It accounts for the missing data under the missing at random assumption, while also estimating higher-order statistics of the causal estimands. CausalFI recovers the conditional distribution of missing confounders given the observed confounders from the decentralized data sources to identify causal effects. The proposed federated framework and its instances are an important step towards a privacy-preserving causal learning model.

\end{abstract}

\section{Introduction}

Causal inferences are important in real-life applications: \emph{What is the impact of a new drug on patient survival? How do vaccination campaigns help reduce  the incidence and spread of infectious diseases? What are the impacts of social and public policies across various domains such as poverty alleviation, education reform, and labor market interventions?}

Causal inference has been applied in a wide range of domains, including economics \citep{tiffin2019machine,finkelstein2020}, 
medicine \citep{henderson2016bayesian,powers2018some}, 
and social welfare \citep{Gutman:2017}. The large amount of experimental and/or observation data needed to accurately estimate the causal effects often resides across different sites. In most cases, the data sources cannot be combined to support centralized processing due to some inherent organizational or policy constraints. For example, in many countries, medical or health records of cancer patients are kept strictly confidential at local hospitals; direct exchange or sharing of the records among hospitals, especially for research purposes, are not allowed \citep{gostin2009beyond}. 
The main research question is: How can a causal model be learned to estimate causal effects across multiple data sources, while simultaneously minimizing the potential risks associated with compromising data records' confidentiality and sensitivity?

Current causal inference approaches \citep[e.g.,][]{shalit2017estimating,yao2018representation} require the shared data to be put in one place for processing.  
Current federated learning algorithms \citep[e.g.,][]{sattler2020,wang2020} allow collaborative learning of joint models based on non-independent and identically distributed \emph{(non iid)} data; they cannot, however, directly support causal inference due to several problems. First, the different data sources might have \emph{disimilar distributions} that would lead to biased causal effect estimation.
For example, the demographic profile and average age for cancer patients from two different hospitals may be drastically different. If the two data sets are combined to support causal inference, one distribution may dominate over the other, leading to biased causal effect estimation.  
Second, it is important to know whether the causal estimands are reliable. Thus, estimating a confidence interval of the relevant causal effect together with its point estimates would give helpful insights into the uncertainty of the causal estimand. For example, a narrow confidence interval for individual treatment effect of smoking on lung cancer, where zero falls outside the confidence interval, means that the patient is at a higher risk of getting cancer. 
Third, different data sources might contain missing values and the missing data attributed to a multitude of reasons such as erroneous data entry or processing, data loss, data deletion, etc, and it is commonly appeared in real-life application  \citep{allison2001missing,baraldi2010introduction,cismondi2013missing}. Dealing with missing data is crucial for accurate estimation of causal effects as it might introduce biases and result in misleading conclusions.

In this article, we present a comprehensive framework along with essential assumptions for federated estimation of causal effects. Subsequently, we introduce three distinct instances of this federated framework designed to estimate causal effects in various scenarios:
\begin{enumerate}
    \item[($i$)] addressing dissimilar distributions among data sources,
    \item[($ii$)]  handling missing values within the dataset, and
    \item[($iii$)] estimating the distribution of causal effects.
\end{enumerate}
These instances encapsulate diverse real-world challenges encountered in causal inference, offering tailored solutions within the federated setting. Our contributions are summarised as follows:
\begin{itemize}
    \item We introduce a federated framework and necessary assumptions for estimating causal effects from multiple data sources without sharing raw data, thus enabling privacy-preserving causal inference. The framework minimizes information transmitted among the sources, and fuses federated learning and causal inference to incorporate multiple data sources while maintaining the sources at their local sites. 
    \item We introduce three instances of the federated method that estimate causal effects in difference scenarios:
    \begin{itemize}
        \item FedCI\footnote{Published in \citep{vo2022bayesian}. Source code: \texttt{https://github.com/vothanhvinh/FedCI}.}: We introduce FedCI (Bayesian \underline{\textbf{Fed}}erated \underline{\textbf{C}}ausal \underline{\textbf{I}}nference), a Bayesian framework that can learn the causal effects of interest without combining data sources to a central site, and, at the same time, learn higher-order statistics of the causal effects to understand their uncertainty. FedCI generalizes the Bayesian imputation approach \citep[][]{imbens2015causal} to a more generic model based on Gaussian processes (GPs); the resulting model is decomposed into multiple components, each of which handles a distinct data source.  \item CausalRFF\footnote{Published in \citep{vo2022adaptive}. Source code: \texttt{https://github.com/vothanhvinh/CausalRFF}}: To address the dissimilar distributions among multiple data sources, we propose CausalRFF (\underline{\textbf{R}}andom Fourier \underline{\textbf{F}}eatures for \underline{\textbf{F}}ederated \underline{\textbf{Causal}} Inference)
a new approach to federated causal inference from multiple, decentralized, and disimilarly distributed data sources. CausalRFF is 
    based on the structural causal model (SCM) \citep{pearl2009causal}, it leverages the Random Fourier Features \citep{rahimi2007random} for federated estimation of causal effects. The Random Fourier Features allow the objective function to be divided into multiple components to support federated training of the model. We provide the minimax lower bounds to explicate the limits of estimation and optimization procedures in our federated causal inference framework.
    \item CausalFI\footnote{Source code: \texttt{https://github.com/vothanhvinh/CausalFI}}: Although FedCI and CausalRFF can estimate causal effects from multiple data sources in a federated setting, these methods also cannot handle data sources that contain missing values. On the other hand, methods that estimate causal effects in the presence of missing data \citep[e.g.,][]{crowe2010comparison,mitra2011estimating,seaman2014inverse,yang2019causal,hillis2021causal} are not designed within the context of a federated setting, where data needs to be on the same machine for performing inference. 
To address this problem, 
we propose a new method called CausalFI (\textbf{\underline{Causal}} \textbf{\underline{F}}ederated inference on \textbf{\underline{I}}ncomplete data), which recovers the distribution of missing confounders conditioned on the observed ones. This enables estimating causal effects from multiple data sources with missing values. We conduct federated causal inference with CausalFI using data from multiple sources with missing data. 
\end{itemize}
\item The proposed framework and its instances decompose the objective function to multiple components, each associated with a data source, which enable \emph{pure} federated learning of causal effects. We discuss the possibility
of combining differential privacy with the proposed framework at the end of the article.
\end{itemize}

\section{Related Work}

\noindent\textbf{On causal inference:} The authors \citet{hill2011bayesian,Alaa:2017,Alaa:2018,shalit2017estimating,Yoon:2018ganite,yao2018representation,kunzel2019metalearners,nie2020quasi} proposed learning causal effects directly from local data sources; these methods adopt the standard \emph{ignorability assumption} \citep{rosenbaum1983central}.   \citet{louizos2017causal,madras2019fairness} adapted the structural causal model (SCM) of \cite{pearl1995causal} to estimate the causal effects with the existence of latent confounding variables. 

Our works are closely related to and extends the notion of \emph{transportability}, where
\citet{pearl2011transportability,lee2013causal,bareinboim2016causal,lee2020generalized} and related work
formulated and provided theoretical analysis of intervention tools on one population to compute causal effects on another population. \cite{lee2020generalized} generalized transportability to support identification of causal effects in the target domain from the observational and interventional distributions on subsets of observable variables,
forming a foundation for drawing conclusions for observational and experimental data \citep{tsamardinos2012towards,bareinboim2016causal}. Causal inference from multiple, decentralized, dissimilarly distributed sources that cannot be combined or processed in a central site is not addressed. 
Recently,  \citet{aglietti2020multi}, conducted randomized experiments on the source to collect data and then estimated a joint model of the interventional data from source population and the observational data from target population.  Our work is different in that we do not work with randomized data; we estimate causal effects through transfers using only \emph{observational data}. This corresponds to an important setting in real-life, where only retrospective observational data are available, e.g., Covid-19 related case and intervention records, bank and financial transaction records.

\noindent\textbf{On federated learning:} Federated learning enables collaboratively learning a shared prediction model while keeping all the training data decentralized at source \citep{mcmahan2017communication}. Some federated learning approaches combine federated stochastic gradient descent \citep{shokri2015privacy} and federated averaging \citep{mcmahan2017communication} to address regression problems \citet{alvarez2019non,zhe2019scalable,de2020mogptk,joukov2020fast} and \citet{hard2018federated,zhao2018federated,sattler2019robust,mohri2019agnostic}. Recent federated learning algorithms allow collaborative learning of joint deep neural network models based on non-iid data \citep{sattler2020,wang2020}.
All these algorithms, however, do not directly support causal inference as the different data sources might have \emph{dissimilar distributions} that would lead to biased causal effect estimation. 
Little work has been focused on federated estimation of causal effects. \citet{xiong2021federated} estimated average treatment effect (ATE) and average treatment effect on the treated (ATT) and assumed that the \emph{confounders are observed}. Our work, on the other hand, estimate conditional average treatment effect (CATE) (which is also known as individual treatment effect, ITE) and average treatment effect (ATE). We consider the existence of \emph{latent confounders} and \emph{dissimilar data distributions} in CausalRFF, uncertainty estimation in FedCI, and missing data in CausalFI.

\noindent\textbf{On causal inference with missing data:} The commonly recognized types of missing data mechanisms \citep{rubin1976inference} include: missing completely at random (MCAR), missing at random (MAR), and missing not at Random (MNAR). \citet{qu2009propensity,crowe2010comparison,mitra2011estimating,seaman2014inverse,kallus2018causal,yang2019causal,mayer2020missdeepcausal,mayer2020doubly,hillis2021causal} are typical works that consider estimating causal effects in the present of missing data. 
\citet{yang2019causal} assumed MNAR and outcome-independent of the missing indicators to derive a non-parametric method for estimating causal effects. \citet{hillis2021causal} introduce a missing data mechanism that integrates with an iterative multivariate matching method, leveraging random forest to construct a distance matrix, and then facilitating subsequent optimal matching. \citet{qu2009propensity,crowe2010comparison,mitra2011estimating,seaman2014inverse} proposed multiple imputation-based methods based on MCAR or MAR assumptions for imputing the missing values, consequently estimating propensity score for estimation of ATE. \citet{mayer2020doubly} proposed a doubly robust estimator to estimate ATE and ATT with missing attributes. These methods, however, required data to be centralized in a central client, which might breaches the confidential and private information. Our approach in CausalFI is distinct in that it does not rely on an outcome-independent assumption and instead focuses on estimating causal effects without the need to aggregate raw data.

\section{A Framework for Federated Estimation of Causal Effects}

In this section, we introduce a framework for learning causal models and aggregating causal affects from multiple data sources in federated settings.

\subsection{Problem Formulation and the Causal Quantities of Interest}
\label{sec:prob-form}

In this work, we aim to estimate the causal effects using multiple data sources. 
Suppose we have $m$ sources of data, each  denoted by $\mathsf{D}^\mathsf{s} = \{( w_i^\mathsf{s}, y_i^\mathsf{s}, x_i^\mathsf{s})\}_{i=1}^{n_\mathsf{s}}$, where $\mathsf{s}\in\bm{\mathcal{S}} \vcentcolon= \{1,2,\!...,m\}$, and the quantities $w_i^\mathsf{s}$, $y_i^\mathsf{s}$ and  $x_i^\mathsf{s}$ are the treatment assignments, observed outcome associated with the treatment, and covariates of individual $i$ in source $\mathsf{s}$, respectively. For notational convenience, we denote $y_{i,\textrm{obs}}^\mathsf{s} \equiv y_i^\mathsf{s}$ and use these representations interchangeably. We focus on binary treatment $w_i^\mathsf{s} \in \{0,1\}$, thus $y_{i,\textrm{obs}}^\mathsf{s}$ can be either of the potential outcomes $y_i^\mathsf{s}(0)$ or $y_i^\mathsf{s}(1)$, i.e., for each individual $i$, we can only observe either $y_i^\mathsf{s}(0)$ or  $y_i^\mathsf{s}(1)$, but not both of them.

The objective is to develop a global causal inference model that satisfies \emph{both} of the following two conditions: \textbf{(i)} the causal inference model can be trained in a private setting where the data of each source are not shared to an outsider, and \textbf{(ii)} the causal inference model can incorporate data from multiple sources to improve causal effects estimation in each specific source.

In the following, we give formal definitions of the causal effects of interest. Let $Y$, $W$ and $X$ be random variables associated with the outcomes, treatment, and covariates, respectively.

\noindent\textbf{Average Treatment Effect (ATE):} ATE represents the average difference in outcomes between a group that receives the treatment and a group that does not. For example, in a medical study, ATE could represent the average improvement in health for patients who received a new medication compared to those who did not. ATE can be defined in the notion of either potential outcomes framework or structural causal model as follows:
\begin{align}
    \texttt{ate} &\vcentcolon= E[Y(1) - Y(0)] \label{eq:ate-rubin}\\
    &\vcentcolon= E[Y | \doo(W=1)] - E[Y | \doo(W=0)],\label{eq:ate-pearl}
\end{align}
where $Y(1)$ and $Y(0)$ are two random variables of the outcome associated with treatment $W=1$ and $W=0$. The operator $\doo(W=w)$ denotes an intervention on the treatment that sets $W=w$.

\noindent\textbf{Conditional Average Treatment Effect (CATE):} CATE is the average treatment affect conditioned on some fixed variables. For example, CATE could measure the average treatment effect of a new medication designed to address the medical condition given an age group of patients. Formally, given a fixed value $X=x$, CATE is defined as follows:
\begin{align}
    \texttt{cate}(x) &\vcentcolon= E[Y(1) - Y(0) | X=x] \label{eq:cate-rubin}\\
    &\vcentcolon= E[Y |\doo(W=1), X=x] -  E[Y |\doo(W=0), X=x].\label{eq:cate-pearl}
\end{align}

\noindent\textbf{Individual Treatment Effect (ITE):} Given an individual $i$, the ITE is defined to be the difference of its potential outcomes:
\begin{align}
    \texttt{ite}_i \vcentcolon= y_i(1) - y_i(0).\label{eq:ite}
\end{align}
In general, it is impossible to compute ITE since we cannot observe both of the potential outcomes of $i$. Hence, we would need to impute the unobserved/missing potential outcome.

It is clearly that $\texttt{cate}(x_i) \neq \texttt{ite}_i$; nevetherless, the literature often conflates ITE and CATE, and the CATE evaluated at $x_i$ is certainly a good estimation for the ITE of individual $i$ and vice versa.

\subsection{Assumptions}
\label{sec:assumptions}
Let $Z$ be random variable of the confounder. 
The following assumptions are made to enable federated causal estimations: \begin{assumption}
[Strong Ignorability] \label{assumption:ignorability}  ($i$) $ Y(1), Y(0) \ind W|Z$, i.e., the potential outcomes are independent of the treatment assignment conditional on $Z$ (unconfoundedness), and ($ii$) every individual has some positive probability to be assigned to every treatment (positivity), i.e., $0 < \p(W=1|Z) < 1$. \citep{rosenbaum1983central}
\end{assumption}
\begin{assumption} [Stable Unit Treatment Value Assumption or SUTVA]
\label{assumption:sutva}
($i$) The potential outcomes for any individual do not vary with the treatments assigned to other individuals, and ($ii$) there are no different forms or versions of each treatment level, which would lead to different potential outcomes. 
\citep{imbens2015causal}
\end{assumption}
Assumption~\ref{assumption:ignorability} and~\ref{assumption:sutva} are standards in causal inference, as discussed in, e.g., \citet{imbens2015causal,shalit2017estimating}. In many estimators such as \citet{shalit2017estimating,oprescu2019orthogonal,nie2021quasi}, $Z$ is the set of the observed covariates, i.e., $Z = X$. In notation of the structural causal model (SCM), $Z$ can be understand as the set of confounders, which is the common cause of treatment and outcome. 
\begin{assumption} \label{assumption:share-covariates}
The individuals from all sources share the same set of common covariates.
\end{assumption}
\begin{assumption}\label{assumption:unique-ident} There exists a set of features such that any individual is uniquely identified across different sources. We refer to this set as `primary key'.
\end{assumption}
\begin{assumption}
\label{assumption:homogeneous-heterogeneous}
Data in different sources are drawn from parts of the population. The multi-source data, which may be homogeneous or heterogeneous in nature, together reflect the characteristics of the population.
\end{assumption}
Most hospitals should collect common covariates of their patients, thus Assumption~\ref{assumption:share-covariates} is reasonable, e.g., decentralized data in \citet{choudhury2019predicting,vaid52federated,flores2021federated} (to name a few) satisfy this assumption for federated learning. 
In Assumption~\ref{assumption:unique-ident}, a `primary key' is not limited to the observed data used for inference as described in Section~\ref{sec:prob-formu}, but it can include any features to uniquely identify an individual, such as $\{\text{nationality, national id}\}$ of a patient.  
Assumption~\ref{assumption:unique-ident} allows a secure preprocessing procedure to remove repeated individual records in different sources, if necessary, without sharing raw data among the sources (see Appendix~\ref{sec:appendix-preprocessing} for details). Assumption~\ref{assumption:homogeneous-heterogeneous} ensures that there is no imbalanced data bias across the sources. In the subsequent sections, we assume that all of the above assumptions 
are satisfied, and the preprocessing procedure is already performed if necessary.

\subsection{Federated Learning of Causal Model}
\label{sec:general-framework}
Federated learning of causal model is important in real-life. For example, multiple healthcare institutions collaboratively analyze patient records without centrally pooling the sensitive information. Each institution retains control over its data, ensuring privacy and security, while contributing insights to a shared model for improved medical research and outcomes. 

\subsubsection{Challenges in federated learning of causal model}
Federated learning of causal model pose a challenge when:
\begin{enumerate}
[label=(c\arabic*)]
\item Raw data cannot be shared  due to privacy issues. For example, patient data is highly confidential and cannot be disclosed to anyone without the explicit consent of the patient. \label{c1}
    \item Estimate distribution of the causal affects, e.g., in Bayesian setting, posterior distributions of the causal effects are computed using data from all sources. \label{c2}
    \item Various data sources may exhibit distinct data distributions, and constructing a causal model based on these sources could introduce bias in estimating heterogeneous causal effects. \label{c3}
    \item The observational data is incomplete, i.e., there are missing values in each source of the data. This situation might lead to a biased estimation. \label{c4}
\end{enumerate}
In the following, we illustrate theses challenges through two examples.

\begin{example}\label{exp1}
Let us illustrate the challenge \ref{c1}+\ref{c2} through an example of estimating distribution of causal effects with potential outcomes framework. 
    The concept of potential outcomes was proposed in \citet{neyman1923application} for randomized trial experiments. \citet{Rubin:1975,rubin:1976a,rubin1977assignment,Rubin:1978} re-formalized the framework for observational studies. We consider the causal effects of a binary treatment $w$, with $w=1$ indicating assignment to `treatment' and $w=0$ indicating assignment to `control'. Following the literature, the causal effect for individual $i$ is defined as a comparison of the two potential outcomes, $y_i(0)$ and $y_i(1)$, where these are the outcomes that would be observed under $w_i=0$ and $w_i=1$, respectively. We can never observe both $y_i(0)$ and $y_i(1)$ for any individual $i$, because it is not possible to go back in time and expose the $i$--th  individual to the other treatment. Consider the Bayesian imputation model of \citet[][]{imbens2015causal}:
\begin{align}
    y_i(0) &= \beta_0^\top x_i + \epsilon_{0i}, &y_i(1) &= \beta_1^\top x_i + \epsilon_{1i},\label{eq:rubin-model}
\end{align}
where $\epsilon_{0i}$ and $\epsilon_{1i}$  are the Gaussian noises. The key to compute treatment effects is $y_i(0)$ and $y_i(1)$. So we need to impute one of the two outcomes. Let $y_{i,\textrm{obs}}$, $y_{i,\textrm{mis}}$  be the observed and unobserved (or missing) outcome. The idea is to find the marginal distribution $\p(y_{i,\textrm{mis}}|\mathbf{y}_{\textrm{obs}},\mathbf{X},\mathbf{w})$, where $\mathbf{y}_{\textrm{obs}},\mathbf{X},\mathbf{w}$ are vectors/matrix of data from all data sources. Once the missing outcomes are imputed, the treatment effects can be estimated. To proceed, \citet[][]{imbens2015causal} suggested four steps based on the following equation $
    \p(y_{i,\textrm{mis}}| \mathbf{y}_{\textrm{obs}},\mathbf{X},\mathbf{w}) = \int \p(y_{i,\textrm{mis}}| \mathbf{y}_{\textrm{obs}},\mathbf{X},\mathbf{w},\bm{\upbeta})\p(\bm{\upbeta}|\mathbf{y}_{\textrm{obs}},\mathbf{X},\mathbf{w})d\bm{\upbeta}$, where $\bm{\upbeta} = \{\beta_0, \beta_1\}$. The aim is to find $\p(y_{i,\textrm{mis}}| \mathbf{y}_{\textrm{obs}},\mathbf{X},\mathbf{w},\bm{\upbeta})$ and $\p(\theta|\mathbf{y}_{\textrm{obs}},\mathbf{X},\mathbf{w})$, and then compute the integral to obtain  $\p(y_{i,\textrm{mis}}| \mathbf{y}_{\textrm{obs}},\mathbf{X},\mathbf{w})$, which is a non-parametric prediction. 
    
\emph{The above procedure shows that learning the distribution $\p(y_{i,\emph{\textrm{mis}}}| \mathbf{y}_{\emph{\textrm{obs}}},\mathbf{X},\mathbf{w})$ would require data from all sources since it is conditional on $\mathbf{y}_{\emph{\textrm{obs}}}$, $\mathbf{X}$, and $\mathbf{w}$. Thus, it violates the data privacy constraint.} 
\end{example}

\begin{example}\label{exp2}
    Consider another example to address the challenges \ref{c1}+\ref{c3} with the structural causal model. Suppose that the confounder $Z\equiv X$ is observed. The key in this problem is to estimate $E[Y|\doo(W=w)]$. This quantity can be estimated from obsrevational data due to the following relation: $E[Y|\doo(W=w)] = \int y p(y|w,\mathbf{x})p(\mathbf{x})d\mathbf{x}dy$, 
    where $p(\mathbf{x})$ can be samples from the observed datasets, and $p(y|w,\mathbf{x})$ can be parameterized and learned from the multiple data sources. To address the dissimilar distribution among the sources, we can learn $p(y|w,x)$ with a transfer kernel approach, where the mean of this distribution takes the following form:
    \begin{align}
        \mu(x, w) = (1-w)\sum_{\mathsf{s}=1}^m\sum_{j=1}^{n_\mathsf{s}} \alpha_{j0}^\mathsf{s}k(x,  x_i^{\mathsf{s}}) + w\sum_{\mathsf{s}=1}^m\sum_{j=1}^{n_\mathsf{s}} \alpha_{j1}^\mathsf{s}k(x,  x_i^{\mathsf{s}}),\label{eq:mu-exp2}
    \end{align}
    where $k(x, x')$ is a transfer kernel function which controls for knowledge transer among multiple sources (see Section~\ref{sec:causalrff} for details of the transfer kernel function), $\alpha_{j0}^\mathsf{s}$ and $\alpha_{j1}^\mathsf{s}$ are learnable parameters. Eq.~(\ref{eq:mu-exp2}) requires \emph{data from all source to perform prediction for the outcome}.
\end{example}

\begin{remark}
    Both Example~\ref{exp1}~and~\ref{exp2} show that a federated learning procedure is required to learn a global causal model where sharing raw data is not allowed. We will generalize Example~\ref{exp1} and \ref{exp2} in Section~\ref{sec:fedci} and \ref{sec:causalrff}, respectively. In Section~\ref{sec:causalfi}, we also propose a model to address \ref{c1}+\ref{c2}+\ref{c4}. In all of these models, we decompose the objective functions into multiple components, each associated with a data source to train, to train in a federated setting. These proposed methods use the same training procedure as outlined in the following subsections.
\end{remark}

\subsubsection{Federated training procedure} 

To learn causal effects from observational data, especially heterogeneous causal effects, it is essential to train a prediction model for the outcomes. For instance, in Example~\ref{exp1}, we need to learn hyperparameters by maximizing marginal likelihood of the observed outcome $\p( \mathbf{y}_{\textrm{obs}}|\mathbf{X},\mathbf{w})$ and similar for Example~\ref{exp2}. To train a causal model in a federated setting, one would need to design a training strategy that aggregates statistics of the local models from each data source. Let $J(\Theta)$ be objective function, where $\Theta$ is the set of parameters or hyperparameters to optimise. For example, in the Bayesian setting, $J(\Theta)$ could be the negative log-marginal likelihood and $\Theta$ could be the set of hyperparameters. To compute $J(\Theta)$, we need data from all sources, which violates the privacy constraint. To proceed, we can approximate $J_{\Theta}$ and decompose it into different components, each associated with local data sources: 
\begin{align}
    J(\Theta) \simeq \sum_{\mathsf{s}=1}^{m} J^{\mathsf{s}}(\Theta).
\end{align}
Under this form of the objective function, we can compute each component $J^{\mathsf{s}}(\Theta)$ using the local data from each source $\mathsf{s}$. Hence, it enables the training of the model in a federated setting by aggregating either the local models or the gradients.

\begin{figure}
        \centering
    \includegraphics[width=0.45\textwidth]{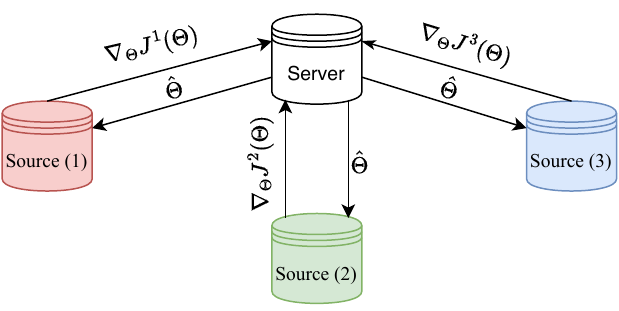}
\caption{An example of our proposed model with three sources. The objective function $J(\Theta) \simeq J^{1}(\Theta)+ J^{2}(\Theta) + J^{3}(\Theta)$ is decomposed to 3 components, each associated with a source.}
        \label{fig:federated-algo}
\end{figure}

\begin{algorithm}\caption{Federated training of causal models}
\label{algo:federated-training}
\SetKwInOut{Parameter}{Parameters}
\Parameter{Let $\Theta$ be set of parameters}
\Begin{
 		Initialize $\Theta$ and send to all source machines\;
	    \Repeat{\emph{stopping condition}}{
    		\For{\emph{ source machine} $\mathsf{s} \in \{1,2,\dots,m\}$}{
    			Compute $\nabla_\Theta J^\mathsf{s}(\Theta)$ and send to server\;
    		}
    		In the central server, do the following steps:\\
    		\Begin{
    		Collect gradients from all sources\;
    		Compute $\nabla_\Theta J(\Theta) = \sum_{\mathsf{s}=1}^m \nabla_\Theta J^\mathsf{s}(\Theta)$\;
    		Update $\Theta \leftarrow \Theta + \mathsf{learning\_rate} \times\nabla_\Theta J(\Theta) $\;
    		Broadcast the new $\Theta$ to all sources\;
    		}
		}
}
\end{algorithm}

\noindent\textbf{Pure federated training:} Figure~\ref{fig:federated-algo} illustrates our proposed federated causal learning algorithm with three sources, where $\Theta$ denotes the set of all parameters to be learned including $\Theta^\mathsf{s}$ and $\lambda^{\mathsf{s},\mathsf{v}}$ from all the sources. The federated learning algorithm can be summarized as follows: First, each source computes the local gradient, $\nabla_\Theta J^{(\mathsf{s})}$, using its own data and sends to the server. The server, then, collects these gradients from all sources and subsequently updates the model. Next, the server broadcasts the new model to all the sources. These steps are repeated until the model converges.

We outline our procedure in Algorithm~\ref{algo:federated-training}. In this algorithm, we introduced a \emph{pure} federated training procedure, which does not incorporate controls for the maximum amount of information that can be disclosed, i.e., differential privacy. Nevertheless, it is feasible to integrate differential privacy within the proposed framework by employing gradient clipping and introducing Gaussian noise to Algorithm~\ref{algo:federated-training}, drawing inspiration from the seminal DP-SGD algorithm proposed by \citet{abadi2016deep}. In this article, our emphasis is on the pure federated training of causal models, deferring the implementation of differential privacy to potential future research endeavors.

A crucial step in implementing Algorithm~\ref{algo:federated-training} is to decompose the objective function. There is no universal method for decomposition, as it depends on the specific model used to represent the outcomes. In Sections~\ref{sec:fedci}, \ref{sec:causalrff}, and \ref{sec:causalfi-incompletedata}, we illustrate various approaches to decompose the objective functions, enabling their training with Algorithm~\ref{algo:federated-training}.

In the following sections, we introduce three instances of the above federated training procedure: FedCI, CausalRFF, and CausalFI, which tailored for estimating causal effects in a federated setting. FedCI employs Bayesian principles and Gaussian processes to effectively calculate the uncertainty associated with causal estimands. CausalRFF strategically addresses data distribution discrepancies among diverse sources, facilitating adaptive knowledge transfer to enhance the precision of causal effect estimation. Meanwhile, CausalFI is specifically designed to handle scenarios involving missing confounders across multiple data sources, ensuring a robust approach to estimating causal effects in complex federated environments.

\section{Bayesian Federated Causal Inference}
\label{sec:fedci}

Estimating the distribution of causal effects is important in real-life applications as it provides uncertainty for decision-making. This section presents a Bayesian method, namely FedCI, that estimates distributions of the causal estimands, hence addressing challenges \ref{c1} and \ref{c2} mentioned in Section~\ref{sec:general-framework}. FedCI is based on the Bayesian imputation model of \citet{imbens2015causal}.

\subsection{The causal estimands}
\label{sec:prob-formu}
We consider the problem setting in Section~\ref{sec:prob-form}. 
For each individual $i$, we can only observe either $y_i^\mathsf{s}(0)$ or  $y_i^\mathsf{s}(1)$, but not both of them. We further denote the unobserved or missing outcome as $y_{i,\textrm{mis}}^\mathsf{s}$. These variables are related to each other through the following equations:
\begin{align}
y_i^\mathsf{s}(1) &= w_i^\mathsf{s} y^\mathsf{s}_{i,\textrm{obs}} + (1-w_i^\mathsf{s})y^\mathsf{s}_{i,\textrm{mis}}, &y_i^\mathsf{s}(0) &= (1-w_i^\mathsf{s}) y^\mathsf{s}_{i,\textrm{obs}} + w_i^\mathsf{s}y^\mathsf{s}_{i,\textrm{mis}}\label{eq:y0-ymis-yobs}.
\end{align}
Thus, $y_i^\mathsf{s}(1) = y^\mathsf{s}_{i,\textrm{obs}}$ when $w_i^\mathsf{s}=1$ and $y_i^\mathsf{s}(1) = y^\mathsf{s}_{i,\textrm{mis}}$ when $w_i^\mathsf{s}=0$, and similarly for $y_i^\mathsf{s}(0)$. 
For notational convenience, we further denote $    \mathbf{y}^\mathsf{s}(0) = [y_1^\mathsf{s}(0),\!...,y_{n_\mathsf{s}}^\mathsf{s}(0)]^\top$,  $\mathbf{y}^\mathsf{s}_{\textrm{obs}} = [y^\mathsf{s}_{1,\textrm{obs}},\!...,y^\mathsf{s}_{n_\mathsf{s},\textrm{obs}}]^\top$, 
and similarly for $\mathbf{y}^\mathsf{s}(1)$, $\mathbf{y}^\mathsf{s}_{\textrm{mis}}$, $\mathbf{X}^\mathsf{s}$ and $\mathbf{w}^\mathsf{s}$. 
We are interested in estimating 
the ITE and the finite-samples ATE:
\begin{align}
\uptau_i^\mathsf{s} &\vcentcolon= y_i^\mathsf{s}(1) - y_i^\mathsf{s}(0), &\uptau &\textstyle\vcentcolon= \sum_{\mathsf{s}=1}^m\sum_{i=1}^{n_\mathsf{s}}\uptau_i^\mathsf{s}/n,\label{eq:ate}
\end{align}
where $y_i^\mathsf{s}(1)$, $y_i^\mathsf{s}(0)$ are realization outcomes of the corresponding random variables, and $n = \sum_{\mathsf{s}=1}^m n_\mathsf{s}$. Inserting Eq.~(\ref{eq:y0-ymis-yobs}) into (\ref{eq:ate}), we obtain the estimate of ITE:
\begin{align}
&\e[\uptau^\mathsf{s}_i]
=(2w_i^\mathsf{s}-1) (y^\mathsf{s}_{i,\textrm{obs}} - \e\big[y^\mathsf{s}_{i,\textrm{mis}}\big| \mathbf{y}_{\textrm{obs}}, \mathbf{X}, \mathbf{w}\big]),\label{eq:ite-hat-expectation}\\ 
&\V\text{ar}[\uptau^\mathsf{s}_i] =(2w_i^\mathsf{s}-1)^2\mathbb{V} \text{ar}\left[y^\mathsf{s}_{i,\textrm{mis}}\big| \mathbf{y}_{\textrm{obs}}, \mathbf{X}, \mathbf{w}\right],
\end{align}
where $\mathbf{y}_{\textrm{obs}}$, $\mathbf{X}$, $\mathbf{w}$ denote the vectors/matrices of the observed outcomes, covariates and treatments concatenated from all the sources. The estimate of ATE is as follows:
\begin{align}
&\!\!\!\!\e[\uptau] \!=\!(2\mathbf{w}-\mathbf{1})^\top\!(\mathbf{y}_{\textrm{obs}} \!-\! \e[\mathbf{y}_{\text{mis}}| \mathbf{y}_{\textrm{obs}}, \mathbf{X}, \mathbf{w}])/n,\\
&\!\!\!\!\V\text{ar}[\uptau] \!=\!(2\mathbf{w}\!-\!\mathbf{1})^\top\!\mathbb{C} \text{ov}[\mathbf{y}_{\textrm{mis}}| \mathbf{y}_{\textrm{obs}}, \mathbf{X}, \mathbf{w}](2\mathbf{w}\!-\!\mathbf{1})/n^2,\label{eq:tau-hat-variance}
\end{align}
where $\mathbf{1}$ is a vector of ones. 

Hence, the remaining task is to learn the posterior $\p(\mathbf{y}_{\textrm{mis}}\big| \mathbf{y}_{\textrm{obs}}, \mathbf{X}, \mathbf{w})$, which is the predictive distribution of $\mathbf{y}_{\textrm{mis}}$ given \emph{all} the covariates, treatments and observed outcomes from \emph{all} sources.

\subsection{GP-based Imputation}
\label{sec:rubin-to-gp}
The model presented in Eq.~(\ref{eq:rubin-model}) is a simple Bayesian linear model. In this section, we present a more general nonlinear Bayesian model. In particular, since $\bm{\upbeta}_0$ and $\bm{\upbeta}_1$ in Eq.~(\ref{eq:rubin-model}) follow multivariate normal distributions, the two components $\beta_0^\top x_i$ and $\beta_1^\top x_i$ also follow multivariate normal distributions. The generalisation of these two components are $f_0(x_i) = \beta_0^\top\omega(x_i)$ and $f_1(x_i) = \beta_1^\top\omega(x_i)$, where $\omega(x_i)$ is a vector of basis functions with input $\mathbf{x}_i$. This formulation would lead to the fact that the marginal of $f_0(x)$ and $f_1(x)$ are Gaussian processes (GPs). Thus, we propose:\begin{align}
    y_i(0) &= f_0(x_i) + \epsilon_{0i}, &y_i(1) &= f_1(x_i) + \epsilon_{1i},\label{eq:gp-model}
\end{align}
where $f_0(x_i)$ and $f_1(x_i)$ are two random functions evaluated at $x_i$, i.e., $f_0(x_i) \sim \mathsf{GP}(\mu_0(\mathbf{X}), \mathbf{K})$ and $f_1(x_i) \sim \mathsf{GP}(\mu_1(\mathbf{X}), \mathbf{K})$, where $\mathbf{K}$ denotes the covariance matrix computed with a kernel function $\mathsf{k}(x,x')$. Similar to the imputation model of \citet{imbens2015causal}, 
this model also requires finding the marginal distribution $\p(y_{i,\textrm{mis}}\,|\,\mathbf{y}_{\textrm{obs}},\mathbf{X},\mathbf{w})$, \emph{through accessing the observed data from all the sources.} 

\emph{Similarly, although this model is generic, it requires access to all the observed data to compute $\mathbf{K}$, which is impossible without violating the privacy constraints mentioned above.} 
In the subsequent sections, we propose a federated model to address this problem.

\subsection{The Proposed Model}
\label{sec:model}
Recall that the aim is to find $\p(\mathbf{y}_{\textrm{mis}}\,|\, \mathbf{y}_{\textrm{obs}}, \mathbf{X}, \mathbf{w})$ so that we may in turn compute Eqs.~(\ref{eq:ite-hat-expectation})-(\ref{eq:tau-hat-variance}) to arrive at the quantities of interest. To that end, we propose to model the joint distribution of the potential outcomes as follows:
\begin{align}
\begin{bmatrix}
	y_i^\mathsf{s}(0)\\
	y_i^\mathsf{s}(1)
	\end{bmatrix} = \Psi^{\frac{1}{2}}\left(\begin{bmatrix}
	f_0^\mathsf{s}(x_i)\\
	f_1^\mathsf{s}(x_i)
	\end{bmatrix} + \begin{bmatrix}
	g_0^\mathsf{s}\\
	g_1^\mathsf{s}
	\end{bmatrix}\right) + \Sigma^{\frac{1}{2}}\bm{\upvarepsilon}_i^\mathsf{s},
\label{eq:the-model}
\end{align}
where $\bm{\upvarepsilon}_i^\mathsf{s} \sim \mathsf{N}(\mathbf{0}, \mathbf{I}_2)$  is to handle the noise of the outcomes. 

As mentioned in Section \ref{sec:rubin-to-gp}, all the outcomes from all sources are \emph{interdependent} in the Bayesian imputation approach, which is problematic for federated learning. This dependency is handled via $f_j^\mathsf{s}(x_i)$ and $g_j^\mathsf{s}$ ($j\in\{0,1\}$), which enable federated learning for the proposed model. We refer to the dependency handled by $f_j^\mathsf{s}(x_i)$ as intra-dependency and the one captured by $g_j^\mathsf{s}$ as inter-dependency. 

\noindent\textbf{Intra-dependency.} 
$f_0^\mathsf{s}(x_i)$ and $f_1^\mathsf{s}(x_i)$ are GP-distributed functions, which allows us to model each source dataset simultaneously along with its heterogeneous correlations. Specifically, we model $f_0^\mathsf{s}(x_i) \sim \mathsf{GP}(\mu_0(\mathbf{X}^\mathsf{s}), \mathbf{K}^\mathsf{s})$ and $f_1^\mathsf{s}(x_i) \sim \mathsf{GP}(\mu_1(\mathbf{X}^\mathsf{s}), \mathbf{K}^\mathsf{s})$, where $\mathbf{K}^\mathsf{s}$ is a covariance matrix computed by a kernel function $\mathsf{k}(x_i^\mathsf{s}, x_j^\mathsf{s})$, and $\mu_0(\cdot)$, $\mu_1(\cdot)$ are functions modelling the mean of these GPs. Parameters of these functions and hyperparameters in the kernel function are shared across multiple sources. 
The above GPs handle the correlations within one source only. 

\noindent\textbf{Inter-dependency.} 
To capture \textit{dependency} among the sources, we introduce variable $\mathbf{g} = [\mathbf{g}_0, \mathbf{g}_1]$, where $\mathbf{g}_0 = [g_0^{1},\!...,g_0^{m}]^\top \sim \mathsf{N}(\bm{r}_0, \mathbf{M})$ and $\mathbf{g}_1 = [g_1^{1},\!...,g_1^{m}]^\top \sim \mathsf{N}(\bm{r}_1, \mathbf{M})$. 
Both $g_0^\mathsf{s}$ and $g_1^\mathsf{s}$ are shared within the source $\mathsf{s}$, and they are correlated across multiple sources $\mathsf{s} \in \{1,\!...,m\}$. The correlations among the sources are modelled via the covariance matrix $\mathbf{M}$ 
which is computed with a kernel function.  The inputs to the kernel function are the sufficient statistics (we used mean, variance, skewness, and kurtosis) of each covariate $x^\mathsf{s}$ within the source $\mathsf{s}$. We denote the first four moments of covariates as $\mathbf{\widetilde{x}}^\mathsf{s} \in \mathbb{R}^{4 d_x \times 1}$ and the kernel function as $\gamma(\mathbf{\widetilde{x}}^\mathsf{s}, \mathbf{\widetilde{x}}^{\mathsf{s}'})$, which evaluates the correlation of two sources $\mathsf{s}$ and $\mathsf{s}'$. This formulation implies that $\mathbf{g}_0$ and $\mathbf{g}_1$ are GPs. The elements of $\bm{r}_0$ and $\bm{r}_1$ are computed with the mean functions $r_0(\mathbf{\widetilde{x}}^\mathsf{s})$ and $r_1(\mathbf{\widetilde{x}}^\mathsf{s})$, respectively. Herein, we only share the sufficient statistics of covariates, but not covariates of a specific individual. 

\noindent\textbf{The two variables $\Psi$ and $\Sigma$.} 
These are positive semi-definite matrices capturing the correlations between the two possible outcomes $y_i^\mathsf{s}(0)$ and $y_i^\mathsf{s}(1)$, $\Psi^{\frac{1}{2}}$ and $\Sigma^{\frac{1}{2}}$ are their Cholesky decomposition matrices. Note that $\Psi$ and $\Sigma$ are also random variables. Since these are positive semi-definite matrices, we model their priors using Wishart distribution $\Psi  \sim \mathsf{Wishart}(\mathbf{V}_0, d_0)$, $\Sigma \sim \mathsf{Wishart}(\mathbf{S}_0, n_0)$, where $\mathbf{V}_0, \mathbf{S}_0 \in \mathbb{R}^{2\times 2}$ are predefined positive semi-definite matrices and $d_0, n_0 \ge 2$ are predefined degrees of freedom.

\subsection{The Proposed Algorithm}
\label{sec:inference}
Based on some results on the joint distribution of potential outcomes, we construct a federated objective function for the proposed federated causal inference algorithm (FedCI).

\subsubsection{Joint Distribution of the Outcomes}

We first present some results that are helpful in constructing the federated objective function in Section~\ref{sec:objective-function}. 
The proofs of these results are in the appendices. To simplify the exposition, we denote $\mathbf{g}^\mathsf{s} = [\mathbf{g}_0^\mathsf{s}, \mathbf{g}_1^\mathsf{s}]$, where $\mathbf{g}_0^\mathsf{s} = [g_0^\mathsf{s},\!..., g_0^\mathsf{s}]^\top$ and $\mathbf{g}_1^\mathsf{s} = [g_1^\mathsf{s},\!..., g_1^\mathsf{s}]^\top$. 

\begin{lemma}
\label{lem:joint-prob-s}
Let $\Psi$, $\Sigma$, $\mathbf{K}$, $\mu_0(\mathbf{X}^\mathsf{s})$, $\mu_1(\mathbf{X}^\mathsf{s})$, and $\mathbf{g}^\mathsf{s}$  satisfy the model in Eq.~\emph{(\ref{eq:the-model})}. Then,
\begin{align*}
&\begin{bmatrix}
\mathbf{y}^\mathsf{s}(0)\\
\mathbf{y}^\mathsf{s}(1)
\end{bmatrix}\Big|\Psi, \Sigma, \mathbf{X}^\mathsf{s}, \mathbf{w}^\mathsf{s}, \mathbf{g}^\mathsf{s} \sim \mathsf{N}\left( \left(\Psi^{\frac{1}{2}} \otimes \mathbf{I}_{n_\mathsf{s}}\right)\begin{bmatrix}\mu_0(\mathbf{X}^\mathsf{s}) + \mathbf{g}_0^\mathsf{s}\\\mu_1(\mathbf{X}^\mathsf{s}) + \mathbf{g}_1^\mathsf{s}
\end{bmatrix}, \Psi \otimes \mathbf{K}^\mathsf{s} + \Sigma \otimes \mathbf{I}_{n_\mathsf{s}}\right),
\end{align*}
where $\otimes$ is the Kronecker product.
\end{lemma}
The proof of Lemma~\ref{lem:joint-prob-s} is presented in Appendix~\ref{sec:appendix-proof-lem-1}. From Lemma~\ref{lem:joint-prob-s}, we observe that $\Psi$, $\mathbf{K}^\mathsf{s}$, $\Sigma$, and $\mathbf{I}_{n_\mathsf{s}}$ are positive semi-definite, thus the covariance matrix $\Psi \otimes \mathbf{K}^\mathsf{s} + \Sigma \otimes \mathbf{I}_{n_\mathsf{s}}$ is positive semi-definite due to the fundamental property of Kronecker product. This is why we choose $\Psi$ and $\Sigma$ to be positive semi-definite in our model; otherwise, the covariance matrix is invalid. From Lemma~\ref{lem:joint-prob-s}, we can obtain the result in Lemma~\ref{lem:3} as follows:
\begin{lemma}
\label{lem:3}
Let $\Psi$, $\Sigma$, $\mathbf{K}$, $\mu_0(\mathbf{X}^\mathsf{s})$, $\mu_1(\mathbf{X}^\mathsf{s})$, and $\mathbf{g}^\mathsf{s}$  satisfy the model in Eq.~\emph{(\ref{eq:the-model})}. Then,
\begin{align*}
&\begin{bmatrix}
\mathbf{y}^\mathsf{s}_{\emph{\textrm{obs}}}\\
\mathbf{y}^\mathsf{s}_{\emph{\textrm{mis}}}
\end{bmatrix}\Big|\Psi, \Sigma, \mathbf{X}^\mathsf{s}, \mathbf{w}^\mathsf{s}, \mathbf{g}^\mathsf{s} \sim \mathsf{N}\left( \begin{bmatrix}\mu_{\emph{\textrm{obs}}}(\mathbf{X}^\mathsf{s})\\\mu_{\emph{\textrm{mis}}}(\mathbf{X}^\mathsf{s})
\end{bmatrix} , \begin{bmatrix}
\mathbf{K}_{\emph{\textrm{obs}}}^\mathsf{s}&\mathbf{K}_{\emph{\textrm{om}}}^\mathsf{s}\\
(\mathbf{K}_{\emph{\textrm{om}}}^\mathsf{s})^\top&\mathbf{K}_{\emph{\textrm{mis}}}^\mathsf{s}\end{bmatrix}\right).
\end{align*}
The mean functions $\mu_{\emph{\textrm{obs}}}(\mathbf{X}^\mathsf{s})$ and $\mu_{\emph{\textrm{mis}}}(\mathbf{X}^\mathsf{s})$ are:
\begin{align*}
\mu_{\emph{\textrm{obs}}}(\mathbf{X}^\mathsf{s}) = (\mathbf{1} - \mathbf{w}^\mathsf{s})\odot\mathbf{m}_0 + \mathbf{w}^\mathsf{s} \odot\mathbf{m}_1, \qquad
\mu_{\emph{\textrm{mis}}}(\mathbf{X}^\mathsf{s}) = \mathbf{w}^\mathsf{s} \odot\mathbf{m}_0 + (\mathbf{1} - \mathbf{w}^\mathsf{s}) \odot\mathbf{m}_1,
\end{align*}
where we denote $\mathbf{m}_0 = \psi_{11}^\ast(\mu_0(\mathbf{X}^\mathsf{s}) + \mathbf{g}_0^\mathsf{s})$ and $\mathbf{m}_1 = \psi_{21}^\ast(\mu_0(\mathbf{X}^\mathsf{s}) + \mathbf{g}_0^\mathsf{s}) + \psi_{22}^\ast(\mu_1(\mathbf{X}^\mathsf{s}) + \mathbf{g}_1^\mathsf{s})$ with $\psi^\ast_{ij}$ is the $(i,j)$--th element of Cholesky decomposition matrix of $\Psi$, $\mathbf{1}$ is a vector ones, and $\odot$ is the element-wise product. The covariance matrices $\mathbf{K}^\mathsf{s}_{\textrm{\emph{\textrm{obs}}}}$, $\mathbf{K}^\mathsf{s}_{\textrm{\emph{\textrm{mis}}}}$, and $\mathbf{K}^\mathsf{s}_{\textrm{\emph{\textrm{om}}}}$ are computed by kernel functions:
\begin{align*}
k_{\emph{\textrm{obs}}}(x_i, x_j) &= \big[(1-w_i)(1-w_j)\psi_{11} + w_iw_j\psi_{22} + (1-w_i)w_j\psi_{12} + w_i(1-w_j)\psi_{21}\big] \mathsf{k}(x_i, x_j)\\&\,\,\,\,\, + \big[(1-w_i)\sigma_{11} + w_i\sigma_{22}\big] \mathds{1}_{i=j},\\k_{\emph{\textrm{mis}}}(x_i, x_j) &= \big[w_iw_j\psi_{11} + (1-w_i)(1-w_j)\psi_{22} + (1-w_i)w_j\psi_{21} + w_i(1-w_j)\psi_{12}\big] \mathsf{k}(x_i, x_j)\\& \,\,\,\,\,+ \big[w_i\sigma_{11} + (1-w_i)\sigma_{22}\big] \mathds{1}_{i=j},
\\k_{\emph{\textrm{om}}}(x_i, x_j) &= \big[(1-w_i)(1-w_j)\psi_{21} + w_iw_j\psi_{12} + (1-w_i)w_j\psi_{22} + w_i(1-w_j)\psi_{11}\big] \mathsf{k}(x_i, x_j) \\&\,\,\,\,\,+ \big[(1-w_i)\sigma_{21} + w_i\sigma_{12}\big] \mathds{1}_{i=j},
\end{align*}
where $\psi_{ab}$ and $\sigma_{ab}$ are the $(a,b)$--th elements of $\Psi$ and $\Sigma$, respectively.
\end{lemma}

The proof of Lemma~\ref{lem:3} is in Appendix~\ref{sec:appendix-proof-lem-2}. Lemma~\ref{lem:3} has two important roles in our work. First, we can obtain the conditional likelihood to help infer the parameters and hyperparameters of our proposed model. Second, we can also obtain the posterior of $\mathbf{y}^\mathsf{s}_\textrm{mis}$ to help us estimate ITE and ATE.

\subsubsection{Federated Objective Function}
\label{sec:objective-function}
The proposed model in Eq.~(\ref{eq:the-model}) would lead to an objective function that can be decomposed into $m$ components, each associated with a data source. Since estimating $\p(\mathbf{y}^\mathsf{s}_{\textrm{mis}}\,\big|\, \mathbf{y}^\mathsf{s}_{\textrm{obs}}, \mathbf{X}^\mathsf{s}, \mathbf{w}^\mathsf{s})$ exactly is intractable, we sidestep this intractability via a variational approximation.  To achieve this, we maximize the following evidence lower bound (ELBO) $J$:
\begin{align}
    \log\p(\mathbf{y}_{\textrm{obs}}\,|\,\mathbf{X},\mathbf{w}) &= \log \int\p(\mathbf{y}_{\textrm{obs}},\mathbf{g}, \Psi, \Sigma\,|\,\mathbf{X},\mathbf{w}) d\mathbf{g}d\Psi d\Sigma\ge \sum_{\mathsf{s}=1}^m J^\mathsf{s} =\vcentcolon J,\label{eq:loss}
\end{align}
where \begin{align*}
J^\mathsf{s} &= \e_q\Big[ \log\p(\mathbf{y}^\mathsf{s}_{\textrm{obs}} | \cdot)\Big]  -\frac{1}{m}\left(\!\sum_{z \in \{\mathbf{g}, \Psi, \Sigma\}}\!\D_{\text{KL}}[\q(z)\|\p(z)]\right).\end{align*}
Herein, $\D_{\text{KL}}[\cdot]$ is the Kullback–Leibler divergence. Details of the ELBO are presented in Appendix~\ref{sec:appendix-elbo}. The conditional likelihood $\p(\mathbf{y}^\mathsf{s}_{\textrm{obs}} | \cdot)$ is obtained from Lemma~\ref{lem:3} by marginalizing out $\mathbf{y}^\mathsf{s}_{\textrm{mis}}$, i.e.,
\begin{align}
    \!\p(\mathbf{y}^\mathsf{s}_{\textrm{obs}} | \mathbf{X}^\mathsf{s}, \mathbf{w}^\mathsf{s}, \Psi, \Sigma, \mathbf{g}^\mathsf{s}) = \mathsf{N}(\mathbf{y}^\mathsf{s}_{\textrm{obs}};\mu_{\textrm{obs}}(\mathbf{X}^\mathsf{s}), \mathbf{K}_{\textrm{obs}}^\mathsf{s}).
\end{align}
The above conditional likelihood is free of $\sigma_{21}$ and $\sigma_{12}$, which capture the correlation of two potential outcomes. Thus the posteriors of these variables would coincide with their priors, i.e., the correlation cannot be learned but set as a prior. This is well-known as one of the potential outcome cannot be observed \citep{imbens2015causal}. 
In Eq.~(\ref{eq:loss}), the ELBO $J$ is derived from the of joint marginal likelihood of all $m$ sources, and it is factorized into $m$ components $J^\mathsf{s}$, each component corresponds to a source. This enables federated optimization of $J$. The first term of $J^\mathsf{s}$ is expectation of the conditional likelihood with respect to the variational posterior $q(\mathbf{g}, \Psi, \Sigma)$, thus this distribution is learned from data of all the sources. In the following, we present its factorization. 

\noindent\textbf{Variational posterior distributions.} 
We use the typical mean-field approximation to factorize among the variational posteriors $
\q(\Psi, \Sigma, \mathbf{g}) =\q(\Psi)\,\q(\Sigma)\,\q(\mathbf{g})$. Let $\mathbf{\widetilde{y}}_{\textrm{obs}}^\mathsf{s}(0)$, $\mathbf{\widetilde{y}}_{\textrm{obs}}^\mathsf{s}(1)$, $\mathbf{\widetilde{x}}^{\mathsf{s}}$, and $\mathbf{\widetilde{w}}^\mathsf{s}$ ($\mathsf{s} = 1,2,\!...,m$) be the first four moments of the observed outcomes, covariates, and treatment of the $\mathsf{s}$--th source. Let $\mathbf{\widetilde{X}} = [\mathbf{\widetilde{x}}^{1},\!...,\mathbf{\widetilde{x}}^{m}]^\top$, $\mathbf{\widetilde{y}}_{\textrm{obs}}(0) = [\mathbf{\widetilde{y}}_{\textrm{obs}}^{1}(0),\!...,\mathbf{\widetilde{y}}_{\textrm{obs}}^{m}(0)]^\top$, $\mathbf{\widetilde{y}}_{\textrm{obs}}(1) = [\mathbf{\widetilde{y}}_{\textrm{obs}}^{1}(1),\!...,\mathbf{\widetilde{y}}_{\textrm{obs}}^{m}(1)]^\top$, and $\mathbf{\widetilde{w}} = [\mathbf{\widetilde{w}}^{1},\!...,\mathbf{\widetilde{w}}^{m}]^\top$. We parameterize 
\begin{align*}
    \q(\mathbf{g}) =\textstyle \prod_{j\in\{0,1\}}\mathsf{N}(\mathbf{g}_j;h_j(\mathbf{\widetilde{y}}_{\textrm{obs}}(0), \mathbf{\widetilde{y}}_{\textrm{obs}}(1), \mathbf{\widetilde{X}}, \mathbf{\widetilde{w}}), \mathbf{U}),
\end{align*}
where $h_0(\cdot)$ and $h_1(\cdot)$ are the mean functions, $\mathbf{U}$ is the covariance matrix computed with a kernel function $\kappa(u^\mathsf{s}, u^{\mathsf{s}'})$, where $u^\mathsf{s}\vcentcolon= [\mathbf{\widetilde{y}}^\mathsf{s}_{\textrm{obs}}(0), \mathbf{\widetilde{y}}^\mathsf{s}_{\textrm{obs}}(1), \mathbf{\widetilde{x}}^\mathsf{s}, \mathbf{\widetilde{w}}^\mathsf{s}]$. 

Since $\Psi$ and $\Sigma$ are positive semi-definite matrices, we model their variational posterior as Wishart distribution: 
\begin{align*}
    	\q(\Psi) &= \mathsf{Wishart}(\Psi;\mathbf{V}_q, d_q), &\q(\Sigma) &= \mathsf{Wishart}(\Sigma;\mathbf{S}_q, n_q),
\end{align*}
where $d_q, n_q$ are degrees of freedom, $\mathbf{V}_q, \mathbf{S}_q$ are the scale matrices. We set the form of these scale matrices as follows
\begin{align*}
\mathbf{V}_q &= \begin{bmatrix}
\nu_{1}^2&\rho\nu_{1}\nu_2\\
\rho\nu_{1}\nu_2&\nu_{2}^2
\end{bmatrix}, &\mathbf{S}_q &= \begin{bmatrix}
\delta_{1}^2&\eta\delta_{1}\delta_2\\
\eta\delta_{1}\delta_2&\delta_{2}^2
\end{bmatrix},
\end{align*}
where $\nu_i, \rho, \delta_i, \eta$ are parameters to be learned and $\rho, \eta \in [0,1]$.

\noindent\textbf{Reparameterization.} To maximize the ELBO, we approximate the expectation in $J^\mathsf{s}$ with Monte Carlo integration, which requires drawing samples of $\mathbf{g}$, $\Psi$ and $\Sigma$ from their variational distributions. This requires a reparameterization to allow the gradients to pass through the random variables $\mathbf{g}$, $\Psi$ and $\Sigma$. 
The reparameterization trick for $\mathbf{g}$ are:
$
\mathbf{g}_j = h_j(\mathbf{\widetilde{y}}_{\textrm{obs}}(0), \mathbf{\widetilde{y}}_{\textrm{obs}}(1), \mathbf{\widetilde{X}}, \mathbf{\widetilde{w}}) + \mathbf{U}^{\frac{1}{2}} \bm{\xi}_j, j\in\{0,1\}$, 
where $\bm{\xi}_j \sim \mathsf{N}(\bm{0},\mathbf{I}_m)$ and $\mathbf{U}^{\frac{1}{2}}$ is the Cholesky decomposition matrix of $\mathbf{U}$.
Since $\q(\Psi)$ is a Wishart distribution, we introduce the following procedure to draw $\Psi$: $\Psi = \mathbf{V}_q^{\frac{1}{2}} \bm{\zeta} ( \mathbf{V}_q^{\frac{1}{2}})^\top, \bm{\zeta} \sim \mathsf{Wishart}(\mathbf{I}_2, d_q)$, 
where $\mathbf{V}_q^{\frac{1}{2}}$ is the Choleskey decomposition matrix of $\mathbf{V}_q$. Likewise, we also apply this procedure to draw $\Sigma$.

\noindent\textbf{Federated optimization algorithm.} 
With the above model and its objective function, we can compute gradients of the learnable parameters separately in each source without sharing data to a central server. Hence it enables federated learning with Algorithm~\ref{algo:federated-training}.

\subsubsection{Predicting Causal Effects from Multiple Sources}
To understand why data from all the sources can help predict causal effects in a source $\mathsf{s}$, we observe that \begin{align}
    \p(\mathbf{y}^\mathsf{s}_\textrm{mis}\,\big| \mathbf{y}_\textrm{obs}, \mathbf{X}, \mathbf{w}) \label{eq:predictive-dist}
    &\simeq \e_{\q} \big[p(\mathbf{y}^\mathsf{s}_\textrm{mis}\big| \mathbf{y}^\mathsf{s}_\textrm{obs},\mathbf{X}^\mathsf{s}, \mathbf{w}^\mathsf{s}, \Psi, \Sigma, \mathbf{g})\big] \\
&= p(\mathbf{y}^\mathsf{s}_\text{mis}\,\big| \underbrace{\addstackgap[1.3pt]{$\mathbf{y}^\mathsf{s}_\text{obs}, \mathbf{X}^\mathsf{s}, \mathbf{w}^\mathsf{s}$}}_\textbf{(i)}, \underbrace{\addstackgap[4.6pt]{$\Theta $}}_\textbf{(ii)}, \underbrace{\addstackgap[2pt]{$\mathbf{\widetilde{y}}_{\textrm{obs}}(0), \mathbf{\widetilde{y}}_{\textrm{obs}}(1), \mathbf{\widetilde{X}}, \mathbf{\widetilde{w}}$}}_\textbf{(iii)}).\nonumber
\end{align}
Eq.~(\ref{eq:predictive-dist}) is an approximation of the predictive distribution of the missing outcomes $\mathbf{y}^\mathsf{s}_\textrm{mis}$ and it depends on the following three components:
\begin{enumerate}[noitemsep,topsep=0pt,leftmargin=*,label=\textbf{(\roman*).}]
    \item The observed outcomes, covariates and treatment from the same source $\mathsf{s}$.
    \item The shared parameters $\Theta$ learned from data of all the sources.
    \item Sufficient statistics of the observed data from all the sources.\end{enumerate}
The two last components \textbf{(ii)} and \textbf{(iii)} indicate that the predictive distribution in source $\mathsf{s}$ utilizes knowledge from all the sources through $\Theta$ and the sufficient statistics $[\mathbf{\widetilde{y}}_{\textrm{obs}}(0)$, $\mathbf{\widetilde{y}}_{\textrm{obs}}(1)$, $\mathbf{\widetilde{X}}$, $\mathbf{\widetilde{w}}]$. 
This explain why data from all of the sources help predict missing outcomes in source $\mathsf{s}$.

\subsection{Experiments}
\label{sec:experiment}
\textbf{The baselines and experimental objectives.} 
We first examine the performance of FedCI. We then compare the performance of FedCI against recent causal inference methods, such as BART \citep{hill2011bayesian}, 
TARNet, CFR Wass (CFRNet with Wasserstein distance), CFR MMD (CFRNet with maximum mean discrepancy distance) \citep{shalit2017estimating}, 
CEVAE \citep{louizos2017causal}, 
OrthoRF \citep{oprescu2019orthogonal}, 
X-learner \citep{kunzel2019metalearners}, and R-learner \citep{nie2021quasi}. All these methods do not consider learning causal effects in a federated setting. This analysis aims to show the efficacy of FedCI as compared with the baselines trained in three different cases: (\textbf{1}) training a local model on each source data, (\textbf{2}) training a global model with the combined data of all sources, (\textbf{3}) using bootstrap aggregating (also known as bagging, which is an ensemble learning method) of \citet{breiman1996bagging} where $m$ models are locally trained on each source data; then taking average of the predicted treatment effects of each model. Although case (\textbf{2})  \textit{violates} the privacy constraint of federated data, we use it  for comparison purposes. In general, we would like 
to assess the performance of the federated causal inference approach against the baselines using combined data in case~(\textbf{2}).

We use publicly available libraries and source codes to implement the baseline methods. In particular, CEVAE, TARNet, CFR Wass, and CFR MMD are readily available on github. We use the online packages \texttt{BartPy} for BART,  \texttt{causalml} \citep{chen2020causalml} for X-learner  and R-learner, and 
\texttt{econml} \citep*{econml} for OrthoRF. 
For all the methods, we fine-tune the learning rate in $\{10^{-1}, 10^{-2}, 10^{-3}, 10^{-4}\}$ and regularizers in $\{10^{1}, 10^{0}, 10^{-1}, 10^{-2}, 10^{-3}\}$. 

\noindent\textbf{Evaluation metrics.} 
We report two evaluation metrics: (i) precision in estimation of heterogeneous effects (PEHE) \citep{hill2011bayesian} for evaluating ITE: \begin{align*}
    \epsilon_\textrm{PEHE} \vcentcolon= \sum_{\mathsf{s}=1}^m\sum_{i=1}^{n_\mathsf{s}}(\tau^{\mathsf{s}}_i - \hat{\tau}^{\mathsf{s}}_i)^2/(m n_\mathsf{s}),
\end{align*}
and (ii) absolute error for evaluating ATE:
\begin{align*}
    \epsilon_\textrm{ATE} \vcentcolon=  |\tau-\hat{\tau}|, 
\end{align*}
where $\tau_{i}^{\mathsf{s}}$ and $\tau$ are the \textit{true} ITE and \textit{true} ATE, respectively, and $\hat{\tau}_{i}^{\mathsf{s}}$,  $\hat{\tau}$ are their estimates. 

These evaluation 
metrics are for point estimates, which are the mean of ITE and ATE in their estimated distributions. 
We also report the estimated distribution of ATE in our model.

\subsubsection{Synthetic Data}
\label{sec:synthetic-data}
We analyses FedCI in terms of three types of outcomes: (1) real-value, (2)  binary, and (3) count. While (1) is examined in a well-specified case for the outcomes, (2) and (3) are studied in misspecified cases.

\noindent\textbf{Real-value Outcomes}

\noindent\emph{\underline{Data.}} Obtaining ground truth for evaluating causal inference algorithm is challenging. Thus, most methods are evaluated using synthetic or semi-synthetic datasets. In this experiment, we simulate the data with the following distributions:
\begin{align*}
    x_{ij} &\!\sim\! \mathsf{U}[-1,1], &\!\!y_i(0) &\!\sim\! \mathsf{N}(\lambda(b_0 \!+\! x_i^\top \mathbf{b}_1), \sigma_0^2), \\[-0.1cm]
    w_i &\!\sim\! \mathsf{Bern}(\varphi(a_0 \!+\! x_i^\top \mathbf{a}_1)), &\!\!y_i(1) &\!\sim\! \mathsf{N}(\lambda(c_0 \!+\! x_i^\top \mathbf{c}_1), \sigma_1^2),
\end{align*}
where $\varphi(\cdot)$ is the sigmoid function, $\lambda(\cdot)$ is the softplus function, and $x_i = [x_{i1},\!...,x_{id_x}]^\top \in \mathbb{R}^{d_x}$ with $d_x=20$. We simulate two synthetic datasets: DATA-1 and DATA-2. For DATA-1, the ground truth parameters are randomly set as follows: $\sigma_0=\sigma_1=1$,  $(a_0,b_0,c_0)=(0.6,0.9,2.0)$, $\mathbf{a}_1 \sim \mathsf{N}(\mathbf{0}, 2\cdot\mathbf{I}_{d_x})$, $\mathbf{b}_1 \sim \mathsf{N}(\mathbf{0}, 2\cdot\mathbf{I}_{d_x})$, $\mathbf{c}_1 \sim \mathsf{N}(\mathbf{1}, 2\cdot\mathbf{I}_{d_x})$. For DATA-2, we set $(b_0,c_0) = (6,30)$, $\mathbf{b}_1 \sim \mathsf{N}(10\cdot\mathbf{1}, 2\cdot\mathbf{I}_{d_x})$, $\mathbf{c}_1 \sim \mathsf{N}(15\cdot\mathbf{1}, 2\cdot\mathbf{I}_{d_x})$, and the other parameters are set similar to that of DATA-1. The purpose is to make two different scales of the outcomes for the two datasets. For each dataset, we simulate $10$ replications with $n = 5000$ records. We only keep $\{(y_i, w_i, x_i)\}_{i=1}^n$ as the observed data, where $y_i = y_i(0)$ if $w_i=0$ and $y_i = y_i(1)$ if $w_i=1$. We divide the data into five sources, each consists of $n_\mathsf{s}=1000$ records. In each source, we use $50$ records for training, $450$ for testing and $400$ for validation. We report the evaluation metrics and their standard errors over the 10 replications. The parameters chosen for this simulation study satisfy Assumption~\ref{assumption:ignorability} since $y_i(0)$ and $y_i(1)$ are independent of $w_i$ given $x_i$. Assumption~\ref{assumption:sutva} is respected as the treatment on an individual $i$ does not effect the outcome of another individual $j$ ($i\neq j$). Since we fixed the dimension of $x_i$ and draw it from the same distribution, Assumption~\ref{assumption:share-covariates} is implicitly satisfied. Assumption~\ref{assumption:unique-ident} holds true since each record drawn from the above distributions is attributed to one individual. This means that there are no duplicates of individuals in more than one source. Assumption~\ref{assumption:homogeneous-heterogeneous} is also satisfied since we have divided the data equally from one dataset. \begin{figure}
\centering
    \includegraphics[width=0.7\textwidth]{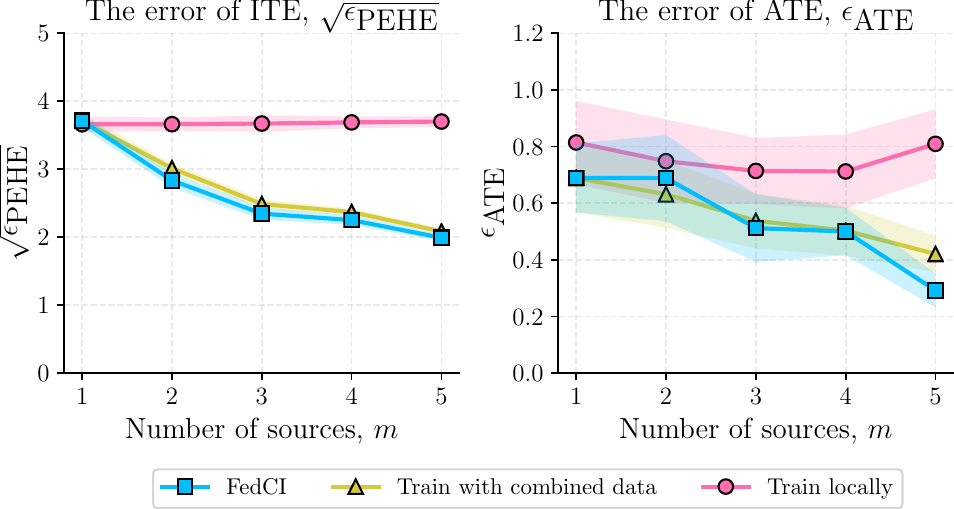}
\caption{Federated inference analysis on DATA-1.}
\label{fig:fedci-alalysis}
\end{figure}
\begin{figure}
\centering
    \includegraphics[width=0.7\textwidth]{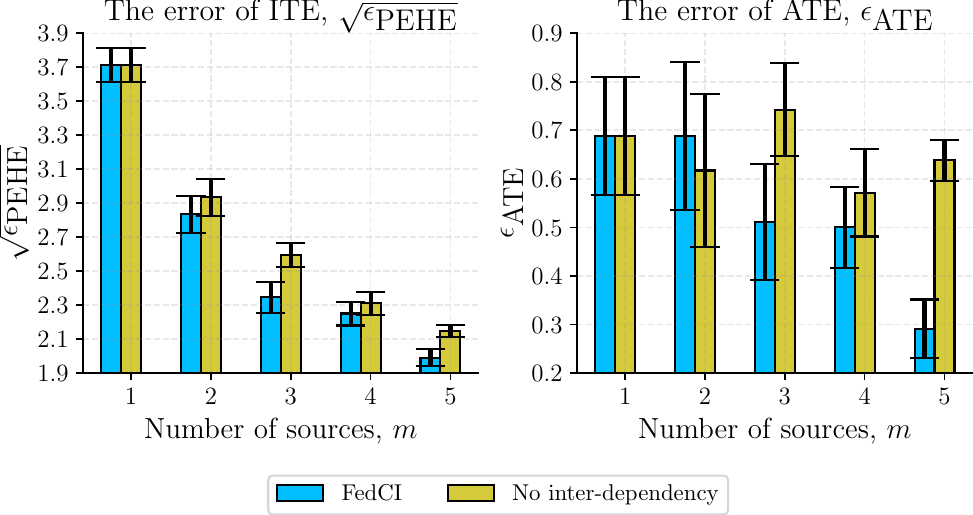}
\caption{The impact of inter-dependency on DATA-1.}
\vspace{6pt}
\label{fig:fedci-inter-dependency-alalysis}
\end{figure}
\begin{table}[!ht]
\centering
\caption{Out-of-sample errors on DATA-1 where top-3 performances are highlighted in bold (lower is better). The dashes (-) in `$\mathsf{loc}$' and `$\mathsf{agg}$' indicate that the numbers are the same as those of `$\mathsf{com}$'.}
\vspace{6pt}
\label{tab:error-synthetic}
\setlength{\tabcolsep}{12pt}
\scriptsize
\begin{tabular}{@{}lcccccc@{}}
\toprule
\multirow{2}{*}{Method}                                           & \multicolumn{3}{c}{The error of ITE ($\sqrt{\epsilon_\text{PEHE}}$)} & \multicolumn{3}{c}{The error of ATE ( $\epsilon_\text{ATE}$)} \\ \cmidrule(lr){2-4}\cmidrule(l){5-7} 
                                                                  & 1 source      & 3 sources     & 5 sources     & 1 source      & 3 sources     & 5 sources    \\ \cmidrule(r){1-1}\cmidrule(lr){2-4}\cmidrule(l){5-7}
BART$_\mathsf{loc}$                                                       & -    & 6.04$\pm$.05    & 6.02$\pm$.04    & -    & 0.59$\pm$.14    & 0.53$\pm$.10   \\
X-Learner$_\mathsf{loc}$                                                  & -    & 5.81$\pm$.13    & 5.77$\pm$.09    & -    & 0.44$\pm$.24    & 0.51$\pm$.13   \\
R-Learner$_\mathsf{loc}$                                                  & -    & 5.94$\pm$.05    & 5.94$\pm$.03    & -    & 0.65$\pm$.05    & 0.66$\pm$.02   \\
OthoRF$_\mathsf{loc}$                                                     & -   & 5.83$\pm$.12    & 6.23$\pm$.13    & -    & \textbf{0.31$\pm$.08}    & 0.52$\pm$.10   \\
TARNet$_\mathsf{loc}$    & -     & 4.25$\pm$.07     & 4.22$\pm$.06     & -     & 0.85$\pm$.04     & 0.81$\pm$.02     \\ 
CFR Wass$_\mathsf{loc}$  & -     & 4.10$\pm$.04 & 3.92$\pm$.03 & -     & 0.81$\pm$.02     & 0.80$\pm$.02     \\ 
CFR MMD$_\mathsf{loc}$   & - & 4.11$\pm$.06 & 3.93$\pm$.03 & -     & 0.80$\pm$.03     & 0.79$\pm$.02 \\
CEVAE$_\mathsf{loc}$                                                     & -   & 3.82$\pm$.09    & 3.50$\pm$.06    & -    & 0.63$\pm$.11    & 0.52$\pm$.03   \\\cmidrule(r){1-1}\cmidrule(lr){2-4}\cmidrule(l){5-7}
BART$_\mathsf{agg}$                                                 & -             & 5.97$\pm$.05              & 5.94$\pm$.03              & -             & 0.64$\pm$.14              & 0.47$\pm$.11             \\
X-Learner$_\mathsf{agg}$ & -             & 5.18$\pm$.09    & 5.09$\pm$.05    & -             & 0.46$\pm$.24    & 0.52$\pm$.13   \\
R-Learner$_\mathsf{agg}$ & -             & 5.94$\pm$.05    & 5.93$\pm$.03    & -             & 0.65$\pm$.05    & 0.66$\pm$.03   \\
OthoRF$_\mathsf{agg}$    & -             & 4.19$\pm$.13    & 3.66$\pm$.08              & -             & \textbf{0.36$\pm$.13}    & 0.48$\pm$.12             \\
TARNet$_\mathsf{agg}$    & -     & 4.02$\pm$.04     & 4.00$\pm$.05     & -     & 0.79$\pm$.04     & 0.77$\pm$.02     \\ 
CFR Wass$_\mathsf{agg}$  & -     & 3.92$\pm$.03 & 3.75$\pm$.03 & -     & 0.78$\pm$.03     & 0.76$\pm$.02     \\ 
CFR MMD$_\mathsf{agg}$   & - & 4.01$\pm$.05 & 3.80$\pm$.02 & -     & 0.78$\pm$.03     & 0.76$\pm$.02 \\
CEVAE$_\mathsf{agg}$                                                     & -    & 3.65$\pm$.10    & 2.99$\pm$.06    & -    & 0.41$\pm$.05    & 0.37$\pm$.04   \\\cmidrule(r){1-1}\cmidrule(lr){2-4}\cmidrule(l){5-7}
BART$_\mathsf{com}$                                                     & 5.98$\pm$.06             & 5.97$\pm$.06    & 5.93$\pm$.03    & 0.83$\pm$.11             & 0.56$\pm$.16    & 0.38$\pm$.09   \\
X-Learner$_\mathsf{com}$    & 5.48$\pm$.15             & 4.60$\pm$.09    & 4.15$\pm$.04    & 0.93$\pm$.22             & 0.60$\pm$.11    & \textbf{0.30$\pm$.07}   \\
R-Learner$_\mathsf{com}$     & 5.93$\pm$.06             & 5.73$\pm$.08    & 5.54$\pm$.06    & 0.78$\pm$.10             & 0.47$\pm$.09    & \textbf{0.30$\pm$.07}   \\
OthoRF$_\mathsf{com}$        & 5.86$\pm$.40             & \textbf{3.60$\pm$.12}    & \textbf{2.94$\pm$.05}    & \textbf{0.55$\pm$.14}             & 0.45$\pm$.14    & 0.34$\pm$.09   \\
TARNet$_\mathsf{com}$    & 3.93$\pm$.07     & 3.87$\pm$.05     & 3.80$\pm$.03     & 0.80$\pm$.04     & 0.77$\pm$.04     & 0.76$\pm$.02     \\ 
CFR Wass$_\mathsf{com}$  & \textbf{3.77$\pm$.05}     & 3.73$\pm$.04 & 3.71$\pm$.02 & 0.80$\pm$.04     & 0.75$\pm$.04     & 0.75$\pm$.02     \\ 
CFR MMD$_\mathsf{com}$   & 3.90$\pm$.06 & 3.73$\pm$.04 & 3.70$\pm$.02 & 0.82$\pm$.05     & 0.75$\pm$.04     & 0.75$\pm$    .02 \\
CEVAE$_\mathsf{com}$        & \textbf{3.79$\pm$.07}             & \textbf{2.85$\pm$.06}    & \textbf{2.72$\pm$.04}    & \textbf{0.51$\pm$.13}             & \textbf{0.23$\pm$.07}    & \textbf{0.20$\pm$.06}   \\\cmidrule(r){1-1}\cmidrule(lr){2-4}\cmidrule(l){5-7}
FedCI                                                             & \textbf{3.71$\pm$.10}     & \textbf{2.35$\pm$.09}    & \textbf{1.99$\pm$.05}    & \textbf{0.69$\pm$.12}    & \textbf{0.31$\pm$.12}    & \textbf{0.29$\pm$.06}   \\ \bottomrule
\end{tabular}
\end{table}

\begin{table}\centering
\caption{Out-of-sample errors on DATA-2. The dashes (-) in `$\mathsf{agg}$' indicate that the numbers are the same as those of `$\mathsf{com}$'. 
}
\vspace{6pt}
\label{tab:error-synthetic-2}
\setlength{\tabcolsep}{12pt}
\scriptsize
\begin{tabular}{@{}lcccccc@{}}
\toprule
\multirow{2}{*}{Method}                                           & \multicolumn{3}{c}{The error of ITE ($\sqrt{\epsilon_\text{PEHE}}$)} & \multicolumn{3}{c}{The error of ATE ( $\epsilon_\text{ATE}$)} \\ \cmidrule(lr){2-4}\cmidrule(l){5-7} 
                                                                  & 1 source      & 3 sources     & 5 sources     & 1 source      & 3 sources     & 5 sources    \\ \cmidrule(r){1-1}\cmidrule(lr){2-4}\cmidrule(l){5-7}
BART$_\mathsf{loc}$                                                       & -    & 18.4$\pm$0.3    & 18.3$\pm$0.2    & -    & 3.37$\pm$0.7    & 2.90$\pm$0.6   \\
X-Learner$_\mathsf{loc}$                                                  & -    & 22.7$\pm$0.5    & 22.8$\pm$0.5    & -    & 3.55$\pm$1.3    & 3.09$\pm$0.8   \\
R-Learner$_\mathsf{loc}$                                                  & -    & 26.3$\pm$0.2    & 26.1$\pm$0.2    & -    & 19.7$\pm$0.3    & 19.5$\pm$0.3   \\
OthoRF$_\mathsf{loc}$                                                     & -   & 38.3$\pm$1.4    & 40.0$\pm$0.9    & -    & 4.09$\pm$0.9    & 4.40$\pm$1.2   \\
TARNet$_\mathsf{loc}$    & -     & 37.6$\pm$0.6     & 37.1$\pm$0.4     & -     & 7.31$\pm$0.4     & 7.25$\pm$0.3     \\ 
CFR Wass$_\mathsf{loc}$  & -     & 37.2$\pm$0.7 & 37.0$\pm$0.5 & -     & 7.24$\pm$0.3     & 7.12$\pm$0.2     \\ 
CFR MMD$_\mathsf{loc}$   & - & 37.2$\pm$0.6 & 36.8$\pm$0.4 & -     & 7.21$\pm$0.4     & 7.11$\pm$0.3     \\
CEVAE$_\mathsf{loc}$                                                     & -   & $21.4\pm$0.7    & 19.8$\pm$0.6    & -    & 2.11$\pm$0.4    & 1.97$\pm$0.2   \\\cmidrule(r){1-1}\cmidrule(lr){2-4}\cmidrule(l){5-7}
BART$_\mathsf{agg}$                                                 & -             & 17.9$\pm$0.2              & 17.7$\pm$0.2              & -             & 3.91$\pm$0.8              & 3.15$\pm$0.7             \\
X-Learner$_\mathsf{agg}$ & -             & 18.2$\pm$0.4    & 17.1$\pm$0.2    & -             & 3.43$\pm$1.3    & 3.07$\pm$0.8   \\
R-Learner$_\mathsf{agg}$ & -             & 26.2$\pm$0.3    & 26.1$\pm$0.2    & -             & 19.7$\pm$0.4    & 19.6$\pm$0.3   \\
OthoRF$_\mathsf{agg}$    & -             & 25.0$\pm$1.3    & 17.3$\pm$0.6              & -             & 4.56$\pm$1.1    & \textbf{1.30$\pm$0.4}             \\
TARNet$_\mathsf{agg}$    & -     & 36.5$\pm$0.3     & 36.1$\pm$0.3     & -     & 7.26$\pm$0.3     & 7.18$\pm$0.3     \\ 
CFR Wass$_\mathsf{agg}$  & -     & 35.2$\pm$0.5 & 35.0$\pm$0.3 & -     & 7.13$\pm$0.3     & 6.97$\pm$0.2     \\ 
CFR MMD$_\mathsf{agg}$   & - & 35.2$\pm$0.5 & 35.1$\pm$0.4 & -     & 7.10$\pm$0.4     & 7.05$\pm$0.2     \\
CEVAE$_\mathsf{agg}$                                                     & -    & $19.2\pm$0.8    & 18.3$\pm$0.7    & -    & 2.02$\pm$0.3    & 1.91$\pm$0.4   \\\cmidrule(r){1-1}\cmidrule(lr){2-4}\cmidrule(l){5-7}
BART$_\mathsf{com}$                                                     & \textbf{18.0$\pm$0.4}             & \textbf{17.7$\pm$0.2}    & 17.4$\pm$0.1    & \textbf{3.54$\pm$1.3}             & 2.94$\pm$0.8    & \textbf{1.84$\pm$0.5}   \\
X-Learner$_\mathsf{com}$    & 21.1$\pm$0.9             & 17.9$\pm$0.4    & \textbf{16.2$\pm$0.2}    & 4.55$\pm$1.4             & 3.29$\pm$1.0    & 2.37$\pm$0.8   \\
R-Learner$_\mathsf{com}$     & 25.9$\pm$0.6             & 23.5$\pm$0.5    & 21.3$\pm$0.4    & 19.0$\pm$0.8             & 15.6$\pm$0.7    & 12.3$\pm$0.6   \\
OthoRF$_\mathsf{com}$        & 37.8$\pm$2.7             & \textbf{10.7$\pm$0.5}    & \textbf{9.83$\pm$0.5}    & 7.88$\pm$2.2             & \textbf{1.99$\pm$0.4}    & 2.36$\pm$0.6   \\
TARNet$_\mathsf{com}$    & 36.1$\pm$0.4     & 35.5$\pm$0.2     & 35.0$\pm$0.2     & 7.11$\pm$0.4     & 7.10$\pm$0.3     & 7.08$\pm$0.2     \\ 
CFR Wass$_\mathsf{com}$  & 35.1$\pm$0.4     & 34.5$\pm$0.2 & 34.1$\pm$0.2 & 7.10$\pm$0.4     & 7.01$\pm$0.3     & 6.90$\pm$0.2     \\ 
CFR MMD$_\mathsf{com}$   & 35.1$\pm$0.4 & 35.0$\pm$0.2 & 34.9$\pm$0.2 & 7.12$\pm$0.4     & 7.02$\pm$0.3     & 7.01$\pm$0.2     \\
CEVAE$_\mathsf{com}$        & \textbf{20.1$\pm$0.5}             & 18.4$\pm$0.6    & 16.6$\pm$0.6    & \textbf{1.50$\pm$0.3}             & \textbf{1.38$\pm$0.4}    & 1.89$\pm$0.2   \\\cmidrule(r){1-1}\cmidrule(lr){2-4}\cmidrule(l){5-7}
FedCI                                                             & \textbf{9.28$\pm$0.4}     & \textbf{6.34$\pm$0.2}    & \textbf{5.53$\pm$0.1}    & \textbf{2.37$\pm$0.5}    & \textbf{1.47$\pm$0.4}    & \textbf{0.74$\pm$.2}   \\ \bottomrule
\end{tabular}
\end{table}

\noindent\emph{\underline{FedCI vs. training on combined data.}} Figure~\ref{fig:fedci-alalysis} reports the three evaluation metrics of FedCI compared with two data source settings: training on combined data and training locally on each data source. As expected, the figures show that the errors of FedCI are as low as those of training on the combined data. This result verifies the efficacy of the proposed federated algorithm.

\noindent\emph{\underline{Inter-dependency component analysis.}} We study the impact of the inter-dependency component (see Section~\ref{sec:model}) by removing it from the model. Figure~\ref{fig:fedci-inter-dependency-alalysis} presents the errors of FedCI compared with `no inter-dependency' (FedCI without inter-dependency). The figures show that the errors in predicting ITE and ATE of `no inter-dependency' seem to be higher than those of FedCI. This result showcases the importance of our proposed inter-dependency component.

In Figure~\ref{fig:fedci-alalysis}, the error $\epsilon_\text{ATE}$ of FedCI increases as the number of sources increases from 1 to 2. In Figure~\ref{fig:fedci-inter-dependency-alalysis}, $\epsilon_\text{ATE}$ of FedCI is larger than that of without inter-dependency. These results might be due to the non-convex optimisation which could lead to a local minima. A potential direction to improve is to use a minibatch stochastic gradient descent for GPs \citep{chen2020stochastic}.

\noindent\emph{\underline{Contrasting with existing baselines.}} In this experiment, we compare FedCI with the existing causal inference methods. All these baseline methods do not consider estimating causal effects on multiple sources. Thus, we train them in three cases as explained earlier: \textbf{(1)} train locally ($\mathsf{loc}$), \textbf{(2)} train with combined data ($\mathsf{com}$), and \textbf{(3)} train with bootstrap aggregating ($\mathsf{agg}$). Note that case (\textbf{2}) violates constraint that data are stored at their local sites. We expect that the error of FedCI to be close to case \textbf{(2)} of the baselines. 
Table~\ref{tab:error-synthetic}~and~\ref{tab:error-synthetic-2} report the  performance of each method in estimating ATE and ITE. Regardless of different scales on the two synthetic datasets, the figure shows that FedCI achieves competitive results as compared with all the baselines. FedCI is in the top-3 performances among all the methods. Importantly, FedCI obtains lower errors than those of BART$_\mathsf{com}$, X-Learner$_\mathsf{com}$, R-Learner$_\mathsf{com}$, OthoRF$_\mathsf{com}$, TARNet$_\mathsf{com}$, CFR~Wass$_\mathsf{com}$, and CFR~MMD$_\mathsf{com}$, which were trained on combined data and thus violate constraint of federated data setting. 
Compared with CEVAE$_\mathsf{com}$, FedCI is better than this method in predicting ITE and comparable with this method in predicting ATE (slightly higher errors). However, we emphasize again that this result is expected since FedCI is a federated learning algorithm while CEVAE$_\mathsf{com}$ works directly on combined data.

\begin{figure}
\centering
    \includegraphics[width=0.65\textwidth]{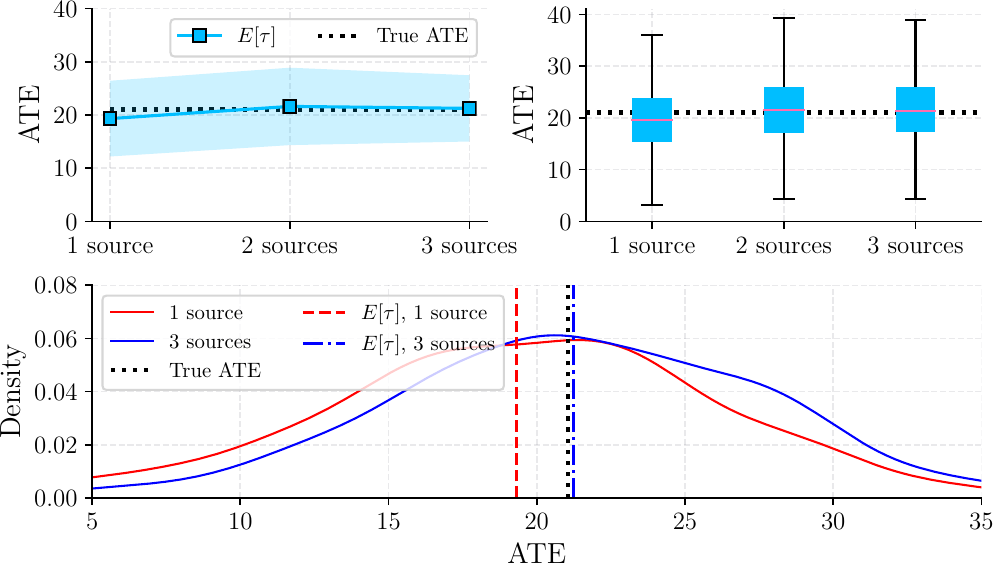}
\caption{
       Estimated distribution of ATE on source \#1 of DATA-2. The dotted black lines represent the true ATE.
    } \label{fig:fedci-uncertainty-analysis-synthetic}
\end{figure}

\noindent\emph{\underline{The estimated distribution of ATE.}} To analyse uncertainty, we present in  Figure~\ref{fig:fedci-uncertainty-analysis-synthetic} the estimated distribution of ATE in the first source ($\mathsf{s}=1$). The figures show that the true ATE is covered by the estimated interval and the estimated mean ATE shifts towards its true value (dotted lines) when more data sources are used. This result might provide useful information about the application in practice. 

\noindent\textbf{Misspecification Analysis: Binary and Count Outcomes} 

\noindent\emph{\underline{Data.}} In this experiment, we analyse the performance when the model is misspecified. We compare FedCI with the baselines in two cases: binary outcomes and count outcomes. We reuse the ground truth distributions of $x_{ij}$ and $w_i$ as in analyses on real-value outcomes. For the outcomes, we simulate them with the following distributions:
\begin{align*}
    &\text{Binary outcomes:} &&y_i(0) \sim \mathsf{Bern}(\varphi(b_0 + x_i^\top \mathbf{b}_1)),\\
   &&& y_i(1) \sim \mathsf{Bern}(\varphi(c_0 + x_i^\top \mathbf{c}_1)).\\
    &\text{Count outcomes:} &&y_i(0) \sim \mathsf{Poisson}(\exp(b_0 + x_i^\top \mathbf{b}_1)),\\
    &&&y_i(1) \sim \mathsf{Poisson}(\exp(c_0 + x_i^\top \mathbf{c}_1)).
\end{align*}
\emph{\underline{Results and discussion.}}  
From Table~\ref{tab:error-synthetic-binary}~and~\ref{tab:error-synthetic-count}, FedCI gives competitive results compared with the baselines trained on combined data. The reason for the good performance for FedCI and some baselines in these misspecification cases is because they provide good estimates for the mean of the missing outcomes. This might in turn be due to the mean estimation of Gaussian distribution in FedCI coincides with the mean estimation of the other distributions.  
Nevertheless, since these are misspecified cases, the continuous posterior distribution is not a good estimation. To obtain better posterior distributions of the missing outcomes and the causal estimands, we would need to consider some other appropriate distributions in our model.

\begin{table}
\centering
\caption{Out-of-sample errors on binary outcomes data.}
\vspace{6pt}
\label{tab:error-synthetic-binary}
\setlength{\tabcolsep}{12pt}
\scriptsize
\begin{tabular}{@{}lcccccc@{}}
\toprule
\multirow{2}{*}{Method}                                           & \multicolumn{3}{c}{The error of ITE ($\sqrt{\epsilon_\text{PEHE}}$)} & \multicolumn{3}{c}{The error of ATE ( $\epsilon_\text{ATE}$)} \\ \cmidrule(l){2-4}\cmidrule(l){5-7} 
           & 1 source       & 3 sources      & 5 sources      & 1 source       & 3 sources      & 5 sources      \\ \cmidrule(r){1-1}\cmidrule(lr){2-4}\cmidrule(l){5-7}
BART$_\mathsf{com}$      & 0.77$\pm$.01     & 0.73$\pm$.01     & 0.70$\pm$.01     & 0.41$\pm$.01     & 0.31$\pm$.01     & 0.24$\pm$.01     \\ 
X-Learner$_\mathsf{com}$ & 0.69$\pm$.01     & 0.60$\pm$.01     & 0.56$\pm$.01     & \textbf{0.13$\pm$.03}     & 0.10$\pm$.02     & \textbf{0.09$\pm$.01}    \\ 
R-Learner$_\mathsf{com}$ & 0.65$\pm$.01     & 0.64$\pm$.01     & 0.62$\pm$.01     & \textbf{0.05$\pm$.01} & \textbf{0.03$\pm$.01} & \textbf{0.03$\pm$.01} \\ 
OthoRF$_\mathsf{com}$    & 0.94$\pm$.04     & 0.60$\pm$.01     & 0.56$\pm$.01     & 0.17$\pm$.03     & 0.18$\pm$.03     & 0.16$\pm$.03     \\ 
TARNet$_\mathsf{com}$    & 0.68$\pm$.02     & 0.68$\pm$.02     & 0.65$\pm$.01     & 0.33$\pm$.01     & 0.33$\pm$.01     & 0.32$\pm$.01     \\ 
CFR Wass$_\mathsf{com}$  & 0.61$\pm$.02     & \textbf{0.50$\pm$.01} & \textbf{0.50$\pm$.01} & 0.32$\pm$.01     & 0.30$\pm$.01     & 0.30$\pm$.01     \\ 
CFR MMD$_\mathsf{com}$   & \textbf{0.55$\pm$.01} & \textbf{0.50$\pm$.01} & \textbf{0.50$\pm$.01} & 0.32$\pm$.01     & 0.30$\pm$.01     & 0.30$\pm$.01     \\ 
CEVAE$_\mathsf{com}$     & \textbf{0.39$\pm$.01} & \textbf{0.37$\pm$.01} & \textbf{0.37$\pm$.01} & \textbf{0.08$\pm$.02} & \textbf{0.05$\pm$.01} & \textbf{0.05$\pm$.01} \\ \cmidrule(r){1-1}\cmidrule(lr){2-4}\cmidrule(l){5-7}
FedCI                           & \textbf{0.41$\pm$.01}                      & \textbf{0.40$\pm$.01}                      & \textbf{0.39$\pm$.01}                      & \textbf{0.05$\pm$.01}                      & \textbf{0.04$\pm$.01}                      & \textbf{0.03$\pm$.01}                      \\ \bottomrule
\end{tabular}
\end{table}

\begin{table}
\centering
\caption{Out-of-sample errors on count outcomes data.
}
\vspace{6pt}
\label{tab:error-synthetic-count}
\setlength{\tabcolsep}{12pt}
\scriptsize
\begin{tabular}{@{}lcccccc@{}}
\toprule
\multirow{2}{*}{Method}                                           & \multicolumn{3}{c}{The error of ITE ($\sqrt{\epsilon_\text{PEHE}}$)} & \multicolumn{3}{c}{The error of ATE ( $\epsilon_\text{ATE}$)} \\ \cmidrule(lr){2-4}\cmidrule(l){5-7} 
           & 1 source       & 3 sources      & 5 sources      & 1 source       & 3 sources      & 5 sources      \\ \cmidrule(r){1-1}\cmidrule(lr){2-4}\cmidrule(l){5-7}
BART$_\mathsf{com}$      & 6.30$\pm$.06     & 6.29$\pm$.04     & 6.26$\pm$.03     & 0.75$\pm$.14     & 0.59$\pm$.18     & 0.47$\pm$.13     \\ 
X-Learner$_\mathsf{com}$ & 6.10$\pm$.10     & 5.16$\pm$.06     & 4.72$\pm$.03     & 1.34$\pm$.29     & 0.63$\pm$.12     & 0.42$\pm$.08    \\ 
R-Learner$_\mathsf{com}$ & 6.27$\pm$.06     & 6.09$\pm$.05     & 5.89$\pm$.04     & 0.82$\pm$.13 & 0.66$\pm$.15 & 0.56$\pm$.10 \\ 
OthoRF$_\mathsf{com}$    & 6.02$\pm$.29     & 4.15$\pm$.06     & \textbf{3.74$\pm$.05}     & 0.75$\pm$.18     & 0.54$\pm$.17     & \textbf{0.41$\pm$.10}     \\ 
TARNet$_\mathsf{com}$    & 4.54$\pm$.14     & \textbf{3.98$\pm$.05}     & 3.80$\pm$.02     & 0.77$\pm$.10     & 0.66$\pm$.02     & 0.62$\pm$.03     \\ 
CFR Wass$_\mathsf{com}$  & \textbf{4.08$\pm$.04}     & 4.03$\pm$.03 & 3.78$\pm$.02 & 0.72$\pm$.04     & \textbf{0.51$\pm$.03}     & 0.50$\pm$.03     \\ 
CFR MMD$_\mathsf{com}$   & 4.15$\pm$.06 & 4.05$\pm$.04 & 3.77$\pm$.02 & \textbf{0.69$\pm$.07}     & 0.54$\pm$.03     & 0.50$\pm$.03     \\ 
CEVAE$_\mathsf{com}$     & \textbf{3.40$\pm$.09} & \textbf{3.31$\pm$.07} & \textbf{3.08$\pm$.05} & \textbf{0.56$\pm$.16} & \textbf{0.40$\pm$.12} & \textbf{0.35$\pm$.08} \\ \cmidrule(r){1-1}\cmidrule(lr){2-4}\cmidrule(l){5-7}
FedCI                           & \textbf{4.02$\pm$.10}                      & \textbf{3.05$\pm$.08}                      & \textbf{2.66$\pm$.04}                      & \textbf{0.54$\pm$.09}                      & \textbf{0.48$\pm$.08}                      & \textbf{0.25$\pm$.05}                      \\ \bottomrule
\end{tabular}
\end{table}

\subsubsection{IHDP Data}
\label{sec:ihdp}
\noindent\emph{\underline{Data.}} The Infant Health and Development Program (IHDP) \citep{hill2011bayesian} is a dataset with 747 data points, each has 25 covariates. These data are obtained from a randomized study on the impact of specialist visits to children's cognitive development. Herein, specialist visit is the treatment and children's cognitive development is the outcome. We use the NPCI package \citep{dorie2016npci} to simulate two potential outcomes for the treatment (with or without specialist visit) of each child. Hence, the \textit{true} individual treatment effect can be computed for evaluation purposes. 
There are 10 replicates of the dataset, and each of them is  divided into three sources of size 249. For each source, we then split it into three equal sets for the purpose of training, testing, and validating the models. The mean and standard error of the aforementioned evaluation metrics are reported over the above 10 replicates of the data. 

\noindent\emph{\underline{Results and discussion.}}  Similar to the experiment for synthetic datasets,  here we also train the baselines in three cases as explained earlier. We also expect that the errors of FedCI to be close to the baselines trained with combined data ($\mathsf{com}$). The results reported in Table~\ref{tab:error-ihdp} show that FedCI achieves competitive results compared to the baselines. Indeed, FedCI is in the top-3 performances among all the methods.
The reason is because FedCI has access to all the data sources in a federated fashion while the `baselines trained locally' ($\mathsf{loc}$) and the `baselines trained with bootstrap aggregating' ($\mathsf{agg}$) only have access to a local data source. 
This result again verifies that FedCI can be used to estimate causal effects effectively under privacy-perserving, federated data settings.

\begin{figure}\centering
    \includegraphics[width=0.7\textwidth]{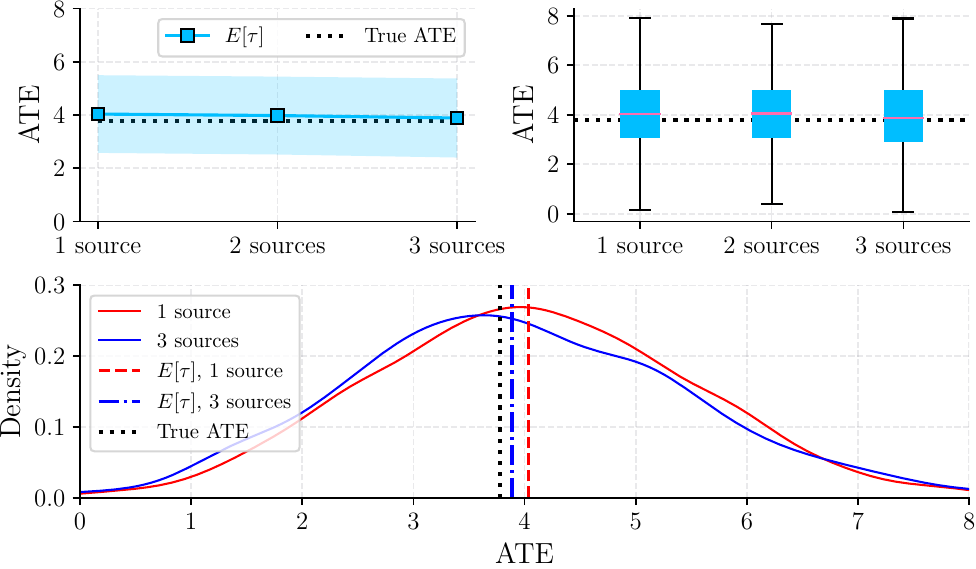}
\caption{The estimated ATE distribution on source \#1 of IHDP dataset. The dotted black lines represent the true ATE.}
\label{fig:fedci-uncertainty-analysis-ihdp}
\end{figure}

Similar to the experiment on synthetic data, the estimated distribution of ATE in the first source ($\mathsf{s}=1$) is presented in~Figure~\ref{fig:fedci-uncertainty-analysis-ihdp}. Again, the figures show that the true ATE is inside the estimated interval and the estimated mean ATE shifts towards its true value (dotted lines) when more data sources are used.

\begin{table}\centering
\caption{Out-of-sample errors on IHDP dataset. The dashes (-) in `$\mathsf{agg}$' indicate that the numbers are the same as those of `$\mathsf{com}$'. 
}
\vspace{6pt}
\label{tab:error-ihdp}
\setlength{\tabcolsep}{12pt}
\scriptsize
\begin{tabular}{@{}lcccccc@{}}
\toprule
\multirow{2}{*}{Method}                                           & \multicolumn{3}{c}{The error of ITE ($\sqrt{\epsilon_\text{PEHE}}$)} & \multicolumn{3}{c}{The error of ATE ( $\epsilon_\text{ATE}$)} \\ \cmidrule(lr){2-4}\cmidrule(l){5-7} 
                                                                  & 1 source      & 2 sources     & 3 sources     & 1 source      & 2 sources     & 3 sources    \\ \cmidrule(r){1-1}\cmidrule(lr){2-4}\cmidrule(l){5-7}
BART$_\mathsf{loc}$                                                       & -    & 5.83$\pm$2.6    & 6.56$\pm$3.3    & -    & 2.09$\pm$0.9    & 1.38$\pm$0.5   \\
X-Learner$_\mathsf{loc}$                                                  & -    & 4.14$\pm$1.5    & 4.54$\pm$1.9    & -    & 1.51$\pm$0.7    & 0.77$\pm$0.5   \\
R-Learner$_\mathsf{loc}$                                                  & -    & 6.35$\pm$1.9    & 6.16$\pm$2.0    & -    & 2.13$\pm$0.7    & 1.44$\pm$0.3   \\
OthoRF$_\mathsf{loc}$                                                     & -    & 4.33$\pm$1.6    & 4.59$\pm$1.9    & -    & 1.10$\pm$0.6    & 0.75$\pm$0.3   \\
TARNet$_\mathsf{loc}$    & -     & 3.71$\pm$1.0     & 3.83$\pm$1.1     & -     & 1.31$\pm$0.5     & 0.98$\pm$0.4     \\ 
CFR Wass$_\mathsf{loc}$  & -     & 3.35$\pm$0.8 & 3.12$\pm$0.7 & -     & 0.87$\pm$0.5     & 0.82$\pm$0.4     \\ 
CFR MMD$_\mathsf{loc}$   & - & 3.40$\pm$0.9 & 3.15$\pm$1.2 & -     & 1.17$\pm$0.5     & 0.63$\pm$0.3     \\
CEVAE$_\mathsf{loc}$                                                     & -    & 3.78$\pm$0.7    & 3.93$\pm$0.8    & -    & 1.91$\pm$0.3    & 2.37$\pm$0.2   \\\cmidrule(r){1-1}\cmidrule(lr){2-4}\cmidrule(l){5-7}
BART$_\mathsf{agg}$                                                 & -             & 4.05$\pm$1.9              & 3.69$\pm$1.8              & -             & 2.09$\pm$1.0              & 1.30$\pm$0.5             \\
X-Learner$_\mathsf{agg}$ & -             & 3.98$\pm$1.5    & 4.28$\pm$1.9    & -             & 1.51$\pm$0.7    & 0.83$\pm$0.5   \\
R-Learner$_\mathsf{agg}$ & -             & 4.76$\pm$1.3    & 4.46$\pm$1.6    & -             & 1.92$\pm$0.5    & 1.41$\pm$0.2   \\
OthoRF$_\mathsf{agg}$    & -             & 3.40$\pm$1.1    & 4.26$\pm$1.9              & -             & 0.87$\pm$0.3    & 1.20$\pm$0.6             \\
TARNet$_\mathsf{agg}$    & -     & 3.52$\pm$0.9     & 3.81$\pm$1.2     & -     & 1.23$\pm$0.4     & 0.95$\pm$0.4     \\ 
CFR Wass$_\mathsf{agg}$  & -     & 3.21$\pm$0.7 & 2.93$\pm$0.9 & -     & 0.80$\pm$0.3     & 0.71$\pm$0.2     \\ 
CFR MMD$_\mathsf{agg}$   & - & 3.17$\pm$0.8 & 2.91$\pm$1.3 & -     & 1.12$\pm$0.5     & 0.57$\pm$0.3     \\
CEVAE$_\mathsf{agg}$    & -             & 3.63$\pm$0.7    & 3.73$\pm$0.5              & -             & 0.92$\pm$0.2    & 0.84$\pm$0.5             \\\cmidrule(r){1-1}\cmidrule(lr){2-4}\cmidrule(l){5-7}
BART$_\mathsf{com}$                                                 & 5.98$\pm$2.7             & 4.32$\pm$2.1              & 4.04$\pm$2.0              & 1.80$\pm$1.1             & 2.09$\pm$1.1              & 1.21$\pm$0.6             \\
X-Learner$_\mathsf{com}$ & 4.22$\pm$1.6             & 4.15$\pm$1.5    & 4.06$\pm$1.8    & 1.64$\pm$0.7             & 1.93$\pm$0.8    & 0.84$\pm$0.4   \\
R-Learner$_\mathsf{com}$ & 6.97$\pm$2.1             & 4.43$\pm$1.4    & 4.47$\pm$1.7    & 3.15$\pm$0.5             & 1.34$\pm$0.5    & 1.10$\pm$0.3   \\
OthoRF$_\mathsf{com}$    & 4.49$\pm$1.9             & 3.81$\pm$1.3    & 3.75$\pm$1.5              & 1.86$\pm$0.8             & 1.61$\pm$0.6    & 1.56$\pm$0.8\\ 
TARNet$_\mathsf{com}$    & 4.50$\pm$1.4     & 3.15$\pm$0.8     & 3.79$\pm$1.1     & \textbf{1.52$\pm$0.5}     & 1.18$\pm$0.4     & 0.91$\pm$0.3     \\ 
CFR Wass$_\mathsf{com}$  & \textbf{4.37$\pm$1.2}     & 2.93$\pm$0.6 & 2.85$\pm$0.9 & \textbf{1.18$\pm$0.7}     & \textbf{0.72$\pm$0.2}     & 0.67$\pm$0.1     \\ 
CFR MMD$_\mathsf{com}$   & 4.43$\pm$1.3 & \textbf{2.85$\pm$0.6} & \textbf{2.83$\pm$1.1} & 2.32$\pm$0.8     & \textbf{0.63$\pm$0.2}     & \textbf{0.54$\pm$0.2}     \\
CEVAE$_\mathsf{com}$        & \textbf{3.16$\pm$0.6}             & \textbf{2.34$\pm$0.6}    & \textbf{2.31$\pm$0.7}    & 2.02$\pm$0.4             & \textbf{0.53$\pm$0.1}    & \textbf{0.48$\pm$0.2}   \\\cmidrule(r){1-1}\cmidrule(lr){2-4}\cmidrule(l){5-7}
FedCI                                                             & \textbf{2.88$\pm$0.8}     & \textbf{2.36$\pm$0.5}    & \textbf{2.35$\pm$0.6}    & \textbf{1.43$\pm$0.7}    & 1.03$\pm$0.4    & \textbf{0.51$\pm$0.2}   \\ \bottomrule
\end{tabular}
\end{table}

\subsection{summary}
\label{sec:conclusion-fedci}
We have introduced FedCI, a Bayesian causal inference paradigm via a reformulation of multi-output GPs to learn causal effects, while keeping data at their local sites. An inference method involving the decomposition of ELBO is presented, allowing the model to be trained in a federated setting. Limitations of FedCI are: (1) it assumes data sources have the same data distribution, (2) the uncertainty of the causal estimands only depends on number of data points in the local source, and (3) there are no missing data among the sources. In the following sections, we introduce two other instances of the proposed federated framework, CausalRFF and CausalFI, that address these challenges.

\section{An Adaptive Federated Inference Algorithm}
\label{sec:causalrff}
Althought FedCI presented in Section~\ref{sec:fedci} can estimate causal effects from multiple data sources without combining or sharing raw data, it assumes the same data distribution among different sources. In this section, we introduce CausalRFF that adaptively transfer knowledge among different sources and allowing different data distribution among the sources, hence relax Assumption~\ref{assumption:homogeneous-heterogeneous} of FedCI.

\begin{figure}
    \centering
\includegraphics[width=0.25\textwidth]{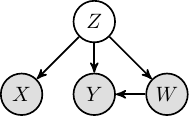}
\caption{The causal graph with latent confounder $Z$, treatment $W$, outcome $Y$, covariate/proxy variable $X$.}
        \label{fig:the-model-causal-graph}
\end{figure}

\noindent \textbf{Problem setting \& notations.}
We consider the problem setting defined in Section~\ref{sec:prob-form} with $m$ data sources $\mathsf{D}^\mathsf{s}$, where $\mathsf{s} = 1,2,\!...,m$. 
The distributions of these data sources might be completely different. All the sources share the same causal graph as shown in Figure~\ref{fig:the-model-causal-graph}, but the data distributions may be different, e.g., $\p_{\mathsf{s}_1}(x, w, y) \neq \p_{\mathsf{s}_2}(x, w, y)$, where $\p_{\mathsf{s}_1}(\cdot)$ and $\p_{\mathsf{s}_2}(\cdot)$ denote the two distributions on two sources $\mathsf{s}_1$ and $\mathsf{s}_2$, respectively. Similarly, the marginal and the conditional distributions with respect to these variables can also be different (or similar). The objective is to develop a global causal inference model that satisfies \emph{both} of the following two conditions: \textbf{(i)} the causal inference model can be trained in a private setting where the data of each source are not shared to an outsider, and \textbf{(ii)} the causal inference model can incorporate data from multiple sources to improve causal effects estimation in each specific source.

\noindent\textbf{Causal effects of interest.} We are interested in estimating the average treatment effect (ATE) and conditional average treatment effect (CATE) as defined in Eq.~(\ref{eq:ate-pearl})~and~(\ref{eq:cate-pearl}), respectively. 
Given a set of $n$ \emph{new} individuals whose covariates/observed proxy variables are $\{x_i\}_{i=1}^{n}$, the CATE and ATE in this sub-population are obtained by $\texttt{cate}(x_i)$ and $\uptau = \sum_{i=1}^{n}\texttt{cate}(x_i)/n$.

The central task to estimate CATE and ATE is to find $\e[Y\,|\, \doo(W=w),X=x]$. Since the data distribution of each source might be different from (or similar to) each other, we use the notation $\e[Y|\doo(W=w^\mathsf{s}, X=x^\mathsf{s}]$ to denote the expectation of the outcome $Y$ under an intervention on $W$ of an individual in source $\mathsf{s}$. With the existence of the latent confounder $Z$, we can further expand this quantity using $do$-calculus \citep{pearl1995causal}. In particular, from the backdoor adjustment formula, we have \begin{align}
&\e\big[Y|\doo(W=w^\mathsf{s}), X=x^\mathsf{s}\big] =  \int\e\big[Y|W=w^\mathsf{s},Z=z^\mathsf{s}\big]\p_\mathsf{s}(z^\mathsf{s}|x^\mathsf{s})dz^\mathsf{s}.\label{eq:est-causal-eff}\end{align}
Eq.~(\ref{eq:est-causal-eff}) shows that the causal effect is identifiable if we can find the conditional distributions $\p_\mathsf{s}(y^\mathsf{s}|w^\mathsf{s},z^\mathsf{s})$ and $\p_\mathsf{s}(z^\mathsf{s}|x^\mathsf{s})$ for each source $\mathsf{s}$. The second distribution can be further expanded by $\p_\mathsf{s}(z^\mathsf{s}|x^\mathsf{s}) = \sum_{w^\mathsf{s}}\int p_\mathsf{s}(z|x^\mathsf{s}, y_i^\mathsf{s},w^\mathsf{s})  \p_\mathsf{s}(y^\mathsf{s}|x^\mathsf{s},w^\mathsf{s}) \p_\mathsf{s}(w^\mathsf{s}|x^\mathsf{s}) \mathrm{d}y^\mathsf{s}$. 
Following the forward sampling strategy, the remaining is to find the following distributions
\begin{align}
    &\p_\mathsf{s}(w^\mathsf{s}|x^\mathsf{s}), &\p_\mathsf{s}(y^\mathsf{s}|x^\mathsf{s},w^\mathsf{s}), &&\p_\mathsf{s}(z^\mathsf{s}|x^\mathsf{s}, y^\mathsf{s},w^\mathsf{s}), &&\p_\mathsf{s}(y^\mathsf{s}|w^\mathsf{s},z^\mathsf{s}),\label{eq:target-est}
\end{align}
and then systematically draw samples from these estimated distributions to obtain the empirical expectation of $Y$ given $\doo(W=w^\mathsf{s})$ and $ X=x^\mathsf{s}$.

\noindent\textbf{Identification.} The CATE and ATE are identifiable if we are able to learn the distributions in Eq.~(\ref{eq:target-est}), which involve latent confounder $Z$. \citet{louizos2017causal} showed that this is possible if $Z$ has a relationship to the observed variables $X$, and there are many cases that it is identifiable such as: $Z$ is categorical and $X$ is a Gaussian mixture model \citep{anandkumar2014tensor}, $X$ includes three independent views of $Z$ \citep{goodman1974exploratory,allman2009identifiability,anandkumar2012method},  $Z$ is a multivariate binary and $X$ are
noisy functions of $Z$ \citep{jernite2013discovering,arora2017provable}, to name a few. Following the works by \citet{louizos2017causal,madras2019fairness}, we use variational inference in the spirit of the variational auto-encoder (VAE) to recover the latent confounders, since it can learn a rich class of latent-variable models, and thus recovering the causal effects. 
Identification of our work follows closely from the literature, however our main contribution is in the \emph{federated setting} of the model. Please refer to Appendix~\ref{ap:proof-iden} 
for the proof of identifiability.

\subsection{The Structural Equations}
\label{sec:struct-eqs}
This section presents how the causal relations are modeled. Since $Z$ is the root node in the causal graph, we model it as a multivariate normal distribution: $Z \sim \mathsf{N}(\bm{\mu},\sigma_z^2\mathbf{I}_{d_z})$ for the all sources. 
We now detail the structural equations of $Y$, $W$ and $X$. Let $V$ be a univariate variable that represents a node or a dimension of a node in the causal graph (Figure~\ref{fig:the-model-causal-graph}), i.e., $V$ can be $Y$, $W$ or a dimension of $X$. Let $\mathsf{pa}(V)$ be set of $V$'s parent variables in the causal graph, i.e, the nodes with directed edges to $V$. We model the structural equation of $V$ in two cases as follows:
\begin{align}
&\text{if $V$ is continuous: } &V &= f_v(\mathsf{pa}(V)) + \epsilon_v,\\
    &\text{if $V$ is binary: } &V &= \mathds{1}[\varphi(f_v(\mathsf{pa}(V)))> \epsilon_v],
\end{align}
where $\epsilon_v \sim \mathsf{N}(0,\sigma_v^2)$ for the former case and $\epsilon_v \sim \mathsf{U}[0,1]$ for the latter case, $\varphi(\cdot)$ is the logistic function and $\mathds{1}(\cdot)$ is the indicator function. The latter case implies that $V$ given $\mathsf{pa}(V)$ follows Bernoulli distribution with $\p(V=1|\mathsf{pa}(V)) = \varphi(f_v(\mathsf{pa}(V)))$. Furthermore, if $W \in \mathsf{pa}(V)$, then we further model
\begin{align}
    f_v(\mathsf{pa}(V)) &\,\,=\,\, (1-W)f_{v0}(\mathsf{pa}(V)\setminus\{W\})  \,\,+ \,\, Wf_{v1}(\mathsf{pa}(V)\setminus\{W\}).
\end{align}
\textbf{Example.} If $Y\in \mathbb{R}$, $W\in\{0,1\}$ and $X_k \in \mathbb{R}$ ($X_k$ is the $k$--th dimension of $X$), then the structural equations are as follows:
\begin{align*}
    Y &= (1-W)f_{y0}(Z) + Wf_{y1}(Z) + \epsilon_y, \qquad W = \mathds{1}[\varphi(f_w(Z)) > \epsilon_w], \qquad X_k = f_{x_k}(Z) + \epsilon_{X_k}.
\end{align*}
In the subsequent sections, we present how to learn the functions $f_v$ ($v\in \{y0,y1,w,x\})$ in a federated setting and then use them to estimate the causal effects of interest.

\subsection{Learning Distributions Involving Latent Confounder}
\label{sec:latent-dists}

To estimate causal effects, we need to estimate the four quantities detailed in Eq.~(\ref{eq:target-est}). This section presents how to learn $\p_\mathsf{s}(z^\mathsf{s}|x^\mathsf{s}, y^\mathsf{s},w^\mathsf{s})$ and $\p_\mathsf{s}(y^\mathsf{s}|w^\mathsf{s},z^\mathsf{s})$. Since the marginal likelihood has no analytical form, we learn the above distributions using variational inference which maximizes the evidence lower bound (ELBO)
\begin{align}
    \mathcal{L} &= \sum_{\mathsf{s}\in\bm{\mathcal{S}}}\sum_{i=1}^{n_\mathsf{s}}\Big(\e_q\big[\log\p_\mathsf{s}(y_i^\mathsf{s}|w_i^\mathsf{s},z_i^\mathsf{s}) + \log\p_\mathsf{s}(w_i^\mathsf{s}|z_i^\mathsf{s}) + \log\p_\mathsf{s}(x_i^\mathsf{s}|z_i^\mathsf{s})\big] - \text{KL}[\q(z_i^\mathsf{s})\|\p(z_i^\mathsf{s})]\Big),
\end{align}
where $\q(z^\mathsf{s}) = \mathsf{N}(z^\mathsf{s}; f_q(y^\mathsf{s},w^\mathsf{s},x^\mathsf{s}),\sigma_q^2\mathbf{I})$ is the variational posterior distribution. The function $f_q(\cdot)$ is modeled as follows: $f_q(y^\mathsf{s},w^\mathsf{s},x^\mathsf{s}) = (1-w^\mathsf{s})f_{q0}(y^\mathsf{s},x^\mathsf{s}) + w^\mathsf{s}f_{q1}(y^\mathsf{s},x^\mathsf{s})$, where $f_{q0}$ and $f_{q1}$ are two functions to be learned. 
The density functions $\p_\mathsf{s}(y^\mathsf{s}|w^\mathsf{s},z^\mathsf{s})$,  $\p_\mathsf{s}(w^\mathsf{s}|z^\mathsf{s})$ and $\p_\mathsf{s}(x^\mathsf{s}|z^\mathsf{s})$ are obtained from the structural equations as described in Section~\ref{sec:struct-eqs}.  
 Please refer to Appendix~\ref{ap:derivation-loss} 
 for details on derivation of the ELBO.

\noindent\textbf{Adaptive modeling.} Since the observed data from each source might come from different (or similar) distributions, we would model them separately and adaptively learn their similarities. In particular, we propose a kernel-based approach to learn these distributions. To proceed, we first obtain the empirical loss function $\widehat{\mathcal{L}}$ from negative of the ELBO $\mathcal{L}$ by generating $M$ samples of each latent confounder $Z$ using the reparameterization trick \citep{kingma2013auto}: $z_i^\mathsf{s}[l] = f_q(y_i^\mathsf{s},w_i^\mathsf{s},x_i^\mathsf{s}) + \sigma_q\epsilon_i^\mathsf{s}[l]$, where $\epsilon_i^\mathsf{s}[l]$ is drawn from the standard normal distribution. We obtain a complete dataset
\begin{align}
    \widetilde{\mathsf{D}}^\mathsf{s} = \bigcup_{l=1}^{M}\big\{( w_i^\mathsf{s}, y_i^\mathsf{s}, x_i^\mathsf{s}, z_i^\mathsf{s}[l])\big\}_{i=1}^{n_\mathsf{s}}, \quad \forall\mathsf{s} \in \bm{\mathcal{S}}.
\end{align}
Using this complete dataset, we minimize the following objective function
\begin{align}
    J = \widehat{\mathcal{L}} + \sum_{c\in\mathcal{A}} R(f_c)\label{eq:objective-J}
\end{align}
with respect to $f_c$, where $\mathcal{A} = \{y0,y1,w,x,q0,q1\}$, and $R(\cdot)$ denotes a regularizer. The minimizer of $J$ would result in the following form of $f_c$
\begin{align}
    f_c(\bm{u}^\mathsf{s}) = \sum_{\mathsf{v}\in\bm{\mathcal{S}}}\sum_{j=1}^{n_\mathsf{v}\times M}\kappa(\bm{u}^\mathsf{s}, \bm{u}_j^\mathsf{v})\bm{\alpha}_j^\mathsf{v},\label{eq:fun-rt}
\end{align}
where $\bm{u}_j^\mathsf{v}$ is obtained from the $j$--th tuple of the dataset $\tilde{\mathsf{D}}^\mathsf{v}$. Details are presented in Appendix. Since data from the sources might come from a completely different (or similar) distribution, we would use an adaptive kernel to measure their similarity. In particular, let $k(\bm{u}^\mathsf{s}, \bm{u}^\mathsf{v})$ be typical kernel function such as squared exponential kernel, rational quadratic kernel, or Matérn kernel. The kernel used in Eq.~(\ref{eq:fun-rt}) is as follows: $\kappa(\bm{u}^\mathsf{s}, \bm{u}^\mathsf{v}) = 
    \lambda^{\mathsf{s},\mathsf{v}} k(\bm{u}^\mathsf{s}, \bm{u}^\mathsf{v})$, if $\mathsf{s}\neq\mathsf{v}$; otherwise, $\kappa(\bm{u}^\mathsf{s}, \bm{u}^\mathsf{v}) = k(\bm{u}^\mathsf{s}, \bm{u}^\mathsf{v})$,
where $\lambda^{\mathsf{s},\mathsf{v}} \in [0,1]$ is the adaptive factor and it is learned from the observed data.

\noindent\textbf{Remark.} Eq.~(\ref{eq:fun-rt}) indicates that computing $f_c(\bm{u}^\mathsf{s})$ requires collecting all data points from all sources, and so the objective function in Eq.~(\ref{eq:objective-J}) cannot be optimized in a federated setting. Next, we present a method known as Random Fourier Features to address the problem.

\noindent\textbf{Random Fourier Features.}  We show how to adapt Random Fourier Features \citep{rahimi2007random} into our model. Let $k(\bm{u},\bm{u}')$ be any translation-invariant kernel (e.g., squared exponential kernel, rational quadratic kernel, or Matérn kernel). Then, by Bochner's theorem \citep[][Theorem~6.6]{wendland2004scattered}, it can be written in the following form:
\begin{align}
k(\bm{u},\bm{u}') = \int e^{\mathsf{i}\bm{\omega}^\top(\bm{u}-\bm{u}')}s(\bm{\omega})d\bm{\omega} = \int \cos\left(\bm{\omega}^\top(\bm{u}-\bm{u}')\right)s(\bm{\omega})d\bm{\omega},
\end{align}
where $s(\bm{\omega})$ is a spectral density function associated with the kernel (please refer to Appendix for spectral density of some popular kernels). The last equality follows from the fact that the kernel function is real-valued and symmetric. This type of kernel can be approximated by
\begin{align}
    &k(\bm{u},\bm{u}') \simeq \frac{1}{B}\sum_{b=1}^B\cos(\bm{\omega}_b^\top(\bm{u}-\bm{u}')) = \phi(\bm{u})^\top\phi(\bm{u}'), \qquad\{\bm{\omega}_b\}_{b=1}^B \overset{i.i.d.}{\sim} s(\bm{\omega}),
\end{align}
where  $\phi(\bm{u}) = B^{-\frac{1}{2}}[\cos(\bm{\omega}_1^\top \bm{u}),\!...,\cos(\bm{\omega}_B^\top \bm{u}),\sin(\bm{\omega}_1^\top \bm{u}),\!...,\sin(\bm{\omega}_B^\top \bm{u})]^\top$. The last equality follows from the trigonometric identity: $\cos(u-v) = \cos u\cos v + \sin u\sin v$. 
 Substituting the above random Fourier Features into Eq.~(\ref{eq:fun-rt}), we obtain 
\begin{align}
    f_c(\bm{u}^\mathsf{s}) \simeq \Big(\theta_c^\mathsf{s} + \sum_{\mathsf{v}\in\bm{\mathcal{S}}\setminus\{\mathsf{s}\}}\lambda^{\mathsf{s},\mathsf{v}}\theta_c^\mathsf{v}\Big)^\top\phi(\bm{u}^\mathsf{s}), \label{eq:approx-fc}
\end{align}
where $\theta_c^\mathsf{s} = \sum_{i=1}^{n_\mathsf{s}}\phi(\bm{u}^\mathsf{s})\bm{\alpha}^\mathsf{s}_i$ and $\lambda^{\mathsf{s},\mathsf{v}}$ ($\mathsf{s},\mathsf{v}\in\bm{\mathcal{S}}$). While optimizing the objective function $J$, instead of learning $\bm{\alpha}_i^\mathsf{s}$, we can directly consider $\theta^\mathsf{s}$ as parameter to be optimized. This has been used in several works such as \citet{rahimi2007random, chaudhuri2011differentially,rajkumar2012differentially}.  
This approximation allows us to rewrite the objective function $J$ as a summation of local objective functions in each source:
\vskip -15pt
\begin{align}
J \simeq \sum_{\mathsf{s}\in\bm{\mathcal{S}}}J^\mathsf{(s)}, \qquad \text{where } J^\mathsf{(s)} = \widehat{\mathcal{L}}^{(\mathsf{s})} + \frac{1}{m}\sum_{\mathsf{v}\in\bm{\mathcal{S}}} \zeta\|\theta^\mathsf{v}\|_2^2,
\end{align}
where $\zeta \in \mathbb{R}^+$ is a regularizer factor. Each component $J^\mathsf{(s)}$ is associated with the source $\mathsf{s}$ and it can be computed with the local data in this source. Hence, it enables federated optimization for the objective function $J$. This objective function can be optimised with Algorithm~\ref{algo:federated-training}.

\noindent\textbf{Minimax lower bound.} We now compute the minimax lower bound of the proposed model, which gives the rate at which our estimator can converge to the population quantity of interest as the sample size increases. We first state the following result that concerns the last two terms in Eq.~(\ref{eq:target-est}):
\begin{lemma}[With presence of latent variables]
\label{lem:minimax-bounds1}
Let $\bm{\uptheta} = \{\theta_c^\mathsf{s}:c\in\{y0,y1,x,w\}, \mathsf{s}\in \bm{\mathcal{S}}\}$ and $\hat{\bm{\uptheta}}$ be its estimate. Let $y_i^\mathsf{s}\in\mathbb{R}$ and $x_i^\mathsf{s}\in\mathbb{R}^{d_x}$. Let $\bm{\mathcal{S}}_{\backslash\mathsf{s}} = \bm{\mathcal{S}} \setminus \{\mathsf{s}\}$. Then,
\begin{align}
&\displaystyle\inf_{\hat{\bm{\uptheta}}} \sup_{P\in\mathcal{P}} \mathbb E_{P} \!\!\left[\! \|\hat{\bm{\uptheta}}-\bm{\uptheta}(P)\|_2 \!\right] \ge  \frac{\sqrt{m(d_x+3)}\log(2\sqrt{m})}{64\sqrt{B}\sum_{\mathsf{s}\in\bm{\mathcal{S}}}n_\mathsf{s}\big(1+\sum_{\mathsf{v}\in\bm{\mathcal{S}}_{\backslash\mathsf{s}}}\!\!\lambda^{\mathsf{s},\mathsf{v}}\big)^2}.\label{eq:lem-1}
\end{align}
\end{lemma}
The LHS of Eq.~(\ref{eq:lem-1}) can be seen as the worst case of the best estimator, whereas the RHS depicts the behavior of the convergence. The bounds do not only depend on the number of samples ($n_\mathsf{s}$, training size) of each source but also the adaptive factors $\lambda^{\mathsf{s},\mathsf{v}}$. When the adaptive factors are small, the lower bounds are large since data from a source $\mathsf{s}$ are only used to learn its own parameter $\theta^\mathsf{s}$. When the adaptive factors are large, the lower bounds are smaller, which suggests that data from a source would help infer parameters associated with the other sources.
This bound gives a guarantee on how data from all the sources impact the learned parameters that modulate the two distributions $\p_\mathsf{s}(z^\mathsf{s}|x^\mathsf{s}, y^\mathsf{s},w^\mathsf{s})$ and $\p_\mathsf{s}(y^\mathsf{s}|w^\mathsf{s},z^\mathsf{s})$. The proof of Lemma~\ref{lem:minimax-bounds1} 
can be found in Appendix~\ref{ap:proof-lem-1}.

\subsection{Learning Auxiliary Distributions}
\label{sec:auxiliary-dists}
The previous section has shown how to learn $\p_\mathsf{s}(z^\mathsf{s}|x^\mathsf{s}, y^\mathsf{s},w^\mathsf{s})$ and $\p_\mathsf{s}(y^\mathsf{s}|w^\mathsf{s},z^\mathsf{s})$. To compute treatment effects, we need to learn two more conditional distributions, namely $\p_\mathsf{s}(w^\mathsf{s}|x^\mathsf{s})$ and $\p_\mathsf{s}(y^\mathsf{s}|x^\mathsf{s},w^\mathsf{s})$. Since all the variables in these two distributions are observed, we estimate them using maximum likelihood estimation. In the following, we present a federated setting to learn $\p_\mathsf{s}(w^\mathsf{s}|x^\mathsf{s})$. Similar to the previous section, the objective function here can also be decomposed into $m$ components as follows: $J_w \simeq \sum_{\mathsf{s}\in\bm{\mathcal{S}}}J_w^{(\mathsf{s})}$, where $J_w^{(\mathsf{s})} = \sum_{i=1}^{n_\mathsf{s}}\ell(w_i^\mathsf{s}, \varphi(g(x_i^\mathsf{s}))) + m^{-1}\sum_{\mathsf{v}\in\bm{\mathcal{S}}}\zeta_w\|\psi^\mathsf{v}\|_2^2$ and $g(x_i^\mathsf{s}) = \sum_{\mathsf{v}\in\bm{\mathcal{S}}}\phi(x_i^\mathsf{s})^\top(\psi^\mathsf{s} + \gamma^{\mathsf{s},\mathsf{v}}\psi^\mathsf{v})$, $\gamma^{\mathsf{s},\mathsf{v}}\in[0,1]$ is the adaptive factor, $\psi^\mathsf{s}$ is the parameter associated with source $\mathsf{s}$, and $\ell(\cdot)$ denotes the cross-entropy loss function since $w_i^\mathsf{s}$ is a binary value. The first component of $J_w^{(\mathsf{s})}$ is obtained from the negative log-likelihood. Learning of $\p_\mathsf{s}(y^\mathsf{s}|x^\mathsf{s},w^\mathsf{s})$ is similar. For convenience, in the subsequent analyses, we denote the parameters and adaptive factors of this distribution as $\beta^\mathsf{s}$ and $\eta^{\mathsf{s},\mathsf{v}}$, where $\mathsf{s},\mathsf{v}\in \bm{\mathcal{S}}$ and $\mathsf{s} \neq \mathsf{v}$. 
The next lemma shows the minimax lower bound for the first two sets of parameters $\bm{\uppsi}$ and $\upbeta$ in Eq.~(\ref{eq:target-est}), but this time without involving the latent variables:
\begin{lemma}[Without the presence of latent variables]
\label{lem:mimimax-bounds2}
Let $\bm{\uppsi} = \{\psi^\mathsf{s}\}_{\mathsf{s}=1}^m$, $\bm{\upbeta} = \{\beta^\mathsf{s}\}_{\mathsf{s}=1}^m$ and $\hat{\bm{\uppsi}}$, $\hat{\bm{\upbeta}}$ be their estimates, respectively. Let $y_i^\mathsf{s}\in\mathbb{R}$. Then, 
\begin{align}
&\,\textbf{\emph{(i)}}\, \inf_{\hat {\bm{\uppsi}}} \sup_{P\in\mathcal{P}} \mathbb E_{P} \left[ \|\hat{\bm{\uppsi}} - \bm{\uppsi}(P)\|_2 \right] \ge \frac{m\log(2\sqrt{m})}{256\sum_{\mathsf{s}\in\bm{\mathcal{S}}}n_\mathsf{s}\big(1 + \sum_{\mathsf{v}\in\bm{\mathcal{S}}_{\backslash\mathsf{s}}}\gamma^{\mathsf{s},\mathsf{v}}\big)},\\&\textbf{\emph{(ii)}}\,\inf_{\hat{\bm{\upbeta}}} \sup_{P\in\mathcal{P}} \mathbb E_{P} \left[ \|\hat{\bm{\upbeta}} - \bm{\upbeta}(P)\|_2 \right] \ge \frac{\sigma}{2^{\frac{9}{2}}}\bigg(\frac{m\log(2\sqrt{m})}{B\sum_{\mathsf{s}\in\bm{\mathcal{S}}}n_\mathsf{s}\big(1 + \sum_{\mathsf{v}\in\bm{\mathcal{S}}_{\backslash\mathsf{s}}}\eta^{\mathsf{s},\mathsf{v}}\big)^2}\bigg)^{1/2}.
\end{align}
\end{lemma}
The proof of Lemma~\ref{lem:mimimax-bounds2} 
can be found in Appendix~\ref{ap:proof-lem-2}. 
The bounds presented in Lemma~\ref{lem:minimax-bounds1} and \ref{lem:mimimax-bounds2} give helpful information about the number of samples to be observed and the cooperation of multiple sources of data through the transfer factors. Since we used variational inference and maximum likelihood to learn the parameters in our model, these methods give consistent estimation as shown in \citet{kiefer1956consistency,van2000asymptotic,wang2019frequentist,yang2020alpha}.

\subsection{Computing Causal Effects}
The key to estimate causal effects in our model is to compute the outcome in Eq.~(\ref{eq:est-causal-eff}). We proceed by drawing samples from the distributions in Eq.~(\ref{eq:target-est}). 
Generating samples from the conditional distributions $\p_\mathsf{s}(w^\mathsf{s}|x^\mathsf{s})$, $\p_\mathsf{s}(y^\mathsf{s}|x^\mathsf{s},w^\mathsf{s})$, and $\p_\mathsf{s}(y^\mathsf{s}|w^\mathsf{s},z^\mathsf{s})$ is straightforward since they are readily available as shown in either Section~\ref{sec:latent-dists}~or~\ref{sec:auxiliary-dists}. There are two options to draw samples from the posterior distribution of confounder $\p_\mathsf{s}(z^\mathsf{s}|x^\mathsf{s}, y^\mathsf{s},w^\mathsf{s})$. The first one is to draw from its approximation, $\q(z^\mathsf{s})$, since maximizing the ELBO in Section~\ref{sec:latent-dists} is equivalent to minimizing $\textrm{KL}(q(z^\mathsf{s})\|\p_\mathsf{s}(z^\mathsf{s}|x^\mathsf{s}, y^\mathsf{s},w^\mathsf{s}))$. 
As a second option, we note that the exact posterior of confounder can be rewritten as  $\p_\mathsf{s}(z^\mathsf{s}|x^\mathsf{s}, y^\mathsf{s},w^\mathsf{s}) \propto \p_\mathsf{s}(y^\mathsf{s}|z^\mathsf{s},w^\mathsf{s})\p_\mathsf{s}(w^\mathsf{s}|z^\mathsf{s})\p_\mathsf{s}(x^\mathsf{s}|z^\mathsf{s})\p(z^\mathsf{s})$, whose components on the right hand side are also available in Section~\ref{sec:latent-dists}. Thus, we can draw from this distribution using the Metropolis-Hastings (MH) algorithm. Since $Z$ is a multidimensional random variable, the traditional MH algorithm would require a long chain to converge. We overcome this problem by using the MH with independent sampler \citep{liu1996metropolized} where the proposal distribution is the variational posterior distribution $q(z^\mathsf{s})$ learned in Section~\ref{sec:latent-dists}. The second approach would give more accurate samples since we select the samples based on exact acceptance probability of the posterior  $\p_\mathsf{s}(z^\mathsf{s}|x^\mathsf{s}, y^\mathsf{s},w^\mathsf{s})$. This would help estimate the CATE given $x_i^\mathsf{s}$. The local ATE is the average of CATE of individuals in a source $\mathsf{s}$. These quantities can be estimated in a local source machine. 
To compute a global ATE, the server would collect all the local ATE in each source and then compute their weighted average. Further details are in Appendix.

\subsection{Experiments}
\label{sec:experiment-causalrff}
\textbf{The baselines.} In this section, we first carry out the experiments to examine the performance of CausalRFF against standard baselines such as BART \citep{hill2011bayesian}, TARNet \citep{shalit2017estimating}, CFR-wass (CFRNet with Wasserstein distance) \citep{shalit2017estimating}, CFR-mmd (CFRNet with maximum mean discrepancy distance) \citep{shalit2017estimating}, CEVAE \citep{louizos2017causal}, 
OrthoRF \citep{oprescu2019orthogonal}, 
X-learner \citep{kunzel2019metalearners}, R-learner \citep{nie2020quasi}, and FedCI \citep{vo2022bayesian}. In contrast to CausalRFF, these methods (except FedCI) do not consider causal inference within a federated setting. We compare our method to these baselines trained in two ways:  (\textbf{a}) training a global model with the combined data from all the sources, (\textbf{b}) using bootstrap aggregating of \citet{breiman1996bagging} where $m$ models are trained separately on each source data and then averaging the predicted treatment effects based on each trained model. Note that case (\textbf{a})  \textit{violates} federated data setting and is only used for comparison purposes. In general, we expect that the performance of CausalRFF to be close to that of the performance of the baselines in case (\textbf{a}) when the data distribution of all the sources are the same. In addition, we also show that the performance of CausalRFF is better than that of the baselines in case (\textbf{a}) when the data distribution of all the source are different.

\noindent\textbf{Implementation of the baselines.} The implementation of CEVAE is from \citet{louizos2017causal}. Implementation of TARNet, CFR-wass, and CFR-mmd are from \citet{shalit2017estimating}. For these methods, we use Exponential Linear Unit (ELU) activation function and fine-tune the number of nodes in each hidden later from 10 to 200 with step size of addition by 10. For BART, we use package \texttt{BartPy}, which is readily available. For X-learner and R-learner, we use the package \texttt{causalml} \citep{chen2020causalml}. For OrthoRF, we use the package \texttt{econml} \citep*{econml}. For FedCI, we use the code from \citet{vo2022bayesian}. For all methods, the learning rate is fine-tuned from $10^{-4}$ to $10^{-1}$ with step size of multiplication by $10$. Similarly, the regularizer factors are also fine-tuned from $10^{-4}$ to $10^0$ with step size of multiplication by $10$. 
We report two error metrics: $\epsilon_\mathrm{PEHE}$ (precision in estimation of heterogeneous effects) and $\epsilon_\mathrm{ATE}$ (absolute error) to compare the methods.  We report the mean and standard error over 10 replicates of the data. Further details are presented in Appendix.

\subsubsection{Synthetic Data}
\label{sec:syn-data}
\textbf{Data description.} Obtaining ground truth for evaluating causal inference algorithm is a challenging task. Thus, most of the state-of-the-art methods are evaluated using synthetic or semi-synthetic datasets. In this experiment, the synthetic data is simulated with the following distributions:
\begin{align*}
    z_i^\mathsf{s} &\sim \mathsf{Cat}(\rho),\\
    x_{ij}^\mathsf{s} &\sim \mathsf{Bern}(\varphi(a_{j0} + (z_i^\mathsf{s})^\top \mathbf{a}_{j1})),\\
    w_i^\mathsf{s} &\sim \mathsf{Bern}(\varphi(b_0 + (z_i^\mathsf{s})^\top (\mathbf{b}_1+\Delta))),\\
    y_i^\mathsf{s}(0) &\sim \mathsf{N}(\mathsf{sp}(c_0 + (z_i^\mathsf{s})^\top (\mathbf{c}_1+\Delta)), \sigma_0^2),\\
    y_i^\mathsf{s}(1) &\sim \mathsf{N}(\mathsf{sp}(d_0 + (z_i^\mathsf{s})^\top (\mathbf{d}_1+\Delta)), \sigma_1^2),
\end{align*}
where $\mathsf{Cat}(\cdot)$, $\mathsf{N}(\cdot)$, and $\mathsf{Bern}(\cdot)$ denote the categorical distribution, normal distribution, and Bernoulli distribution, respectively. $\varphi(\cdot)$ denotes the sigmoid function, $\mathsf{sp}(\cdot)$ denotes the softplus function, and $x_i = [x_{i1},\!...,x_{id_x}]^\top \in \mathbb{R}^{d_x}$ with $d_x=30$. Herein, we convert $z_i^\mathsf{s}$ to a one-hot vector. To simulate data, we randomly set the ground truth parameters as follows: $\rho=[.11, .17, .34, .26, .12]^\top$, $(c_0,d_0)=(0.9,7.9)$, $(\mathbf{c}_1,\mathbf{d}_1,\mathbf{d}_1)$ are drawn i.i.d from $\mathsf{N}(\bm{0},2\mathbf{I}_5)$, $a_{j0}$ and elements of $\mathbf{a}_{j1}$ are drawn i.i.d from $\mathsf{N}(0,2)$. 
For each source, we simulate $10$ replications with $n_\mathsf{s} = 1000$ records. We only keep $\{(y_i^\mathsf{s}, w_i^\mathsf{s}, x_i^\mathsf{s})\}_{i=1}^{n_\mathsf{s}}$ as the observed data, where $y_i^\mathsf{s} = y_i^\mathsf{s}(0)$ if $w_i^\mathsf{s}=0$ and $y_i^\mathsf{s} = y_i^\mathsf{s}(1)$ if $w_i^\mathsf{s}=1$. In each source, we use $50$ data points for training, $450$ for testing and $400$ for validating. We report the evaluation metrics and their standard errors over the 10 replications.

\begin{figure}
    \centering
\includegraphics[width=0.65\textwidth]{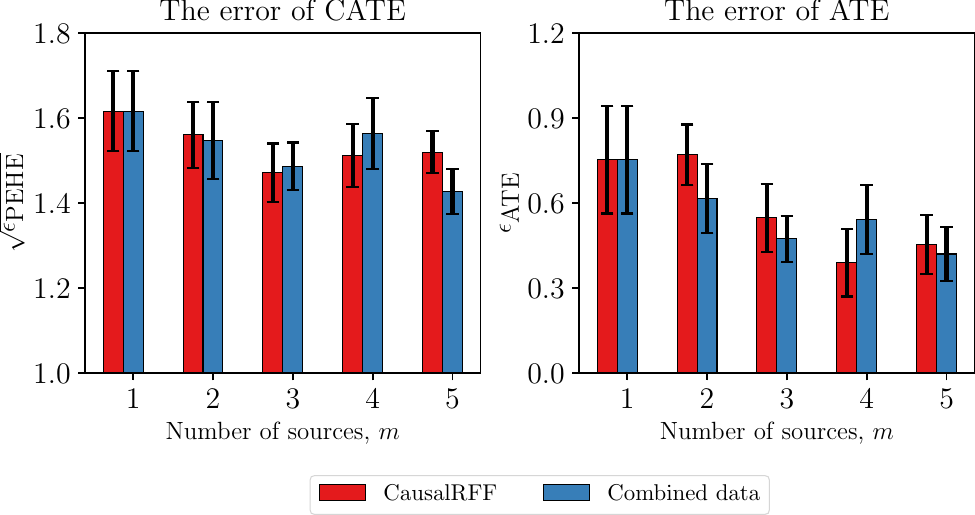}
\caption{Experimental results on DATA$^\mathsf{same}$.}
        \label{fig:analysis1}
\end{figure}

\begin{figure}
    \centering
\includegraphics[width=0.65\textwidth]{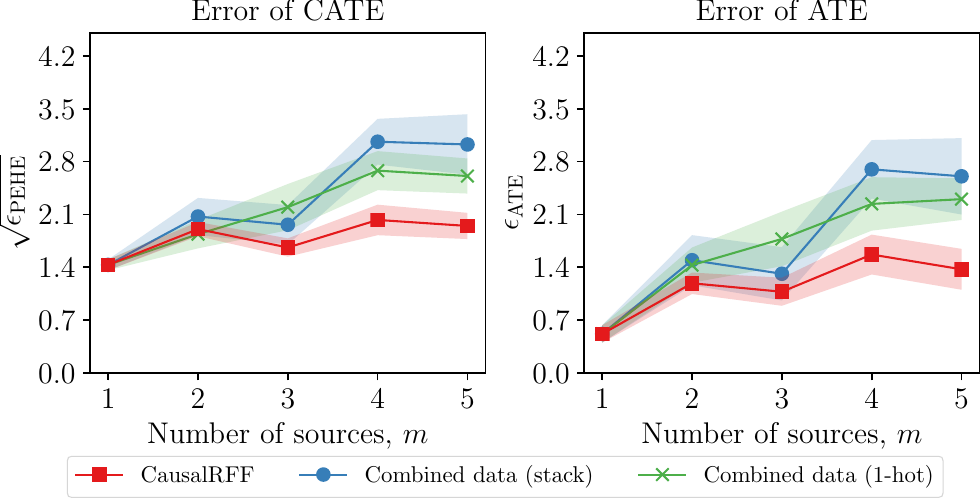}
\caption{Experimental results on DATA$^\mathsf{diff}$.}
        \label{fig:analysis2}
\end{figure}

\begin{table}
    \centering
\caption{Out-of-sample errors on DATA$^\mathsf{same}$ where top-3 performances are highlighted in bold (lower is better). The dashes (-) in `$\text{ag}$' (bootstrap aggregating)  indicate that the numbers are the same as that of `$\text{cb}$' (combined data).}
\vspace{6pt}
\label{tb:data-same}
\setlength{\tabcolsep}{12pt}
\centering
\scriptsize
\begin{tabular}{@{}lcccccc@{}}
\toprule
\multirow{2}{*}{Method} & \multicolumn{3}{c}{The error of CATE, $\sqrt{\epsilon_\text{PEHE}}$}                                              & \multicolumn{3}{c}{The error of ATE, $\epsilon_\text{ATE}$}                                               \\ \cmidrule(lr){2-4}\cmidrule(l){5-7} 
& 1 source                     & 3 sources                    & 5 sources                    & 1 source                     & 3 sources                   & 5 sources                    \\ \cmidrule(r){1-1}\cmidrule(lr){2-4}\cmidrule(l){5-7}
BART$_\text{ag}$       & -                            & 3.8$\pm$.10                     & 3.8$\pm$.09                     & -                            & 2.3$\pm$.15                     & 2.3$\pm$.14                     \\
X-Learner$_\text{ag}$  & -                            & 3.2$\pm$.07                     & 3.1$\pm$.06                     & -                            & 0.6$\pm$.11                     & 0.5$\pm$.13                     \\
R-Learner$_\text{ag}$  & -                            & 3.5$\pm$.17                     & 3.9$\pm$.46                     & -                            & 1.5$\pm$.35                     & 2.0$\pm$.70                     \\
OthoRF$_\text{ag}$     & -                            & 5.4$\pm$.21                     & 4.5$\pm$.12                     & -                            & 0.5$\pm$.10                     & 0.7$\pm$.16                     \\
TARNet$_\text{ag}$          & -                     & 3.9$\pm$.04                     & 3.4$\pm$.03                     & -                     & 2.2$\pm$.07                     & 2.0$\pm$.02                     \\
CFR-wass$_\text{ag}$          & -                     & 3.0$\pm$.05                     & 3.6$\pm$.02                     & -                     & 2.1$\pm$.03                     & 1.8$\pm$.02                     \\
CFR-mmd$_\text{ag}$          & -                     & 4.0$\pm$.03                     & 3.9$\pm$.02                     & -                     & 2.3$\pm$.03                     & 2.0$\pm$.01                     \\
CEVAE$_\text{ag}$      & -                            & \textbf{2.9$\pm$.0}4 & \textbf{2.5$\pm$.04} & -                            & 0.7$\pm$.08 & \textbf{0.5$\pm$.10} \\\cmidrule(r){1-1}\cmidrule(lr){2-4}\cmidrule(l){5-7}
BART$_\text{cb}$           & \textbf{3.7$\pm$.12}                     & 3.2$\pm$.07                     & 3.1$\pm$.03                     & 2.1$\pm$.20                     & 1.0$\pm$.18                     & 0.6$\pm$.13                     \\
X-Learner$_\text{cb}$       & 3.3$\pm$.06                     & 3.4$\pm$.06                     & 3.3$\pm$.04                     & \textbf{0.5$\pm$.11}                      & \textbf{0.4$\pm$.06}                     & \textbf{0.5$\pm$.12}                     \\
R-Learner$_\text{cb}$       & 4.2$\pm$.46                     & 3.4$\pm$.07                     & 3.4$\pm$.04                     & 2.2$\pm$.72                      & 0.6$\pm$.15                     & 0.9$\pm$.15                     \\
OthoRF$_\text{cb}$          & 7.6$\pm$.29                     & 4.3$\pm$.10                     & 3.7$\pm$.07                     & 1.4$\pm$.30                     & \textbf{0.4$\pm$.12}                     & \textbf{0.5$\pm$.10}                     \\
TARNet$_\text{cb}$          & 4.2$\pm$.07                     & 3.8$\pm$.03                     & 3.5$\pm$.02                     & 2.2$\pm$.13                     & 2.1$\pm$.06                     & 2.1$\pm$.03                     \\
CFR-wass$_\text{cb}$          & 4.0$\pm$.11                     & 3.8$\pm$.02                     & 3.7$\pm$.02                     & 2.1$\pm$.06                     & 2.0$\pm$.03                     & 1.9$\pm$.02                     \\
CFR-mmd$_\text{cb}$          & 3.8$\pm$.05                     & 3.8$\pm$.02                     & 3.7$\pm$.02                     & 2.1$\pm$.04                     & 2.1$\pm$.03                     & 2.0$\pm$.02                     \\
CEVAE$_\text{cb}$           & \textbf{2.5$\pm$.03} & \textbf{2.4$\pm$.03} & \textbf{2.4$\pm$.03} & \textbf{0.5$\pm$.08} & \textbf{0.3$\pm$.06} & \textbf{0.3$\pm$.06} \\\cmidrule(r){1-1}\cmidrule(lr){2-4}\cmidrule(l){5-7}
FedCI           & \textbf{2.5$\pm$.03} & \textbf{2.4$\pm$.03} & \textbf{2.5$\pm$.03} & \textbf{0.4$\pm$.06} & \textbf{0.3$\pm$.11} & \textbf{0.3$\pm$.10} \\\cmidrule(r){1-1}\cmidrule(lr){2-4}\cmidrule(l){5-7}
CausalRFF                & \textbf{1.6$\pm$.09}                     & \textbf{1.5$\pm$.07}                     & \textbf{1.5$\pm$.05}                     & \textbf{0.8$\pm$.19}                     & \textbf{0.5$\pm$.12}                     & \textbf{0.4$\pm$.10}                     \\ \bottomrule
\end{tabular}
\end{table}

\noindent\textbf{Result and discussion (I).} In the first experiment, we study the performance of CausalRFF on multiple sources whose data distributions are the same. To do that, we simulate $m=5$ sources from the same distribution, i.e., we set the ground truth $\Delta=0.0$ for all the sources. We refer to this dataset as DATA$^\mathsf{same}$. In this experiment, we expect that the result of CausalRFF, which is trained in federated setting, is as good as training on combined data. The results in Figure~\ref{fig:analysis1} show that the error in two cases seem to move together in a correlated fashion, which verifies our hypothesis.

In addition, to study the performance of CausalRFF on the sources whose data distributions are different, we also simulate $m=5$ sources. However, the first source is with $\Delta=0.0$ and the other four sources are with $\Delta=4.0$. We refer to this dataset as DATA$^\mathsf{diff}$. We test the error of CATE and ATE on the first source. In this case, we expect that the errors of CausalRFF to be lower than that of training on combined data since CausalRFF learns the adaptive factors which prevent negative impact of the other four sources to the first source. The results in Figure~\ref{fig:analysis2} show that CausalRFF achieves lower errors compared to training on combined data (there are two cases of combining: stacking data, and adding one-hot vectors to indicate the source of each data point), which is as expected.

\begin{figure}
    \centering
\includegraphics[width=0.65\textwidth]{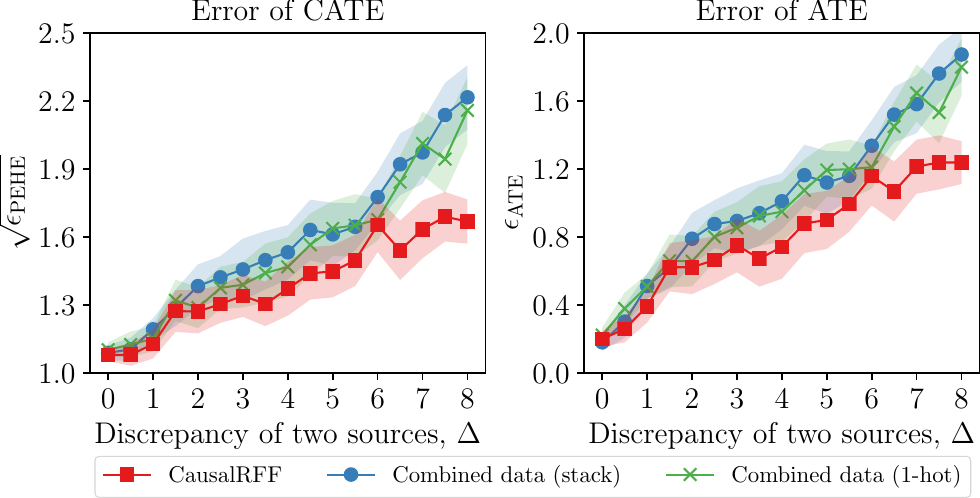}
\caption{Experimental results on different levels of discrepancy, $\Delta$.}
        \label{fig:analysis3}
\end{figure}

In the third experiment, we study the effect of $\Delta$ on the performance of CausalRFF. We simulate $m=2$ sources with different values of $\Delta$. In particular, the first source is with $\Delta=0.0$ and the second source is with $\Delta$ varying from 0.0 to 8.0. We compare our CausalRFF method with that of training on combined data. Again, Figure~\ref{fig:analysis3} shows that CausalRFF achieves lower errors as expected.

\begin{table}
    \centering
\caption{Out-of-sample errors on DATA$^\mathsf{diff}$. The dashes (-) in `$\text{ag}$' indicate that the numbers are the same as that of `$\text{cb}$'.
}
\vspace{6pt}
\label{tb:data-diff}
\setlength{\tabcolsep}{12pt}
\centering
\scriptsize
\begin{tabular}{@{}lcccccc@{}}
\toprule
\multirow{2}{*}{Method} & \multicolumn{3}{c}{The error of CATE, $\sqrt{\epsilon_\text{PEHE}}$}                                              & \multicolumn{3}{c}{The error of ATE, $\epsilon_\text{ATE}$}                                               \\ \cmidrule(l){2-4}\cmidrule(l){5-7} 
& 1 source                      & 3 sources                     & 5 sources                    & 1 source                      & 3 sources                    & 5 sources                    \\ \cmidrule(r){1-1}\cmidrule(lr){2-4}\cmidrule(l){5-7}
BART$_\text{ag}$       & -                            & \textbf{3.0$\pm$.01}                     & \textbf{3.0$\pm$.02}                     & -                            & 1.3$\pm$.05                     & \textbf{1.4$\pm$.10}                     \\
X-Learner$_\text{ag}$  & -                            & 3.3$\pm$.03                     & 3.3$\pm$.04                     & -                            & \textbf{1.2$\pm$.09}                     & \textbf{1.3$\pm$.09}                     \\
R-Learner$_\text{ag}$  & -                            & 3.2$\pm$.03                     & 3.1$\pm$.02                     & -                            & \textbf{1.0$\pm$.07}                     & \textbf{1.2$\pm$.09}                     \\
OthoRF$_\text{ag}$     & -                            & 3.6$\pm$.05                     & 3.6$\pm$.05                             & -                            & 1.3$\pm$.09                     &  1.6$\pm$.10                            \\
TARNet$_\text{ag}$         & -                     & 6.1$\pm$.19                     & 5.7$\pm$.05                     & -                     & 2.5$\pm$.06                     & 3.0$\pm$.05                     \\
CFR-wass$_\text{ag}$         & -                     & 5.6$\pm$.09                     & 5.7$\pm$.07                     & -                     & 2.7$\pm$.05                     & 2.8$\pm$.04                     \\
CFR-mmd$_\text{ag}$         & -                     & 5.9$\pm$.08                     & 5.6$\pm$.05                     & -                     & 2.5$\pm$.03                     & 2.8$\pm$.02                     \\
CEVAE$_\text{ag}$      & -                            & 4.2$\pm$.07 & 3.9$\pm$.05 & -                            & 2.1$\pm$.09 & 1.8$\pm$.10 \\\cmidrule(r){1-1}\cmidrule(lr){2-4}\cmidrule(l){5-7}
BART$_\text{cb}$           & 3.1$\pm$.05                     & 4.1$\pm$.10                     & 4.2$\pm$.10                     & 0.8$\pm$.17                     & 2.8$\pm$.15                     & 2.9$\pm$.14                     \\
X-Learner$_\text{cb}$      & 3.3$\pm$.03                     & 5.0$\pm$.08                     & 4.6$\pm$.10                     & \textbf{0.5$\pm$.12}                     & 3.3$\pm$.11                     & 3.1$\pm$.13                     \\
R-Learner$_\text{cb}$      & 3.3$\pm$.05                     & 3.5$\pm$.05                     & 3.3$\pm$.05                     & 0.7$\pm$.18                     & \textbf{1.1$\pm$.10}                     & \textbf{1.3$\pm$.10}                     \\
OthoRF$_\text{cb}$         & 3.9$\pm$.06                     & 5.2$\pm$.10                     & 4.6$\pm$.09                     & \textbf{0.5$\pm$.11}                     & 3.3$\pm$.14                     & 3.0$\pm$.12                     \\
TARNet$_\text{cb}$         & 4.2$\pm$.07                     & 5.9$\pm$.09                     & 5.8$\pm$.06                     & 2.2$\pm$.13                     & 2.3$\pm$.04                     & 2.9$\pm$.02                     \\
CFR-wass$_\text{cb}$         & 4.0$\pm$.11                     & 5.7$\pm$.08                     & 5.5$\pm$.08                     & 1.9$\pm$.06                     & 2.4$\pm$.03                     & 2.9$\pm$.04                     \\
CFR-mmd$_\text{cb}$         & 3.8$\pm$.05                     & 5.7$\pm$.08                     & 5.5$\pm$.04                     & 2.1$\pm$.04                     & 2.4$\pm$.03                     & 2.9$\pm$.04                     \\
CEVAE$_\text{cb}$          & \textbf{2.4$\pm$.03} & 5.0$\pm$.06 & 4.4$\pm$.07 & \textbf{0.3$\pm$.08} & 2.6$\pm$.10 & 2.0$\pm$.07 \\\cmidrule(r){1-1}\cmidrule(lr){2-4}\cmidrule(l){5-7}
FedCI           & \textbf{2.5$\pm$.03} & \textbf{2.6$\pm$.04} & \textbf{2.8$\pm$.04} & \textbf{0.2$\pm$.06} & \textbf{1.2$\pm$.12} & 1.5$\pm$.13 \\\cmidrule(r){1-1}\cmidrule(lr){2-4}\cmidrule(l){5-7}
CausalRFF                & \textbf{1.4$\pm$.07}                     & \textbf{1.7$\pm$.12}                     & \textbf{1.9$\pm$.17}                     & \textbf{0.5$\pm$.11}                     & \textbf{1.1$\pm$.19}                     & \textbf{1.4$\pm$.27}                     \\ \bottomrule
\end{tabular}
\end{table}

\noindent\textbf{Result and discussion (II).} 
This section aims to compare CausalRFF with the baselines on both datasets: DATA$^\mathsf{same}$ and DATA$^\mathsf{diff}$. Except FedCI (which is a Bayesian federated method), the other baselines are trained on two cases: combined data (cb) and bootstrap aggregating (ag) as mentioned earlier. On DATA$^\mathsf{same}$, we expect that the performance of the proposed method is as good as the baselines trained on combined data. The results in Table~\ref{tb:data-same} show that the performance of CausalRFF is as expected. 
For DATA$^\mathsf{diff}$, we report the results on Table~\ref{tb:data-diff}. The figures reveal that the performance of CausalRFF is as good as the baselines in predicting ATE. In terms of predicting CATE, the performance of the baselines significantly reduces as we add more data sources whose distribution are different from the first source. Meanwhile, the performance of CausalRFF in predicting CATE is slightly reduced, but it is still much better than those of the baselines. The reason of this is because we used adaptive factors to learn for the similarity of data distributions among the sources.

\subsubsection{Large-scale Synthetic Data}

\textbf{Data description.} In this section, we conduct experiments on a large number of sources. The set up in this section is similar to that of Section~\ref{sec:syn-data}. We simulate two cases: (1) DATA-LARGE$^\mathsf{same}$: a dataset of 100 sources, where we set $\Delta=0$ for all sources so that their distributions are the same. (2) DATA-LARGE$^\mathsf{diff}$: a dataset of 100 sources, where we draw uniformly the discrepancy factor $\Delta \sim \mathsf{U}[0,8]$ for each source so that their distributions are different. In both cases, we use test set from the first 20 sources for evaluation.

\noindent\textbf{Result and discussion.}  Table~\ref{tb:data-large-same}  shows that CausalRFF achieves competitive results in estimating ATE and CATE when the sources have the same distribution. Table~\ref{tb:data-large-diff} shows that CausalRFF outperforms the baselines when the sources have different distributions. These results are consistent with our discussions in Section~\ref{sec:syn-data}.

\subsubsection{IHDP Data}

\begin{table}
    \caption{Errors on DATA-LARGE$^\mathsf{same}$ dataset. }
\vspace{6pt}
\label{tb:data-large-same}
\setlength{\tabcolsep}{12pt}
\centering
\scriptsize
\begin{tabular}{@{}lcccccc@{}}
\toprule
\multirow{2}{*}{Method} & \multicolumn{3}{c}{The error of CATE, $\sqrt{\epsilon_\text{PEHE}}$}                                              & \multicolumn{3}{c}{The error of ATE, $\epsilon_\text{ATE}$}                                               \\ \cmidrule(l){2-4}\cmidrule(l){5-7} 
& \makecell[c]{20\\ sources}      & \makecell[c]{50\\ sources}      & \makecell[c]{100\\ sources}     & \makecell[c]{20\\ sources}     & \makecell[c]{50\\ sources}     & \makecell[c]{100\\ sources}     \\ \cmidrule(r){1-1}\cmidrule(lr){2-4}\cmidrule(l){5-7}
BART$_\text{cb}$      & 3.4$\pm$.03 & 3.4$\pm$.01 & 3.3$\pm$.01 & 1.4$\pm$.06 & 1.3$\pm$.02 & 1.3$\pm$.01 \\
X-Learner$_\text{cb}$ & 3.0$\pm$.01 & 2.9$\pm$.01 & 2.9$\pm$.01 & \textbf{.16$\pm$.02} & \textbf{.12$\pm$.02} & \textbf{.13$\pm$.02} \\
R-Learner$_\text{cb}$ & 3.0$\pm$.01 & 2.9$\pm$.01 & 2.9$\pm$.01 & \textbf{.07$\pm$.01} & \textbf{.10$\pm$.02} & \textbf{.10$\pm$.02} \\
OthoRF$_\text{cb}$    & 3.4$\pm$.03 & 3.3$\pm$.01 & 3.2$\pm$.01 & 1.2$\pm$.06 & 1.1$\pm$.02 & 1.0$\pm$.02 \\
TARNet$_\text{cb}$    & 3.8$\pm$.03 & 3.7$\pm$.01 & 3.3$\pm$.01 & 1.1$\pm$.02 & 1.0$\pm$.01 & .93$\pm$.01 \\
CFR-wass$_\text{cb}$  & 3.7$\pm$.02 & 3.6$\pm$.01 & 3.2$\pm$.01 & 1.1$\pm$.02 & .99$\pm$.01 & .87$\pm$.01 \\
CFR-mmd$_\text{cb}$   & 3.7$\pm$.02 & 3.6$\pm$.01 & 3.2$\pm$.01 & 1.1$\pm$.02 & .98$\pm$.01 & .87$\pm$.01 \\
CEVAE$_\text{cb}$     & \textbf{2.3$\pm$.01} & \textbf{2.2$\pm$.01} & \textbf{2.0$\pm$.01} & \textbf{.19$\pm$.03} & \textbf{.17$\pm$.01} & .17$\pm$.01 \\ \cmidrule(r){1-1}\cmidrule(lr){2-4}\cmidrule(l){5-7}
FedCI           & \textbf{2.2$\pm$.02} & \textbf{2.2$\pm$.01} & \textbf{1.9$\pm$.01} & .23$\pm$.04 & .21$\pm$.01 & .19$\pm$.01 \\\cmidrule(r){1-1}\cmidrule(lr){2-4}\cmidrule(l){5-7}
CausalRFF                & \textbf{1.6$\pm$.05} & \textbf{1.6$\pm$.01} & \textbf{1.5$\pm$.01} & 0.3$\pm$.04 & 0.2$\pm$.02 & \textbf{.16$\pm$.02}                     \\ \bottomrule
\end{tabular}
\end{table}

\begin{table}
    \caption{Errors on DATA-LARGE$^\mathsf{diff}$ dataset. } \label{tb:data-large-diff}
\vspace{6pt}
\setlength{\tabcolsep}{12pt}
\scriptsize
\centering
\begin{tabular}{@{}lcccccc@{}}
\toprule
\multirow{2}{*}{Method} & \multicolumn{3}{c}{The error of CATE, $\sqrt{\epsilon_{\text{PEHE}}}$} & \multicolumn{3}{c}{The error of ATE, $\epsilon_{\text{ATE}}$} \\ \cmidrule(lr){2-4}\cmidrule(l){5-7} 
& \makecell[c]{20\\ sources}      & \makecell[c]{50\\ sources}      & \makecell[c]{100\\ sources}     & \makecell[c]{20\\ sources}     & \makecell[c]{50\\ sources}     & \makecell[c]{100\\ sources}     \\ \cmidrule(r){1-1}\cmidrule(lr){2-4}\cmidrule(l){5-7}
BART$_\text{cb}$      & 3.4$\pm$.03          & 3.5$\pm$.01          & 3.5$\pm$.01           & 1.4$\pm$.06         & 1.5$\pm$.02         & 1.5$\pm$.01          \\
X-Learner$_\text{cb}$ & 3.3$\pm$.04          & \textbf{3.2$\pm$.01}          & 3.2$\pm$.01           & \textbf{1.1$\pm$.08}         & \textbf{1.2$\pm$.02}         & \textbf{1.2$\pm$.02}          \\
R-Learner$_\text{cb}$ & \textbf{3.2$\pm$.03}          & \textbf{3.1$\pm$.01}          & \textbf{3.1$\pm$.01}           & \textbf{.88$\pm$.07}         & \textbf{.88$\pm$.02}         & \textbf{.86$\pm$.01}          \\
OthoRF$_\text{cb}$    & 3.4$\pm$.03          & 3.4$\pm$.01          & 3.4$\pm$.01           & 1.2$\pm$.07         & \textbf{1.2$\pm$.02}         & 1.3$\pm$.01          \\
TARNet$_\text{cb}$    & 5.6$\pm$.04          & 5.6$\pm$.02          & 5.7$\pm$.02           & 2.7$\pm$.06         & 2.8$\pm$.02         & 2.8$\pm$.02          \\
CFR-wass$_\text{cb}$  & 5.4$\pm$.05          & 5.5$\pm$.02          & 5.5$\pm$.02           & 2.7$\pm$.05         & 2.7$\pm$.02         & 2.7$\pm$.02          \\
CFR-mmd$_\text{cb}$   & 5.4$\pm$.05          & 5.4$\pm$.02          & 5.5$\pm$.02           & 2.7$\pm$.05         & 2.7$\pm$.02         & 2.7$\pm$.02          \\
CEVAE$_\text{cb}$     & 3.4$\pm$.04          & 3.4$\pm$.02          & 3.3$\pm$.01           & 1.2$\pm$.06         & \textbf{1.2$\pm$.02}         & \textbf{1.2$\pm$.01}       \\\cmidrule(r){1-1}\cmidrule(lr){2-4}\cmidrule(l){5-7}
FedCI           & \textbf{3.2$\pm$.03}          & \textbf{3.2$\pm$.02}          & \textbf{3.0$\pm$.01}           & 1.2$\pm$.07         & \textbf{1.2$\pm$.01}         & \textbf{1.2$\pm$.01} \\\cmidrule(r){1-1}\cmidrule(lr){2-4}\cmidrule(l){5-7}
CausalRFF                & \textbf{1.8$\pm$.03}          & \textbf{1.7$\pm$.03}          & \textbf{1.6$\pm$.01}           & \textbf{.24$\pm$.04}         & \textbf{.19$\pm$.14}         & \textbf{.15$\pm$.01}      \\ \bottomrule
\end{tabular}
\end{table}

This dataset is described in Section~\ref{sec:ihdp}. 
It was `de-randomized' by removing from the treated set children with non-white mothers. 
We use 10 replicates of the dataset in this experiment. 
For each replicate, we  divide into three sources, each consists of 249 data points. For each source, we use the first 50 data points for training, the next 100 for testing and the rest 99 for validating. We report the mean and standard error of the evaluation metrics over 10 replicates of the data.

\begin{table}

\caption{Out-of-sample errors on IHDP dataset. The dashes (-) in `$\text{ag}$' indicate that the numbers are the same as that of `$\text{cb}$'.
} \label{tb:data-ihdp}
\vspace{6pt}
\setlength{\tabcolsep}{12pt}
\scriptsize
\centering
\begin{tabular}{@{}lcccccc@{}}
\toprule
\multirow{2}{*}{Method} & \multicolumn{3}{c}{The error of CATE, $\sqrt{\epsilon_{\text{PEHE}}}$} & \multicolumn{3}{c}{The error of ATE, $\epsilon_{\text{ATE}}$} \\ \cmidrule(lr){2-4}\cmidrule(l){5-7} 
& 1 source      & 2 sources      & 3 sources     & 1 source     & 2 sources     & 3 sources     \\ \cmidrule(r){1-1}\cmidrule(lr){2-4}\cmidrule(l){5-7}
BART$_\text{ag}$       & -             &  2.3$\pm$.26             &  2.4$\pm$.22             & -            & 1.2$\pm$.23              &  1.3$\pm$.18             \\
X-Learner$_\text{ag}$  & -             & \textbf{1.8$\pm$.20}      & 1.8$\pm$.22      & -            & \textbf{0.6$\pm$.15}      & \textbf{0.4$\pm$.11}      \\
R-Learner$_\text{ag}$  & -             & 2.4$\pm$.31      & 2.3$\pm$.21      & -            & 1.3$\pm$.34      & 1.2$\pm$.24      \\
OthoRF$_\text{ag}$     & -             & 2.3$\pm$.21      & 2.1$\pm$.16      & -            & 0.6$\pm$.22      & 0.7$\pm$.13      \\
TARNet$_\text{ag}$          & -       & 2.9$\pm$.13      & 2.7$\pm$.15      & -     & \textbf{0.7$\pm$.12}      & 0.7$\pm$.16      \\
CFR-wass$_\text{ag}$          & -      & 2.3$\pm$.31      & 2.2$\pm$.20      & -     & \textbf{0.7$\pm$.12}      & 0.7$\pm$.11      \\
CFR-mmd$_\text{ag}$          & -      & 2.6$\pm$.21      & 2.4$\pm$.15      & -     & 0.8$\pm$.19      & 0.7$\pm$.18      \\
CEVAE$_\text{ag}$      & -             & 1.9$\pm$.14      & \textbf{1.6$\pm$.17}     & -            & 1.2$\pm$.11      & 0.8$\pm$.10      \\\cmidrule(r){1-1}\cmidrule(lr){2-4}\cmidrule(l){5-7}
BART$_\text{cb}$           & 2.2$\pm$.22      & 2.1$\pm$.26      & 2.1$\pm$.25      & 1.0$\pm$.16     & 0.8$\pm$.20      & 0.7$\pm$.17      \\
X-Learner$_\text{cb}$       & 1.9$\pm$.21      & 1.9$\pm$.21      & 1.8$\pm$.18      & \textbf{0.5$\pm$.21}     & \textbf{0.5$\pm$.18}      & \textbf{0.4$\pm$.11}      \\
R-Learner$_\text{cb}$       & 2.8$\pm$.31      & 2.6$\pm$.23      & 2.6$\pm$.17      & 1.6$\pm$.25     & 1.6$\pm$.26      & 1.6$\pm$.19      \\
OthoRF$_\text{cb}$          & 2.8$\pm$.16      & 2.1$\pm$.14      & 1.9$\pm$.14      & \textbf{0.8$\pm$.15}     & \textbf{0.6$\pm$.10}      & \textbf{0.6$\pm$.10}      \\
TARNet$_\text{cb}$          & 3.5$\pm$.59      & 2.7$\pm$.12      & 2.5$\pm$.15      & 1.6$\pm$.61     & \textbf{0.7$\pm$.12}      & \textbf{0.6$\pm$.17}      \\
CFR-wass$_\text{cb}$          & 2.2$\pm$.15      & 2.1$\pm$.22      & 2.1$\pm$.23      & \textbf{0.7$\pm$.23}     & \textbf{0.6$\pm$.18}      & \textbf{0.6$\pm$.16}      \\
CFR-mmd$_\text{cb}$          & 2.7$\pm$.19      & 2.3$\pm$.26      & 2.2$\pm$.10      & 0.9$\pm$.30     & \textbf{0.7$\pm$.17}      & \textbf{0.5$\pm$.17}      \\
CEVAE$_\text{cb}$           & \textbf{1.8$\pm$.22}      & 2.0$\pm$.11      & \textbf{1.7$\pm$.12}      & \textbf{0.5$\pm$.14}     & 1.4$\pm$.07      & 0.9$\pm$.07      \\\cmidrule(r){1-1}\cmidrule(lr){2-4}\cmidrule(l){5-7}
FedCI           & \textbf{1.6$\pm$.10} & \textbf{1.6$\pm$.12} & \textbf{1.7$\pm$.09} & \textbf{0.5$\pm$.10} & \textbf{0.5$\pm$.24} & \textbf{0.5$\pm$.09} \\\cmidrule(r){1-1}\cmidrule(lr){2-4}\cmidrule(l){5-7}
CausalRFF                & \textbf{1.7$\pm$.34}      & \textbf{1.4$\pm$.33}      & \textbf{1.2$\pm$.18}      & \textbf{0.7$\pm$.14}     & \textbf{0.7$\pm$.17}      &  \textbf{0.5$\pm$.16}      \\ \bottomrule
\end{tabular}
\end{table}

\noindent\textbf{Result and discussion.} Table~\ref{tb:data-ihdp} reports the experimental results on IHDP dataset. Again, we see that the proposed method gives competitive results compared to the baselines. In particular, the error of CausalRFF in predicting ATE is as low as that of the baselines, which is as we expected. In addition, the errors of CausalRFF in predicting CATE are lower than those of the baselines, which verifies the efficacy of the proposed method. Most importantly, CausalRFF is trained in a federated setting which minimizes the risk of privacy breach for the individuals stored in the local dataset.

\subsection{Summary}

We have proposed CausalRFF which learns causal effects from federated, observational data sources with dissimilar distributions. CausalRFF utilizes Random Fourier Features that naturally induce the decomposition of the loss function to individual components. CausalRFF allows for each component data group to inherit different distributions, and requires no prior knowledge on data discrepancy among the sources. We have also proved statistical guarantees which show how multiple data sources are effectively incorporated in our causal model.

One limitation of CausalRFF is that it might not directly work with incomplete data, i.e., there are missing values among data sources. In the next section, we introduce CausalFI that estimates causal effects from multiple data sources with missing values.

\section{Federated Causal Inference from Incomplete Data}
\label{sec:causalfi-incompletedata}

FedCI and CausalRFF proposed in Section~\ref{sec:fedci} and \ref{sec:causalrff} do not work directly in scenarios with missing data. In this section, we introduce CausalFI that estimates causal effects in a federated setting with incomplete confounders. 

In the following, we first describe the problem setting of federated causal inference from missing data. We then state necessary assumptions for identification of causal effects and propose CausalFI, a Bayesian federated estimator, that learns distributions of the causal estimands. 

\subsection{Problem Setting}
\label{sec:problem-setting}

We consider the federated setting as described in Section~\ref{sec:prob-form}. We have the following relations: $Y = (1-W)\cdot Y(0) + W\cdot Y(1)$. This is also known as the consistency assumption in causal inference, i.e., the factual outcome $Y$ equals to the potential outcome $Y(0)$ when $W=0$, and $Y(1)$ when $W=1$. We divide the confounders $Z$ to two sets of confounders $X$ and $U$, where $X$ is always observed and $U$ is partially observed, i.e., there are missing values in $U$. The number of dimensions of $U$ is $d$. Let $R = [R_1,\!...R_d]$ be vector of missing indicators for $Z$, i.e., $U_i$ is missing/unknown if $U_i=0$  and  $U_i$ is observed if $R_i=1$.

The observed data sources are a bit different from the description in Section~\ref{sec:prob-form} since we have two sets of confounders $X$ and $U$ and the missing indicators. Suppose we have $m$ data sources, each locally curated and organized, denoted as $D^\mathsf{s} = \{( w_i^\mathsf{s}, y_i^\mathsf{s}, x_i^\mathsf{s}, u_i^\mathsf{s}, r_i^\mathsf{s})\}_{i=1}^{n_\mathsf{s}}$, where $\mathsf{s} \in [m]$ (i.e., $\mathsf{s} \in \{1,\!...,m\}$). Here, $w_i^\mathsf{s}$, $y_i^\mathsf{s}$, $x_i^\mathsf{s}$, $u_i^\mathsf{s}$, and $r_i^\mathsf{s}$ represent realization samples of the treatment assignment, observed outcome, observed confounders,  partially observed confounders, and missing indicator for individual $i$ in source $\mathsf{s}$, i.e., they are realization values of the random variables $W$, $Y$, $X$, $U$, and $R$. We aim to utilize all $m$ data sources to learn a causal model without combining or sharing raw data among the sources, and at the same time learning higher order statistic of the causal estimands. 
In the subsequent section, we state some necessary assumptions required to estimate causal effects in a federated setting with missing data.

\subsection{Causal Quantities of Interest \& Assumptions}

Let $r$ be a specific missing pattern, i.e.,  it is a realization value of the random vector variable $R$. We further denote $U_r$ and  $U_{\tilde{r}}$ as sets of observed and missing variables in $U$, i.e., $U = \{U_r, U_{\tilde{r}}\}$. 
We would like to estimate conditional average treatment effect (CATE) and average treatment effect (ATE) which are defined as follows:
\begin{align}
    \tau(x,u_r) = E\big[Y(1) - Y(0)|X=x,U_r=u_r\big], &&\tau &= E\big[Y(1) - Y(0)\big], \label{eq:cate-ate}
\end{align}
where $\tau(x,u_r)$ is the average treatment affect conditioned on the observed confounders $X=x$ and $U_r=u_r$. Herein, the confounder $u_{\tilde{r}}$ is missing (or unknown). 

To estimate causal effects from multiple observational data sources, we first make some assumptions:
Assumptions~\ref{assumption:ignorability}~and~\ref{assumption:sutva} would allow the causal quantities in Eq.~(\ref{eq:cate-ate}) to be written in terms of the observed outcome $Y$. However, since the confounder $Z$ is partially observed, further assumptions are required. 
\begin{assumption}[MCAR and MAR]
\label{assumption:missing}
($i$) Missing indicators might depend on the observed confounders, treatment assignments, and observed outcomes, and ($ii$) they are independent of the incomplete confounders given all the observed variables, i.e., $R_i \ind U_j | X, Y, W$ for all $i,j\in [d]$.
\end{assumption}
\begin{assumption}
\label{assumption:missing-pattern}
$\text{Pr}(R_i = 1 | X, Y, W)> c$ for $i\in[d]$.\end{assumption}

Assumptions~\ref{assumption:ignorability}--\ref{assumption:missing-pattern} would allow the causal effects to be estimated from missing observational data.

\subsection{Identification of Causal Effects with Missing Observational Data}
\label{sec:identification}
In this section, we discuss identifiability of $\tau$ and $\tau(x,u_r)$. From Eq.~(\ref{eq:cate-ate}), we would need to estimate $\psi_w  :=E[Y(w)]$ and $\psi_w(x,u_r)  :=E[Y(w)|X=x,U_r=u_r]$, for $w\in\{0,1\}$. Hence, we have:
\begin{align*}
    \tau(x,z_r) &= \psi_1(x,u_r) - \psi_0(x,u_r),  &&\tau = \psi_1 - \psi_0.
\end{align*}
Under Assumptions~\ref{assumption:ignorability}~and~\ref{assumption:sutva}, we can rewrite $\psi_w$ and $\psi_w(x,z_r)$ as follows: 
\begin{align}
\psi_w(x,u_r) &= E_{Z_{\tilde{r}}}\big[E_Y[Y|W=w,X=x,U_r=u_r,U_{\tilde{r}}]\big],\label{eq:phi-w-x-z}\\\psi_w &= E_{X,U_r}\big[\psi_w(X,U_r)\big],\label{eq:phi-w}
\end{align}
where the inner expectation of $\psi_w(x,u_r)$ is taken over $p(y|w,x,z_r,u_{\tilde{r}})=p(y|w,x,u)$, and the outer expectation is taken over $p(u_{\tilde{r}}|x,u_r)$. Details of Eq.~(\ref{eq:phi-w-x-z})~and~(\ref{eq:phi-w}) is presented in Appendix~\ref{ap:causalfi-psi}.

\noindent\textbf{Imputation might be a biased estimator:} Now we show that using imputation and then apply a typical causal inference method might introduce bias. Let $u_{\tilde{r}}^\ast$ be imputed value of $u_{\tilde{r}}$. Then, the estimate of $\psi_w(x,u_r)$ is:
\begin{align}
\hat{\psi}_w(x,z_r,u_r^\ast) = E_Y\big[Y|W=w,X=x,U_r=u_r,U_{\tilde{r}}=u_{\tilde{r}}^\ast\big].
\end{align}
Generally, our objective is to ensure that $u_r^\ast$ satisfies the condition $\hat{\psi}_w(x,u_r,u_r^\ast) = \psi_w(x,u_r)$. However, achieving this alignment might pose certain challenges. \begin{remark}\label{lem:1}
    Let $u_{\tilde{r}}^\ast$ be imputed value by regression, and let $g(u_{\tilde{r}})=E[Y|W=w,X=x,U_r=u_r,U_{\tilde{r}}=u_{\tilde{r}}]$. We have:
\begin{enumerate}[label=(\roman*)]
        \item If $g(u_{\tilde{r}})$ is  non-linear, then $
        \psi_w(x,u_r) \neq \hat{\psi}_w(x,u_r,u_r^\ast)$.
    \item If $g(u_{\tilde{r}})$ is linear, then $
        \psi_w(x,u_r) = \hat{\psi}_w(x,u_r,u_r^\ast)$.
    \end{enumerate}
\end{remark}
Remark~\ref{lem:1} arises from the fact that $\psi_w(x,u_r) = E_{U_{\tilde{r}}}\big[E_Y[Y|W=w,X=x,U_r=u_r,U_{\tilde{r}}]\big] =E_{U_{\tilde{r}}}[g(U_{\tilde{r}})\big]$, 
where the expectation is taken over $p(u_{\tilde{r}}|x,u_r)$. Additionally, we have that
$\hat{\psi}_w(x,u_r,u_r^\ast) = E_Y[Y|W=w,X=x,U_r=u_r,U_{\tilde{r}}=u_{\tilde{r}}^\ast] = g(u_{\tilde{r}}^\ast)$. Since $u_{\tilde{r}}^\ast$ is imputed by regression on $(x,u_r)$, we have that $u_{\tilde{r}}^\ast = E[U_{\tilde{r}}|X=x,U_r=U_r]$. Hence, $\hat{\psi}_w(x,u_r,u_r^\ast) = g(E_{U_{\tilde{r}}}[U_{\tilde{r}}])$. This leads to ($i$) and ($ii$) in Remark~\ref{lem:1}

Remark~\ref{lem:1} indicates that constructing a regression-based imputation approach to forecast a missing value $u_{\tilde{r}}^\ast$, while conditioning on observed variables $X=x$ and $U_z=u_r$, is only viable when the outcome is a \emph{linear} function of the confounders. It could potentially introduce bias into causal estimations when the outcome is a non-linear function of the confounders containing missing values. 
Yet, in practical scenarios, the inherent nature of the function $g(u_{\tilde{r}})$ remains uncertain--whether it adheres to linearity or exhibits non-linear characteristics.

\noindent\textbf{The proposed procedure:} We now explain our approach where we learn a distribution of the missing values rather than a fixed point. To estimate $\psi_w(x,u_r)$ and $\psi_w$, we need to draw samples of $U_{\tilde{r}}$ from
\begin{align}
    p(z_{\tilde{r}}|x,u_r)\propto p(u|x) \label{eq:prop-z}
\end{align}
and samples of $Y$ from $p(y|w,x,u)$. 
Hence, the key is to learn $p(y|w,x,u)$ and $p(u|x)$ from the multiple data sources. However, learning these two distributions is challenging since $Z$ is missing according to the indicator vector $R$. Only using \emph{observed records} in the dataset to learn these distributions is biased since $p(y|w,x,u)\neq p(y|r=1,w,x,u)$ and $p(u|x)\neq p(u|r=1,x)$. So we need to take into account the missing data while performing inference. We have that
\begin{align}
    &p(y|w,x,u) \propto p(u|y,w,x) p(y,w,x), \\
    &p(u|x) =\sum_{w}\int p(u|y,w,x) p(y,w,x)dy.\label{eq:cond-dist}
\end{align}
Eq.~(\ref{eq:cond-dist}) implies that $p(y|w,x,u)$ and $p(u|x)$ are identifiable if $p(u|y,w,x)$ and $p(y,w,x)$ are identifiable. Since $Y$, $W$, and $X$ are all fully observed variables, $p(y,w,x)$ is identifiable. From Assumption~\ref{assumption:missing}, we have that $p(u|y,w,x)=p(u|r=1,y,w,x)$, which enable the learning of $p(u|y,w,x)$ from the \emph{observed records data}. Assumption~\ref{assumption:missing-pattern} ensures that there exists observed records in the multiple sources data for performing inference. 
Once $p(u|y,w,x)$ and $p(y|w,x)$ are identified, a naive approach to estimate $\psi_w$ and $\psi_w(x,u_r)$ is to draw samples of $U_{\tilde{r}}$ and $Y$ from $p(u_{\tilde{r}}|x,u_r)$ and $p(y|w,x,u)$ using Markov Chain Monte Carlo or inverse transform sampling depending on whether the variables are continuous or discrete. Although this is a possible approach, drawing samples of $U_{\tilde{r}}$ and $Y$ using the relation in Eq.~(\ref{eq:prop-z})~and~(\ref{eq:cond-dist}) might face a numerical issue since we need to integrate and sum over $y$ and $w$. Hence, we propose two surrogate $\hat{p}(u_{\tilde{r}}|x,u_r)$ and $\hat{p}(y|w,x,u)$ that are learned from the \emph{pseudo data points of $U$ drawn from $p(u|y,w,x)$}. To obtain the pseudo points, we can use forward sampling by drawing samples $(y,w,x)$ from the observed dataset and then using them as given values in $p(u|y,w,x)$ to draw samples for $U$.

To summarize, we would need to learn $p(u|y,w,x)$, and the two surrogate $\hat{p}(u_{\tilde{r}}|x,u_r)$, $\hat{p}(y|w,x,u)$. 
In the subsequent section, we propose a federated learning approach to learn these distribution from multiple data sources.

\subsection{CausalFI: Federated Causal Inference from Incomplete Data}
\label{sec:causalfi}
In this section, present a Bayesian federated approach to estimate causal effects from multiple data sources.

\noindent \textbf{Learning $p(z|y,w,x)$:} We parameterize different model  depending on each dimension of $U$. Let $p(u;\Lambda)$ be probability density/mass function of $Z$, where $\Lambda$ is the set of parameters of the density. For example, if $p(u;\Lambda)$ is density of a Gaussian distribution, then $\Lambda$ is a set of the mean and variance. To model the conditional distribution $p(u|y,w,x)$, we set
\begin{align}
    \lambda := \lambda(\theta^\lambda) = wf_1^{\lambda}(y,x;\theta_1^{\lambda}) + (1-w)f_0^{\lambda}(y,x;\theta_0^{\lambda})
\end{align}
for $\lambda \in \Lambda$, where $\theta_0^{\lambda}$ and $\theta_1^{\lambda}$ be sets of parameters modelling $\lambda$ when $w=0$ and $w=1$, respectively. 

Let $\theta = \bigcup_{\lambda\in \Lambda}\{\theta_0^{\lambda}, \theta_1^{\lambda}\}$ be set of all parameters. To capture higher-order statistic of the causal estimands in a Bayesian setting, we learn the posterior distribution $p(\theta|\mathbf{y},\mathbf{w},\mathbf{x})$, where $\mathbf{y},\mathbf{w},\mathbf{x}$ are observational data from \emph{all} sources. This would required data to be collected to a central machine to compute the posterior, which violates the federated setting. To overcome this problem, we use a variational approximation which would learn a set of shared variational parameters among the sources and avoid sharing raw data among them. To proceed, we specify a variational posterior distribution $q_\phi(\theta)$ and learn its sets of parameters $\phi$ by maximizing the evidence lower bound:
\begin{align}
    J &= \sum_{\mathsf{s}=1}^m J^\mathsf{s},\label{eq:fed-obj}\\
    J^\mathsf{s} &= E_{\theta}\Big[\log p(\mathbf{u}_r^\mathsf{s}|\mathbf{y}^\mathsf{s},\mathbf{w}^\mathsf{s},\mathbf{x}^\mathsf{s};\theta)\Big] - \frac{1}{m}D_{\text{KL}}[q_\phi(\theta)\|p(\theta)],\nonumber
\end{align}
where $\mathbf{u}_r^\mathsf{s}, \mathbf{y}^\mathsf{s},\mathbf{w}^\mathsf{s},\mathbf{x}^\mathsf{s}$ are observational data in source $\mathsf{s}$, and $p(\theta)$ is a prior distribution. 
The objective function $J$ is decomposed to multiple components $J^\mathsf{s}$, each associated with a data sources. So we can maximize it in a federated setting by aggregating either the local models or the gradients. Hence we can optimize it using Algorithm~\ref{algo:federated-training}.  

The conditional log-likelihood $\log p(\mathbf{u}_r^\mathsf{s}|\mathbf{y}^\mathsf{s},\mathbf{w}^\mathsf{s},\mathbf{x}^\mathsf{s};\theta)$ depends on the domain of $Z$, e.g., for continuous variable, the conditional log-likelihood is in the form of squared error; for binary or categorical, it is cross-entropy; for count variable, it is log of probability mass function of Poisson distribution or Negative binomial distribution. 
We learn $p(u|y,w,x)$ by maximizing the objective functions in the form of Eq.~(\ref{eq:fed-obj}). We obtain its variational posterior denoted by $q_{\phi_u}(\theta_u)$. The variational posterior would enable calculating higher-order statistic of the causal estimands.

\noindent\textbf{Learning $\hat{p}(u_{\tilde{r}}|x,u_r)$ and $\hat{p}(y|w,x,u)$}:
To estimate CATE, it is crucial to sample the missing confounder $u_{\tilde{r}}$ from $p(u_{\tilde{r}}|x,u_r)$ and then sample the outcome from $p(y|w,x,u)$. This can be achieved by utilizing Markov Chain Monte Carlo (MCMC) methods. However, this is not a wise solution 
since identification of $p(u_{\tilde{r}}|x,u_r)$ requires marginalization over $w$ and $y$ as shown in Eq.~(\ref{eq:prop-z})~and~(\ref{eq:cond-dist}), and there is no analytical solution for this marginalization. We overcome this problem by proposing surrogate models $\hat{p}(u_{\tilde{r}}|x,u_r;\theta_u)$, $\hat{p}(y|w,x,u;\theta_y)$, and we learn variational distributions  $q_{\hat{\phi}}(\theta_u)$, $q_{\hat{\phi}}(\theta_y)$, in federated setting. This model is learned from the pseudo data points of $u$ generated from $p(u|y,w,x)$ through Eq.~(\ref{eq:cond-dist}) with forward sampling. The loss functions for learning these two distributions are similar to that of Eq.~(\ref{eq:fed-obj}). We provide details of the loss functions and model structures in Appendix~\ref{ap:causalfi-loss}.

{\setlength{\textfloatsep}{0pt}\begin{algorithm}[tb]
\caption{Federated calculation of ATE and CATE}
\label{alg:ate-cate-global}
\SetKwInOut{Input}{Input}
\Input{$D^{'\mathsf{s}}\!=\!\{(x_i^{'\mathsf{s}},u_i^{'\mathsf{s}},r_i^{'\mathsf{s}})\!:\!i\!\in\! [n_\mathsf{s}']\}$ for $\mathsf{s}\in[m']$.}

\textbf{In the server machine:}

\Begin{
Draw $\{\theta_u^{(k)}\}_{k=1}^K \sim q_{\hat{\phi}}(\theta_u)$ and $\{\theta_y^{(k)}\}_{k=1}^K \sim q_{\hat{\phi}}(\theta_y)$\;
Send $\{(\theta_u^{(k)}, \theta_y^{(k)})\}_{k=1}^K$ to all sources\;
}

\textbf{In each source machine $\mathsf{s} \in [m']$:}

\Begin{
Call Algorithm~\ref{alg:ate-cate-local-source} to obtain $\mathcal{A}_1^\mathsf{s} \!\vcentcolon=\!\{\widehat{\text{ATE}}_k^\mathsf{s}\!:\!k\!\in\![K]\}$,  and $\mathcal{A}_2^\mathsf{s} \vcentcolon= \{\widehat{\text{CATE}}_{ik}\!:\!i\!\in\![n_\mathsf{s}'],k\!\in\![K]\}$\;
Compute CATE \& its higher-order statistics for each individual $i$ in source $\mathsf{s}$ using $\mathcal{A}_2^\mathsf{s}$\;
Send $\mathcal{A}_1^\mathsf{s}$ to the server machine\;
}

\textbf{In the server machine:}

\Begin{
Compute $\widehat{\text{ATE}}_k = (\sum_{\mathsf{s}=1}^{m'}n_\mathsf{s}'\widehat{\text{ATE}}_k^\mathsf{s})/\sum_{\mathsf{s}=1}^{m'}n_\mathsf{s}'$\;
Compute ATE and its higher-order statistics from $\{\widehat{\text{ATE}}_k:k\in[K]\}$\;
}
\end{algorithm}
}

{\setlength{\textfloatsep}{0pt}\begin{algorithm}[tb]
\caption{Samples of ATE and CATE on local source $\mathsf{s}$}
\label{alg:ate-cate-local-source}
\SetKwInOut{Input}{Input}
\SetKwInOut{Output}{Output}
\Input{$D^{'\mathsf{s}}=\{(x_i^{'\mathsf{s}},u_i^{'\mathsf{s}},r_i^{'\mathsf{s}}):i \in [n_\mathsf{s}']\}$, $\{(\theta_z^{(k)}, \theta_y^{(k)}): k\in[K]\}$.}
\Output{Samples of ATE$^\mathsf{s}$ and CATE for each $(\theta_z^{(k)}, \theta_y^{(k)})$.}

\For{$k = 1$ to $K$}{

\For{$i=1$ to $n_\mathsf{s}'$}{
Let $(x',u',r') = (x_i^{'\mathsf{s}},u_i^{'\mathsf{s}},r_i^{'\mathsf{s}})$\; Draw $\{u'_{\tilde{r'}}(j)\}_{j=1}^N \!\sim\! \hat{p}(u'_{\tilde{r'}}|x',u'_{r'};\theta_u^{(k)})$\;

\For{$j = 1$ to $N$}{
Let $u'_j = (u'_{r'}, u'_{\tilde{r'}}(j))$\;
Draw $\{y'_{0jl}\}_{l=1}^{M} \sim \hat{p}(y'|w=0,x',u'_j;\theta_y^{(k)})$\;

Draw $\{y'_{1jl}\}_{l=1}^{M} \sim \hat{p}(y'|w=1,x',u'_j;\theta_y^{(k)})$\;
}
$\widehat{\text{CATE}}_{ik} \!=\! \sum_{j=1}^N\sum_{l=1}^M(y'_{1jl} - y'_{0jl})/(MN)$\;
}
$\widehat{\text{ATE}}_k^\mathsf{s} = \sum_{i=1}^{n_\mathsf{s}'}\widehat{\text{CATE}}_{ik}/n_{\mathsf{s}}'$\;
}
\textbf{return} $\{\widehat{\text{ATE}}_k^\mathsf{s}\!:\!k\!\in\![K]\}$, $\{\widehat{\text{CATE}}_{ik}\!:\!i\!\in\![n_\mathsf{s}'],k\!\in\![K]\}$\;
\end{algorithm}

}

\subsection{Estimating Causal Effects}
Given a set of new data points in $m'$ sources ($m'\le m)$: $D^{'\mathsf{s}}=\{(x_i^{'\mathsf{s}},u_i^{'\mathsf{s}},r_i^{'\mathsf{s}}):i\in [n_\mathsf{s}']\}$, where $\mathsf{s}\in[m']$ and  $u_i^{'\mathsf{s}}$s might contain missing values, we would like to estimate CATE and ATE. To compute these causal quantities of interest and their higher-order statistics, we draw $K$ samples of $\theta_u$ and $\theta_y$ from their variational posterior distributions, denoted as $\{(\theta_u^{(k)}, \theta_y^{(k)}) : k\in[K]\}$. Each sample $(\theta_u^{(k)}, \theta_y^{(k)})$ is used to compute one sample for ATE and CATE, and there are $K$ samples. We can use these $K$ samples of ATE and CATE to compute their mean, variance, skewness, kurtosis, etc. For each sample $(\theta_u^{(k)}, \theta_y^{(k)})$, we empirically calculate $\psi_w(x',u'_{r'})$ and $\psi_w$ using Eq.~(\ref{eq:phi-w-x-z})~and~(\ref{eq:phi-w}). To proceed, for each given data points $(x',u',r') \in D'$, we draw samples of $U_{\tilde{r}}$ from $\hat{p}(u'_{\tilde{r'}}|x',u'_{r'};\theta_u^{(k)})$. We then substitute the above samples to  $\hat{p}(y'|w',x',u';\theta_y^{(k)})$ to draw samples of $Y$ and calculate its empirical expectation. The empirical $\psi_w$ can be calculated using Eq.~(\ref{eq:phi-w}) by averaging $\psi_w(x',u'_{r'})$ over all  data points in $\bigcup_{\mathsf{s}=1}^{m'}D^{'\mathsf{s}}$. We summarise the steps of federated estimating causal effects in Algorithm~\ref{alg:ate-cate-global}~and~\ref{alg:ate-cate-local-source}.

\subsection{Missing Mechanism Testing}

This work assumes MCAR and MAR, and these missing mechanisms are testable. The Little's test \citep{little1988test} is the well-known test for MCAR. The test is available in software packages like SPSS, SAS, and R. Testing for MAR is more complex as it implies that the missingness is related to the observed data but not the unobserved data. There is no single definitive test for MAR, but several approaches can be used to assess the assumption such as \citet{molenberghs2007missing,little2019statistical,enders2022applied}. A drawback of existing missing mechanism tests is their inability to assess missing data across multiple data sources in a federated setting. To address this limitation, we present a straightforward approach that involves conducting individual tests on each data source and subsequently determining the prevailing missing data mechanism through a voting method. This allows us to account for the distinct characteristics of each data source while making a collective inference about the missing mechanism. For future research, an interesting topic is to develop federated tests for missing data mechanism.

\subsection{Experiments}

\label{sec:experiment-causalfi}

\textbf{Baselines \& the aims of experiments.} In this section, we first study the performance of CausalFI compared with the cases of bootstrap aggregating and on combined data. We also study its performance on different portion of missing data. We then compare CausalFI with the existing baselines including: 
BART \citep{hill2011bayesian}, TARNet \citep{shalit2017estimating}, CFR-Wass (CFRNet with Wasserstein distance) \citep{shalit2017estimating}, CFR-MMD (CFRNet with maximum mean discrepancy distance) \citep{shalit2017estimating}, CEVAE \citep{louizos2017causal}, 
OrthoRF \citep{oprescu2019orthogonal}, 
IPW \citep{seaman2014inverse}, 
DR \citep{mayer2020doubly}, X-learner \citep{kunzel2019metalearners}, R-learner \citep{nie2021quasi}, and FedCI (Section~\ref{sec:fedci}) \citep{vo2022bayesian}.  
Note that these methods, except IPW and DR, do not do not directly deal with missing data. Hence, we first impute the missing data and then fit the baselines. We impute with two popular methods: probabilistic principal component analysis ($\texttt{ppca}$) \citep{tipping1999probabilistic}, and multivariate imputation by chained equations ($\texttt{mice}$) \citep{van2011mice}. 
In addition, for most of the baselines that are not federated methods (BART, TARNet, CFR-wass, CFR-mmd, CEVAE, OrthorRF, X-learner, R-learner), we would train them in two cases: 
($i$) using bootstrap aggregating of \citet{breiman1996bagging} where $m$ models are trained separately on each source data and then averaging the predicted treatment effects based on each trained model (denoted as \texttt{avg}), ($ii$) training a global model with the combined data from all the sources (denoted as \texttt{com}). Note that case ($ii$) might breach sensitive information, it is only used for the purposes of comparison.
In summary, there are four different cases: \texttt{ppca+avg}, \texttt{ppca+com}, \texttt{mice+avg}, and \texttt{mice+com}.

\noindent\textbf{Implementation of the baselines.} We reuse source code of the baselines which are publicly available. We train CEVAE using implementation from \citet{louizos2017causal}. Implementation of TARNet and CFR are from \citet{shalit2017estimating}. For these methods, we use Exponential Linear Unit (ELU) activation function and fine-tune the number of nodes in each hidden later from 10 to 200 with step size of addition by 10. We use package \texttt{BartPy} for BART, and package \texttt{causalml} \citep{chen2020causalml} for X-learner and R-learner. Experiments on OrthoRF is with the package \texttt{econml} \citep*{econml}. Implementation of IPW and DR is from \citet{mayer2020doubly}. Finally, implementation of FedCI is  from \citet{vo2022bayesian}. 
The learning rate is fine-tuned from $\{10^{-5}, 10^{-4}, 10^{-3}, 10^{-2}, 10^{-1}\}$ for all methods. For TARNet, CFR-wass, CFR-mmd, OrthoRF, X-learner, R-learner, we additionally fine-tune the regularizer factors in $\{10^{-5}, 10^{-4}, 10^{-3}, 10^{-2}, 10^{-1}, 10^{0}\}$. 
Two error metrics are used to compare the methods: $\epsilon_\mathrm{PEHE}$ (precision in estimation of heterogeneous effects) and $\epsilon_\mathrm{ATE}$ (absolute error) to compare the methods: $\epsilon_\mathrm{PEHE} = \sum_{i=1}^n(\tau(x_i,z_{ri}) - \hat{\tau}(x_i,z_{ri}))^2/n, \epsilon_\mathrm{ATE} = |\tau - \hat{\tau}|$, 
 where $\uptau(x_i,z_{ri}), \uptau$ are the ground truth of CATE and ATE, and $\hat{\tau}(x_i,z_{ri}), \hat{\tau}$ are their estimates. 

For comparing the point estimation, we report the mean and standard error of the error metrics over 10 replicates of the data. 

\subsubsection{Synthetic Data}

\noindent To evaluate the performance of CausalFI, we simulate variables with a ground truth model and divide them into multiple sources for federated learning. We use the following ground truth distributions to simulate the data:
\begin{align*}
    \mathbf{z}_i &\sim \mathsf{N}(\bm{m}, \bm{\Sigma}),\\
    w_i &\sim \mathsf{Bern}(\mathsf{sig}(a_0 + \bm{b}_1^\top\bm{u}_i)),\\
    y_i(0) &\sim \mathsf{N}(\mathsf{sp}(c_0 + \bm{c}_1^\top \bm{u}_i), \sigma_0^2),\\
    y_i(1) &\sim \mathsf{N}(\mathsf{sp}(d_0 + \bm{d}_1^\top \bm{u}_i), \sigma_1^2),
\end{align*}
where $\mathsf{N}(\cdot)$, $\mathsf{Bern}(\cdot)$ denote the Gaussian and Bernoulli distributions, $\mathsf{sig}(\cdot)$ denotes the sigmoid function and  $\mathsf{sp}(\cdot)$ is the softplus function. We only keep $y_i=w_iy_i(1) + (1-w_i)y_i(0)$ as the observed outcome. The vector $\mathbf{z}_i$ contains all confounders and we split it into two sets $u_i$ and $x_i$. $u_i$ is a vectors that might contain missing values, and $x_i$ is always observed. We simulate missing indicators for $u_i$ using:
\begin{align}
    r_{ji} \sim \mathsf{Bern}(\mathsf{sig}(e_{j0} + e_{j1}w_i + e_{j2}y_i + \bm{e}_{j3}^\top x_i)), \label{eq:sim-missing-indicator}
\end{align}
for each dimension $j$ of $z_i$. Herein, $e_{j0}$, $e_{j1}$, $e_{j2}$, $\bm{e}_{j3}$ are the ground truth parameters and they are randomly set. In particular, we set $e_{j0}=5.0$ and randomly draw $e_{j1}, e_{j2} \sim \mathsf{U}[-2,0]$, $\bm{e}_{j3}\sim \mathsf{N}(\bm{0},\mathbf{M})$, where $\mathbf{M}=\mathbf{L} \mathbf{L}^\top$, $\mathbf{L} \in \mathbb{R}^{10\times 5}$ and each element $L_{ij} \sim \mathsf{U}[0,0.5]$. For each dataset, we simulate 10 replications, each has $10,000$ data points. We randomly divide the data into 50 sources, each has 200 data points. For each data source, we split it into three sets of 100, 50, 50 data points for training, testing, and validation. In Table~\ref{tb:data-stats-synthetic}, we report the percentage of missing values of each confounder in the synthetic data. 
In Table~\ref{tb:data-stats-ihdp}, we report the percentage of missing values of each confounder in IHDP data.

\begin{table}\centering
\caption{Percentage of missing entries in each incomplete confounder on synthetic data.
}
 \vspace{6pt}
\label{tb:data-stats-synthetic}
\setlength{\tabcolsep}{9pt}
\scriptsize
\begin{tabular}{@{}lcccccccccc@{}}
\toprule
Replicate &   $z_1$ &   $z_2$ &   $z_3$ &   $z_4$ &   $z_5$ &   $z_6$ &   $z_7$ &   $z_8$ &   $z_9$ &  $z_{10}$ \\\midrule
       \#1 &  30\% &  44\% &  25\% &  49\% &  52 &  24\% &  62\% &  33\% &  49\% &   37\% \\
       \#2 &  72\% &  39\% &  76\% &  81\% &  37\% &  74\% &  85\% &  31\% &  31\% &   31\% \\
       \#3 &  76\% &  37\% &  41\% &  57\% &  40\% &  55\% &  78\% &  37\% &  33\% &   35\% \\
       \#4 &  65\% &  51\% &  32\% &  57\% &  52\% &  53\% &  79\% &  56\% &  44\% &   48\% \\
       \#5 &  38\% &  44\% &  27\% &  54\% &  46\% &  37\% &  69\% &  39\% &  38\% &   34\% \\
       \#6 &  56\% &  79\% &  46\% &  89\% &  73\% &  38\% &  90\% &  56\% &  64\% &   60\% \\
       \#7 &  69\% &  61\% &  33\% &  84\% &  63\% &  42\% &  77\% &  36\% &  51\% &   57\% \\
       \#8 &  48\% &  26\% &  55\% &  61\% &  28\% &  60\% &  65\% &  15\% &  24\% &   22\% \\
       \#9 &  61\% &  54\% &  38\% &  83\% &  54\% &  43\% &  72\% &  18\% &  44\% &   48\% \\
      \#10 &  61\% &  60\% &  31\% &  73\% &  63\% &  31\% &  77\% &  44\% &  56\% &   55\% \\\bottomrule
    \end{tabular}
\end{table}

\begin{table}\centering
\caption{Percentage of missing entries in each incomplete confounder on IHDP data.
}
 \vspace{6pt}
\label{tb:data-stats-ihdp}
\setlength{\tabcolsep}{9pt}
\scriptsize
\begin{tabular}{@{}lcccc@{}}
\toprule
Replicate &    $z_1$ &    $z_2$ &    $z_3$ &    $z_4$ \\
\midrule
       \#1 & 20.16\% & 20.16\% & 20.83\% & 20.56\% \\
       \#2 & 46.91\% & 46.64\% & 46.91\% & 44.09\% \\
       \#3 & 93.01\% & 93.28\% & 92.74\% & 94.09\% \\
       \#4 & 44.89\% & 42.61\% & 46.24\% & 45.43\% \\
       \#5 & 23.39\% & 26.08\% & 25.40\% & 24.46\% \\
       \#6 & 60.62\% & 60.35\% & 58.87\% & 57.12\% \\
       \#7 & 28.76\% & 31.45\% & 29.84\% & 29.03\% \\
       \#8 & 82.93\% & 83.20\% & 82.53\% & 83.60\% \\
       \#9 &  0.13\% &  0.67\% &  0.00\% &  0.27\% \\
      \#10 & 15.46\% & 17.34\% & 16.80\% & 17.34\% \\
\bottomrule
\end{tabular}
\end{table}

\begin{figure}\centering
    \includegraphics[width=0.65\textwidth]{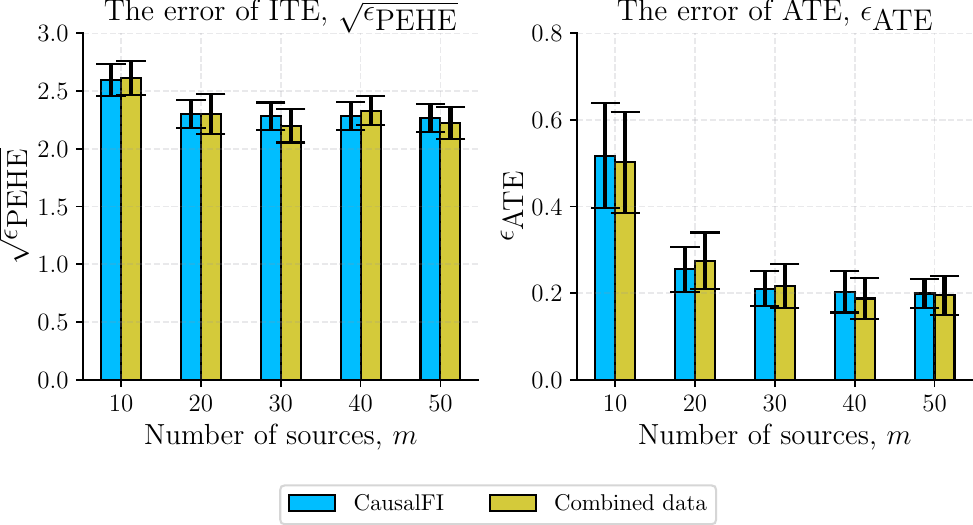}
\caption{Federated inference analysis on synthetic data.}
    \label{fig:compare-vs-combined-data}
\end{figure}

\noindent\textbf{CausalFI vs. training on combined data:} In this experiment, we study the performance of CausalFI compared with the case of combining data. The results in Figure~\ref{fig:compare-vs-combined-data} show that the performance of CausalFI is as good as training on combined data as expected. This verifies the efficacy of our proposed federated training method.

\begin{figure}\centering
    \includegraphics[width=0.65\textwidth]{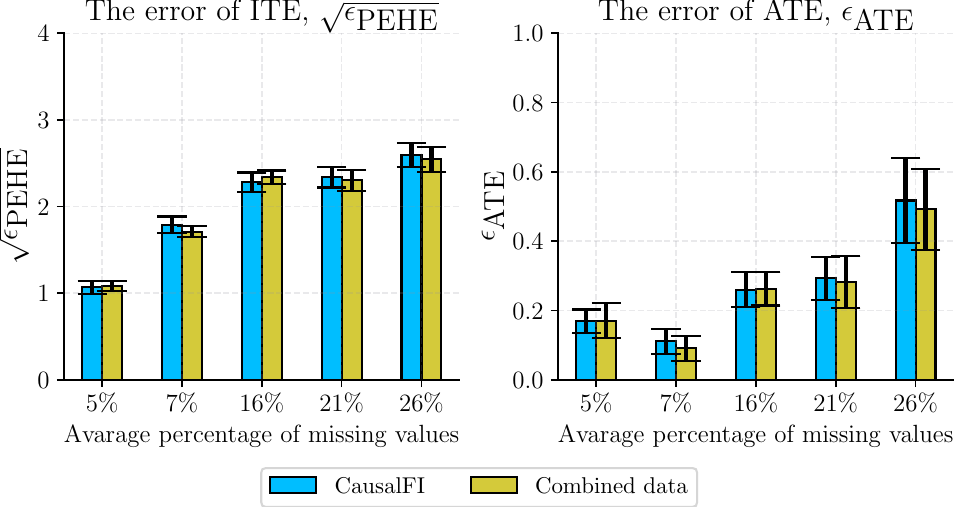}
\caption{The errors on different number of missing features.}
    \label{fig:compare-diff-num-missing-features}
\end{figure}

\begin{figure}\centering
    \includegraphics[width=0.65\textwidth]{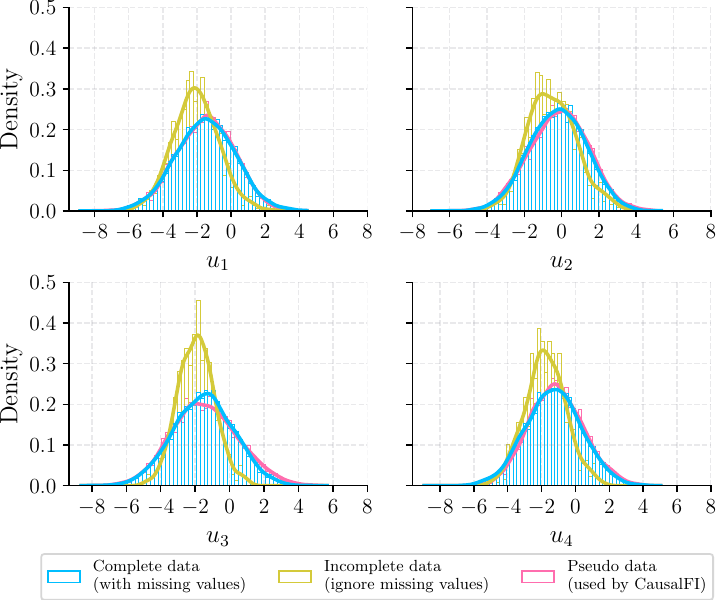}
\caption{Marginal distribution of $U_1,U_2, U_3, U_4$ pseudo data compared with those of complete and incomplete data.}
    \label{fig:hist-data}
\end{figure}

\noindent\textbf{Analysis on learning missing confounders:} This experiment aims to learn the performance of CausalFI on different number of missing features. We train the model when there are 2, 4, 6, 8 and 10 incomplete confounders, which are associated with an average of 5\%, 7\%, 16\%, 21\%, and 26\% missing entries. Herein, we also compare with the case of training on combined data.  Figure~\ref{fig:compare-diff-num-missing-features} shows that when there are more confounders that contain missing values, the errors are a bit higher. This result is expected as there are more missing values. Importantly, the errors of CausalFI are as low as those of training on combined data. In addition, we illustrate marginal distributions of the learned pseudo data compared with distributions of the complete data and incomplete data in Figure~\ref{fig:hist-data}. The figures show that distributions of the learned pseudo data (used by CausalFI) are close to those of complete data. This explains why CausalFI can recover distribution of the missing confounders and hence identifies the causal effects of interest.

\begin{table}\centering
\scriptsize
\caption{Out-of-sample errors of CATE and ATE on synthetic data (lower is better). Top-3 performances are highlighted in bold. 
Methods trained on combined data (com)  violate privacy constraint and they are only used for comparison purpose.
}
\vspace{6pt}
\label{tb:errors}
\setlength{\tabcolsep}{9pt}
\begin{tabular}{@{}lcccccc@{}}
\toprule
\multirow{2}{*}{Method} & \multicolumn{3}{c}{The error of CATE, $\sqrt{\epsilon_\text{ATE}}$}                                              & \multicolumn{3}{c}{The error of ATE, $\epsilon_\text{ATE}$}                                               \\ \cmidrule(l){2-4}\cmidrule(l){5-7} 
        & 10 sources & 30 sources & 50 sources & 10 sources & 30 sources & 50 sources \\\cmidrule(r){1-1}\cmidrule(lr){2-4}\cmidrule(l){5-7}
BART$_{\text{ppca+com}}$        & 7.47$\pm$.58 & 7.45$\pm$.56 & 7.23$\pm$.59 & 3.31$\pm$.61 & 3.34$\pm$.55 & 3.27$\pm$.56 \\
BART$_{\text{mice+com}}$  & 7.43$\pm$.57 & 7.43$\pm$.57 & 7.23$\pm$.57 & 3.22$\pm$.62 & 3.34$\pm$.60 & 3.41$\pm$.60 \\
BART$_{\text{ppca+avg}}$         & 7.52$\pm$.64 & 7.55$\pm$.53 & 7.43$\pm$.52 & 3.43$\pm$.52 & 3.51$\pm$.53 & 3.51$\pm$.61 \\
BART$_{\text{mice+avg}}$         & 7.49$\pm$.61 & 7.51$\pm$.52 & 7.35$\pm$.47 & 3.29$\pm$.61 & 3.49$\pm$.49 & 3.47$\pm$.58 \\
\cmidrule(r){1-1}\cmidrule(lr){2-4}\cmidrule(l){5-7}R-learner$_{\text{ppaca+com}}$   & 5.30$\pm$.27 & 4.65$\pm$.28 & 4.50$\pm$.31 & 1.12$\pm$.28 & 1.19$\pm$.18 & 1.08$\pm$.23 \\
R-learner$_{\text{mice+com}}$  & 5.70$\pm$.30 & 4.91$\pm$.41 & 4.53$\pm$.31 & 1.59$\pm$.33 & 1.53$\pm$.25 & 1.40$\pm$.23 \\
R-learner$_{\text{ppca+avg}}$  & 7.29$\pm$.59  & 7.17$\pm$.59 & 7.22$\pm$.58 & 2.62$\pm$.74 & 2.46$\pm$.74 & 2.49$\pm$.74 \\
R-learner$_{\text{ppca+avg}}$  & 7.52$\pm$.51  & 7.43$\pm$.50 & 7.37$\pm$.49 & 2.47$\pm$.72 & 2.43$\pm$.60 & 2.41$\pm$.72 \\
\cmidrule(r){1-1}\cmidrule(lr){2-4}\cmidrule(l){5-7}X-learner$_{\text{ppca+com}}$  & 5.79$\pm$.39 & 5.61$\pm$.37 & 5.91$\pm$.42 & 1.53$\pm$.31 & 1.70$\pm$.34 & 2.13$\pm$.34 \\
X-learner$_{\text{mice+com}}$  & 5.75$\pm$.35 & 6.19$\pm$.40 & 5.98$\pm$.35 & 2.02$\pm$.32 & 2.56$\pm$.29 & 2.52$\pm$.27 \\
X-learner$_{\text{ppca+avg}}$   & 5.84$\pm$.38 & 5.77$\pm$.38 & 5.81$\pm$.37 & 1.17$\pm$.29 & 1.18$\pm$.26 & 1.22$\pm$.27 \\
X-learner$_{\text{mice+avg}}$   & 5.92$\pm$.37 & 5.64$\pm$.41 & 5.72$\pm$.38 & 1.27$\pm$.21 & 1.21$\pm$.29 & 1.24$\pm$.25 \\
\cmidrule(r){1-1}\cmidrule(lr){2-4}\cmidrule(l){5-7}OthoRF$_{\text{ppca+com}}$      & 7.83$\pm$.64 & 7.72$\pm$.58 & 7.61$\pm$.57 & 3.50$\pm$.73 & 3.37$\pm$.63 & 2.32$\pm$.60 \\
OthoRF$_{\text{mice+com}}$  & 7.95$\pm$.58 & 7.69$\pm$.53 & 7.55$\pm$.52 & 3.34$\pm$.66 & 3.33$\pm$.58 & 3.32$\pm$.57 \\
OthoRF$_{\text{ppca+avg}}$      & 7.97$\pm$.76 & 7.55$\pm$.67 & 6.93$\pm$.62 & 3.80$\pm$.69 & 3.21$\pm$.65 & 2.89$\pm$.50 \\
OthoRF$_{\text{mice+avg}}$      & 8.05$\pm$.71 & 7.61$\pm$.60 & 6.99$\pm$.61 & 3.50$\pm$.71 & 3.32$\pm$.63 & 2.94$\pm$.52 \\
\cmidrule(r){1-1}\cmidrule(lr){2-4}\cmidrule(l){5-7}TARNet$_{\text{ppca+com}}$      & 3.63$\pm$.13 & 3.62$\pm$.15 & 3.63$\pm$.15& 1.38$\pm$.22 & 1.40$\pm$.32 & 1.35$\pm$.30           \\
TARNet$_{\text{mice+com}}$      & 3.52$\pm$.15 & 3.48$\pm$.17 & 3.50$\pm$.18 & 1.29$\pm$.25 & 1.34$\pm$.31 & 1.32$\pm$.30           \\
TARNet$_{\text{ppca+avg}}$      & 4.77$\pm$.48 & 4.51$\pm$.29 & 4.15$\pm$.61 & 1.80$\pm$.39 & 1.76$\pm$.43 & 1.61$\pm$.39            \\
TARNet$_{\text{ppca+avg}}$      & 4.61$\pm$.41 & 4.47$\pm$.31 & 4.01$\pm$.55 & 1.71$\pm$.42 & 1.61$\pm$.41 & 1.50$\pm$.37            \\
\cmidrule(r){1-1}\cmidrule(lr){2-4}\cmidrule(l){5-7}CFR-mmd$_{\text{ppca+com}}$  & 3.72$\pm$.21 & 3.72$\pm$.20 & 3.75$\pm$.22 & 0.83$\pm$.16 & 0.77$\pm$.15 & \textbf{0.74$\pm$.17}            \\
CFR-mmd$_{\text{mice+com}}$      & 3.59$\pm$.22 & 3.59$\pm$.20 & 3.64$\pm$.22 & 0.89$\pm$.26 & 0.89$\pm$.24 & 0.93$\pm$.27           \\
CFR-mmd$_{\text{ppca+avg}}$  & 4.52$\pm$.71 & 4.22$\pm$.54 & 4.11$\pm$.63& 1.75$\pm$.41 & 1.71$\pm$.42 & 1.50$\pm$.29            \\
CFR-mmd$_{\text{ppca+avg}}$  & 4.41$\pm$.60 & 4.11$\pm$.52 & 3.91$\pm$.61& 1.62$\pm$.42 & 1.50$\pm$.41 & 1.28$\pm$.34            \\
\cmidrule(r){1-1}\cmidrule(lr){2-4}\cmidrule(l){5-7}CFR-wass$_{\text{ppca+com}}$ & 3.77$\pm$.16 & 3.78$\pm$.16 & 3.77$\pm$.16& 1.32$\pm$.27 & 1.32$\pm$.27 & 1.30$\pm$.27           \\
CFR-wass$_{\text{mice+com}}$      & 3.95$\pm$.14 & 3.96$\pm$.13 & 3.96$\pm$.14 & 1.52$\pm$.29 & 1.52$\pm$.29 & 1.51$\pm$.28           \\
CFR-wass$_{\text{ppca+avg}}$ & 4.51$\pm$.32 & 4.12$\pm$.31 & 4.07$\pm$.25& 1.77$\pm$.51 & 1.68$\pm$.49 & 1.61$\pm$.39            \\
CFR-wass$_{\text{mice+avg}}$ & 4.82$\pm$.35 & 4.26$\pm$.37 & 4.11$\pm$.21& 1.65$\pm$.53 & 1.50$\pm$.42 & 1.53$\pm$.30            \\
\cmidrule(r){1-1}\cmidrule(lr){2-4}\cmidrule(l){5-7}CEVAE$_{\text{ppca+com}}$      & 3.99$\pm$.31 & 3.41$\pm$.30 & 3.48$\pm$.27 & 1.14$\pm$.31 & 0.75$\pm$.19 & 0.85$\pm$.30 \\
CEVAE$_{\text{mice+com}}$  &  3.09$\pm$.22 & 4.45$\pm$.99 & 3.99$\pm$.62 & \textbf{0.69$\pm$.27} & 1.37$\pm$.73 & 1.24$\pm$.48 \\
CEVAE$_{\text{ppca+avg}}$       & 4.35$\pm$.38 & 3.28$\pm$.29 & 3.34$\pm$.32 & 1.22$\pm$.45 & 0.97$\pm$.26 & 1.01$\pm$.34 \\
CEVAE$_{\text{mice+avg}}$       & 4.47$\pm$.40 & 4.59$\pm$.23 & 4.24$\pm$.32 & 1.01$\pm$.41 & 1.07$\pm$.21 & 1.18$\pm$.29 \\
\cmidrule(r){1-1}\cmidrule(lr){2-4}\cmidrule(l){5-7}FedCI$_{\text{ppca+com}}$           & \textbf{2.72$\pm$.13}           &  \textbf{2.52$\pm$.18}          &   \textbf{2.47$\pm$.17}         &  0.78$\pm$.10          &   \textbf{0.57$\pm$.11}         &  \textbf{0.59$\pm$.11}          \\
FedCI$_{\text{mice+com}}$           & \textbf{2.62$\pm$.13}           &  \textbf{2.54$\pm$.16}          &   \textbf{2.52$\pm$.17}         &  \textbf{0.62$\pm$.15}          &   \textbf{0.55$\pm$.15}         &  \textbf{0.52$\pm$.12}  \\
FedCI$_{\text{ppca+avg}}$           & 3.11$\pm$.17           &  3.04$\pm$.20          &   2.98$\pm$.19         &  0.89$\pm$.12          &   0.72$\pm$.10         &  0.69$\pm$.13          \\
FedCI$_{\text{mice+avg}}$           & 3.18$\pm$.13           &  3.10$\pm$.15          &   3.02$\pm$.16         &  0.85$\pm$.12          &   0.70$\pm$.14         &  0.65$\pm$.12   \\\cmidrule(r){1-1}\cmidrule(lr){2-4}\cmidrule(l){5-7}CausalFI             & \textbf{2.60$\pm$.14} & \textbf{2.28$\pm$.12} & \textbf{2.26$\pm$.12} & \textbf{0.52$\pm$.12} & \textbf{0.21$\pm$.04} & \textbf{0.20$\pm$.03}\\\bottomrule
\end{tabular}
\end{table}

\noindent\textbf{Compare with baselines:} 
As mentioned earlier, we have four settings for the baselines: 
\texttt{ppca+com}, \texttt{mice+com}, \texttt{ppca+avg},  \texttt{mice+avg}. 
For FedCI, since this is a federated method, we only use combined data and bootstrap aggregating when imputing missing values, and the training of FedCI is a federated setting. This experiment is with 10 incomplete confounders (26\% of missing values). 
The results in Table~\ref{tb:errors} show that CausalFI is among top-3 performance. It achieves lower errors compared to BART, R-learner, X-learner, OthoRF, TARNet, and CFR trained on combined data. In comparison with CEVAE$_{\text{ppca+com}}$, CEVAE$_{\text{mice+com}}$,  FedCI$_{\text{ppca+com}}$, and FedCI$_{\text{mice+com}}$, CausalFI achieves competitive results. However, imputation of the missing values in these four baselines require combining data, which violate federated data setting. In addition, we also observe that performance of the baselines depends on the impute method used. Especially in the case of CEVAE, each imputation method (\texttt{ppca} or \texttt{mice}) would result in a very different error of ATE. Meanwhile, CausalFI learns distributions of the missing confounders while training the model. 

For comparison with IPW and DR, these methods are designed for missing data, but they can only estimate \emph{in-sample} ATE, but not CATE. Hence, we compare with them separately in Table~\ref{tab:errors-vs-missing-methods}. The imputation used in these methods are \texttt{pca} (principal components analysis) and \texttt{mice}. The results show that CausalFI significantly outperform these baselines.

\begin{table}
    \centering
\caption{In-sample errors of ATE ($\epsilon_{\text{ATE}}$) on synthetic data.
}\label{tab:errors-vs-missing-methods}
 \vspace{6pt}
\scriptsize
\setlength{\tabcolsep}{12pt}
\begin{tabular}{@{}lccc@{}}
\toprule
Method         & 10 sources & 30 sources & 50 sources \\\cmidrule(r){1-1}\cmidrule(l){2-4}
IPW$_{\text{pca+com}}$    &  2.6$\pm$0.9      & 1.9$\pm$0.7       & 1.3$\pm$0.4       \\ 
IPW$_{\text{mice+com}}$ &  2.9$\pm$0.3      & 2.8$\pm$0.2       & 2.5$\pm$0.2       \\
IPW$_{\text{com}}$ &  3.0$\pm$0.4      & 2.8$\pm$0.3       & 2.6$\pm$0.2       \\ 
\cmidrule(r){1-1}\cmidrule(l){2-4}
DR$_{\text{pca+com}}$    &  2.8$\pm$0.8      & 2.4$\pm$0.6       & 2.2$\pm$0.6       \\ 
DR$_{\text{mice+com}}$ &  3.9$\pm$0.9      & 3.7$\pm$1.0       & 3.2$\pm$0.6       \\ 
DR$_{\text{com}}$ &  3.7$\pm$0.8      & 3.2$\pm$1.0       & 2.8$\pm$0.5       \\ 
\cmidrule(r){1-1}\cmidrule(l){2-4}
CausalFI &  0.3$\pm$.10      & 0.2$\pm$.05       & 0.2$\pm$.04       \\ \bottomrule
\end{tabular}
\end{table}

\begin{table}
    \centering
\caption{In-sample errors of ATE ($\epsilon_{\text{ATE}}$) on IHDP data.
}\label{tab:errors-vs-missing-methods-ihdp}
 \vspace{6pt}
\scriptsize
\setlength{\tabcolsep}{12pt}
\begin{tabular}{@{}lccc@{}}
\toprule
 Method        & 2 sources & \makecell[c]4 sources & 6 sources \\\cmidrule(r){1-1}\cmidrule(l){2-4}
IPW$_{\text{pca+com}}$    &  1.2$\pm$0.5      & 0.6$\pm$0.2       & 0.5$\pm$0.2       \\ 
IPW$_{\text{mice+com}}$ &  1.2$\pm$0.4      & 0.7$\pm$0.2       & 0.6$\pm$0.2       \\ 
IPW$_{\text{com}}$ &  0.9$\pm$0.2      & 0.4$\pm$0.2       & 0.4$\pm$0.1       \\ 
\cmidrule(r){1-1}\cmidrule(l){2-4}
DR$_{\text{pca+com}}$    &  0.8$\pm$0.3     & 0.7$\pm$0.3       & 0.3$\pm$0.1       \\ 
DR$_{\text{mice+com}}$ &  0.8$\pm$0.3      & 0.5$\pm$0.2       & 0.2$\pm$0.1       \\
DR$_{\text{com}}$ &  1.4$\pm$0.6      & 0.6$\pm$0.1       & 0.6$\pm$0.2       \\
\cmidrule(r){1-1}\cmidrule(l){2-4}
CausalFI &  1.1$\pm$0.3      & 0.6$\pm$0.1       & 0.3$\pm$0.1       \\ \bottomrule
\end{tabular}
\end{table}

\begin{figure}
\centering
    \includegraphics[width=0.7\textwidth]{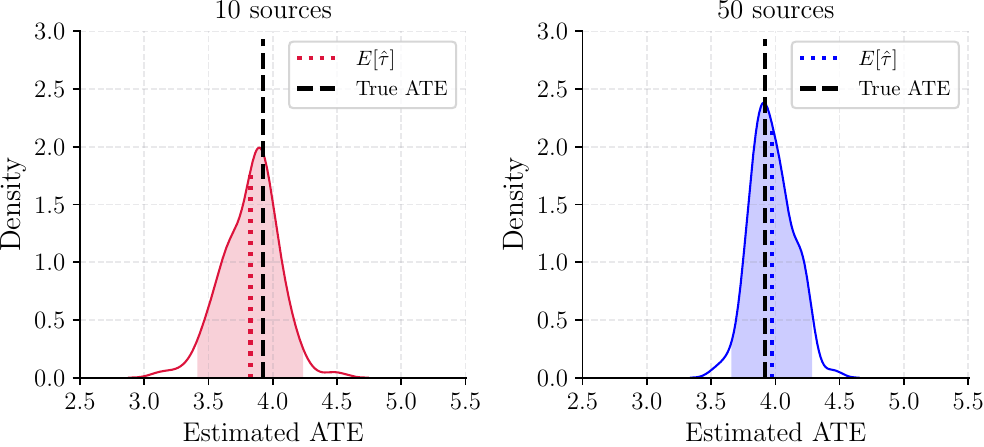}
\caption{Distribution of the estimated ATE.}
    \label{fig:dist-ate}
   
\end{figure}
\begin{table}\centering
\caption{Standard deviation of CausalFI vs. FedCI.
}\label{tab:standard-deviation}
 \vspace{6pt}
 \setlength{\tabcolsep}{12pt}
\scriptsize
\begin{tabular}{@{}lccc@{}}
\toprule
 Method        & 10 sources & 30 sources & 50 sources \\\cmidrule(r){1-1}\cmidrule(l){2-4}
FedCI    & 6.83       & 6.39       & 6.16       \\ 
CausalFI & 0.21       & 0.18       & 0.16       \\ \bottomrule
\end{tabular}
\end{table}

\noindent\textbf{Distribution of the estimated ATE:} We report distributions of the estimated ATE in Figure~\ref{fig:dist-ate}. As expected, the figures show that expectation of the causal estimand shifts toward the true ATE, and the confidence interval shrinks when the model is trained with more sources. It also shows that the true ATE is within the confidence interval, which is helpful for decision making. 
Note that the standard deviation of CausalFI is smaller than that of FedCI as the distribution in CausalFI is learned \emph{from all sources}. Meanwhile, standard deviation of FedCI only depends on the number of data points in \emph{a single source}. This is because FedCI is based on Gaussian Process, and it computes the variance using local training data in a specific source. In Table~\ref{tab:standard-deviation}, we report standard deviations of CausalFI and FedCI on different number of sources, which verifies our hypothesis.

\subsubsection{IHDP Data}
\label{sec:ihdp-causalfi}
\textbf{Data description:} 
Details of this dataset is described in Section~\ref{sec:ihdp}. 
The dataset is replicated ten times, and each replicate is divided into six subsets, each containing 124 data points. For each of these subsets, we further split them into three sets of 80, 24, and 20 data points for training, testing, and validation purposes when building the models. We then calculate the mean and standard error of the evaluation metrics across these ten replicates of the data. 
To simulate missing indicators, we also use Eq.~(\ref{eq:sim-missing-indicator}). We set $e_{j0}=1.0$, $e_{j1}=-0.5$, $e_{j2}=-0.03$, $\bm{e}_{j3}\sim \mathsf{N}(\bm{0},\mathbf{M})$, where $\mathbf{M}=\mathbf{L} \mathbf{L}^\top$, $\mathbf{L} \in \mathbb{R}^{21\times 4}$ and each element $L_{ij} \sim \mathsf{U}[0,0.5]$. We use Eq.~(\ref{eq:sim-missing-indicator}) to simulate missing indicators of 4 incomplete confounders. 
In Table~\ref{tb:data-stats-ihdp}, we report the percentage of missing values of each confounder in IHDP data.

\begin{table}\centering
\caption{Out-of-sample errors of CATE and ATE on IHDP data (lower is better). Top-3 performances are highlighted in bold. 
Methods trained on combined data (com)  violate privacy constraint and they are only used for comparison purpose.
}
\vspace{6pt}
\label{tb:errors-ihdp}
\setlength{\tabcolsep}{9pt}
\scriptsize
\begin{tabular}{@{}lcccccc@{}}
\toprule
    \multirow{2}{*}{Method} & \multicolumn{3}{c} {The error of CATE, $\sqrt{\epsilon_{\text{PEHE}}}$}             & \multicolumn{3}{c}{The error of ATE, $\epsilon_{\text{ATE}}$}   \\ \cmidrule(lr){2-4}\cmidrule(l){5-7}
        & 2 sources & 4 sources & 6 sources & 2 sources & 4 sources & 6 sources \\\cmidrule(r){1-1}\cmidrule(lr){2-4}\cmidrule(l){5-7}BART$_{\text{ppca+com}}$        & 5.35$\pm$1.4 & 5.15$\pm$1.3 & 4.72$\pm$1.4 & 1.72$\pm$1.0 & 0.99$\pm$0.5 & \textbf{0.67$\pm$0.2} \\
BART$_{\text{mice+com}}$ & 5.57$\pm$1.3 & 5.53$\pm$1.6 & 5.38$\pm$1.2 & 2.06$\pm$1.1 & 1.78$\pm$0.7 & 1.31$\pm$0.5 \\
BART$_{\text{ppca+avg}}$         & 5.58$\pm$1.5 & 5.55$\pm$1.3 & 5.36$\pm$1.2 & 2.09$\pm$1.2 & 1.80$\pm$0.7 & 1.28$\pm$0.4 \\
BART$_{\text{mice+avg}}$         & 5.93$\pm$1.7 & 5.79$\pm$1.4 & 5.52$\pm$1.4 & 2.12$\pm$1.3 & 1.88$\pm$0.8 & 1.52$\pm$0.6 \\
\cmidrule(r){1-1}\cmidrule(lr){2-4}\cmidrule(l){5-7}R-learner$_{\text{ppca+com}}$   & 6.08$\pm$1.7 & 5.47$\pm$1.8 & 6.10$\pm$1.5 & 3.34$\pm$1.2 & 3.54$\pm$1.3 & 4.28$\pm$1.4 \\
R-learner$_{\text{mice+com}}$  & 5.84$\pm$1.5 & 5.55$\pm$1.7 & 5.74$\pm$1.7 & 3.50$\pm$1.4 & 3.29$\pm$1.0 & 3.78$\pm$1.2 \\
R-learner$_{\text{ppca+avg}}$  & 6.64$\pm$1.7 & 5.26$\pm$1.3 & 5.70$\pm$1.4 & 3.74$\pm$1.2 & 2.84$\pm$1.3 & 3.55$\pm$1.5 \\
R-learner$_{\text{mice+avg}}$  & 5.49$\pm$1.6 & 5.49$\pm$1.4 & 5.69$\pm$1.3 & 3.02$\pm$1.2 & 2.93$\pm$1.1 & 3.65$\pm$0.9 \\
\cmidrule(r){1-1}\cmidrule(lr){2-4}\cmidrule(l){5-7}X-learner$_{\text{ppca+com}}$  & 4.04$\pm$1.5 & \textbf{3.57$\pm$1.2} & 3.20$\pm$0.9 & 1.06$\pm$0.4 & 0.90$\pm$0.3 & 1.02$\pm$0.4 \\
X-learner$_{\text{mice+com}}$  & \textbf{3.94$\pm$1.5} & 3.63$\pm$1.2 & 3.09$\pm$0.9 & \textbf{0.92$\pm$0.3} & 0.88$\pm$0.3 & 0.82$\pm$0.3 \\
X-learner$_{\text{ppca+avg}}$   & \textbf{3.93$\pm$1.5} & 3.95$\pm$1.6 & 3.74$\pm$1.4 & 1.22$\pm$0.6 & 1.34$\pm$0.6 & 1.18$\pm$0.5 \\
X-learner$_{\text{mice+avg}}$   & 3.97$\pm$1.5 & 4.05$\pm$1.6 & 3.85$\pm$1.5 & 1.33$\pm$0.6 & 1.29$\pm$0.6 & 1.13$\pm$0.6 \\
\cmidrule(r){1-1}\cmidrule(lr){2-4}\cmidrule(l){5-7}OthoRF$_{\text{ppca+com}}$      & 4.31$\pm$1.4 & 3.73$\pm$1.3 & 3.11$\pm$1.2 & 1.38$\pm$0.7 & 0.99$\pm$0.6 & 0.70$\pm$0.5 \\
OthoRF$_{\text{mice+com}}$  & 4.28$\pm$1.5 & 3.89$\pm$1.6 & \textbf{3.07$\pm$1.5} & 1.55$\pm$0.8 & 1.23$\pm$0.7 & 0.83$\pm$0.6 \\
OthoRF$_{\text{ppca+avg}}$      & 4.77$\pm$1.6 & 4.10$\pm$1.4 & 3.62$\pm$1.3 & 1.47$\pm$0.8 & 1.28$\pm$0.6 & 1.11$\pm$0.7 \\
OthoRF$_{\text{mice+avg}}$      & 4.65$\pm$1.6 & 3.98$\pm$1.6 & 3.37$\pm$1.3 & 1.68$\pm$0.7 & 1.41$\pm$0.6 & 0.98$\pm$0.6 \\
\cmidrule(r){1-1}\cmidrule(lr){2-4}\cmidrule(l){5-7}TARNet$_{\text{ppca+com}}$      & 5.97$\pm$1.8 & 5.63$\pm$1.6 & 4.22$\pm$1.2 & 2.05$\pm$0.6 & 1.49$\pm$0.5 & 0.87$\pm$0.1            \\
TARNet$_{\text{mice+com}}$      & 6.53$\pm$1.6 & 6.25$\pm$1.5 & 4.36$\pm$1.2 & 2.04$\pm$0.7 & 1.90$\pm$0.6 & 1.03$\pm$0.1           \\
TARNet$_{\text{ppca+avg}}$     & 6.05$\pm$1.7 & 5.96$\pm$1.6 & 5.31$\pm$1.4 & 2.11$\pm$0.7 & 1.65$\pm$0.5 & 0.92$\pm$0.3           \\
TARNet$_{\text{ppca+avg}}$      & 6.96$\pm$1.8 & 6.45$\pm$1.5 & 4.91$\pm$1.3 & 2.34$\pm$0.8 & 2.04$\pm$0.6 & 1.14$\pm$0.3           \\
\cmidrule(r){1-1}\cmidrule(lr){2-4}\cmidrule(l){5-7}CFR-mmd$_{\text{ppca+com}}$  & 5.53$\pm$1.5 & 6.69$\pm$1.7 & 5.48$\pm$1.1 & 1.19$\pm$0.3 & 1.76$\pm$0.5 & 1.59$\pm$0.3            \\
CFR-mmd$_{\text{mice+com}}$      & 6.20$\pm$1.5 & 6.39$\pm$1.7 & 5.58$\pm$1.2 & 1.58$\pm$0.4 & 1.88$\pm$0.5 & 1.81$\pm$0.4           \\
CFR-mmd$_{\text{ppca+avg}}$  & 5.78$\pm$1.7 & 6.82$\pm$1.8 & 5.63$\pm$1.3 & 1.31$\pm$0.4 & 1.81$\pm$0.5 & 1.63$\pm$0.3            \\
CFR-mmd$_{\text{ppca+avg}}$  & 6.67$\pm$1.6 & 6.52$\pm$1.5 & 5.76$\pm$1.3 & 1.61$\pm$0.5 & 1.91$\pm$0.5 & 1.89$\pm$0.4           \\
\cmidrule(r){1-1}\cmidrule(lr){2-4}\cmidrule(l){5-7}CFR-wass$_{\text{ppca+com}}$ & 5.98$\pm$1.7 & 6.31$\pm$1.5 & 5.07$\pm$1.2 & 1.05$\pm$0.3 & 1.15$\pm$0.3 & 1.39$\pm$0.3           \\
CFR-wass$_{\text{mice+com}}$      & 5.93$\pm$1.7 & 6.18$\pm$1.6 & 5.43$\pm$1.4 & 1.18$\pm$0.3 & 1.64$\pm$0.5 & 1.67$\pm$0.7           \\
CFR-wass$_{\text{ppca+avg}}$ & 6.21$\pm$1.6 & 6.53$\pm$1.6 & 5.31$\pm$1.3 & 1.21$\pm$0.4 & 1.21$\pm$0.4 & 1.46$\pm$0.5           \\
CFR-wass$_{\text{mice+avg}}$ & 6.31$\pm$1.7 & 6.52$\pm$1.8 & 5.61$\pm$1.3 & 1.32$\pm$0.5 & 1.78$\pm$0.5 & 1.72$\pm$0.6           \\
\cmidrule(r){1-1}\cmidrule(lr){2-4}\cmidrule(l){5-7}CEVAE$_{\text{ppca+com}}$      & 4.74$\pm$1.5 & 4.43$\pm$1.4 & 4.54$\pm$1.5 & \textbf{0.65$\pm$0.2} & 1.06$\pm$0.3 & 0.98$\pm$0.3 \\
CEVAE$_{\text{mice+com}}$  & 4.82$\pm$1.4 & 4.41$\pm$1.3 & 4.63$\pm$1.2 & 1.03$\pm$0.4 & 1.21$\pm$0.3 & 1.02$\pm$0.3 \\
CEVAE$_{\text{ppca+avg}}$       & 4.97$\pm$1.6 & 4.65$\pm$1.3 & 4.72$\pm$1.1 & 0.98$\pm$0.4 & 1.23$\pm$0.4 & 1.08$\pm$0.3 \\
CEVAE$_{\text{mice+avg}}$       & 5.03$\pm$1.6 & 4.82$\pm$1.4 & 4.74$\pm$1.2 & 1.24$\pm$0.7 & 1.32$\pm$0.3 & 1.15$\pm$0.3 \\
\cmidrule(r){1-1}\cmidrule(lr){2-4}\cmidrule(l){5-7}FedCI$_{\text{ppca+com}}$           & 3.97$\pm$1.6 & \textbf{3.52$\pm$1.3} & 3.15$\pm$1.4 & 0.98$\pm$0.3 & \textbf{0.78$\pm$0.2} & \textbf{0.61$\pm$0.2}          \\
FedCI$_{\text{mice+com}}$           & 4.12$\pm$1.7 & 3.85$\pm$1.4 & \textbf{3.04$\pm$1.2} & 1.01$\pm$0.4 & \textbf{0.85$\pm$0.3} & 0.78$\pm$0.2          \\
FedCI$_{\text{ppca+avg}}$           & 4.61$\pm$1.7 & 3.89$\pm$1.5 & 3.62$\pm$1.3 & 1.12$\pm$0.3 & 1.02$\pm$0.2 & 0.91$\pm$0.2          \\
FedCI$_{\text{mice+avg}}$           & 4.45$\pm$1.8 & 3.91$\pm$1.5 & 3.52$\pm$1.3 & 1.17$\pm$0.4 & 0.92$\pm$0.2 & 0.89$\pm$0.2             \\\cmidrule(r){1-1}\cmidrule(lr){2-4}\cmidrule(l){5-7}CausalFI             & \textbf{3.67$\pm$1.7} & \textbf{3.33$\pm$1.4} & \textbf{2.99$\pm$1.2} & \textbf{0.73$\pm$0.3} & \textbf{0.53$\pm$0.2} & \textbf{0.32$\pm$0.1} \\\bottomrule
\end{tabular}
\end{table}

\begin{figure}\centering
    \includegraphics[width=0.65\textwidth]{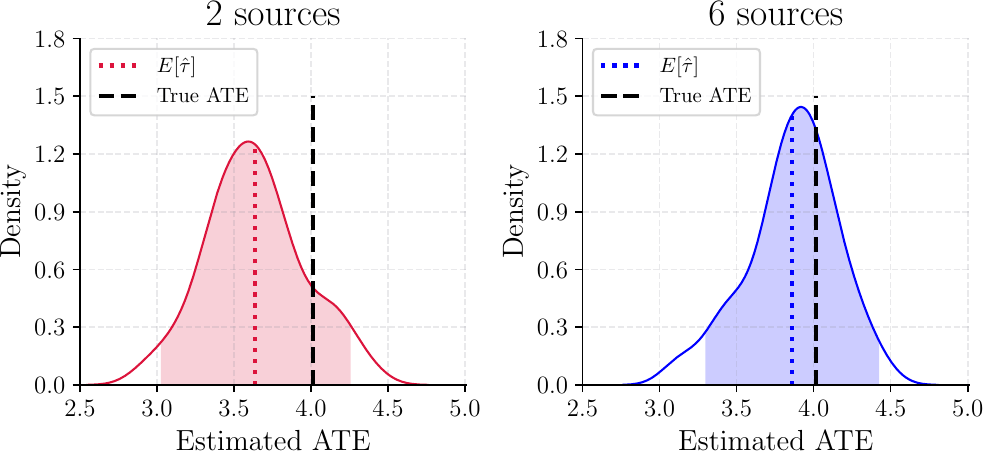}
\caption{Distribution of the estimated ATE on IHDP data in replicate \#1.}
    \label{fig:dist-ate-ihdp}
\end{figure}

\noindent\textbf{Results and discussion:} We report the results in Table~\ref{tb:errors-ihdp}. Similar to experimental results on synthetic data, the results in this dataset show that the performance of CausalFI
is competitive to the baselines. In particular, CausalFI is among top-3 performance. This again verifies the efficacy of the proposed method. We also observe that the errors of CEVAE, CFR-mmd, and CFR-wass in estimating ATE are a bit higher when there are more missing data points. This could be because of the performance of the impute method. We also report comparison with IPW and DR regarding in-sample $\epsilon_{\text{ATE}}$ in Table~\ref{tab:errors-vs-missing-methods-ihdp}, which shows competitive results. However, these baselines are trained on combined data which violates privacy constraint of the data. Similar to the results on synthetic data, the distribution of ATE on IHDP data are shown in Figure~\ref{fig:dist-ate-ihdp}. The figures show that the true ATE are within the confidence interval, and the mean of the estimated ATE shifts towards the true ATE when there are more sources. We also compare standard deviation of CausalFI with FedCI in Table~\ref{tab:standard-deviation-ihdp}. The figures show that standard deviation of CausalFI is smaller than that of FedCI, which demonstrates that CausalFI would give a more confident result than FedCI.

\begin{table}\centering
\caption{Standard deviation of CausalFI vs. FedCI on IHDP data.
}\label{tab:standard-deviation-ihdp}
 \vspace{6pt}
\scriptsize
 \setlength{\tabcolsep}{12pt}
\begin{tabular}{@{}lccc@{}}
\toprule
Method         & 2 sources & 4 sources & 6 sources \\\cmidrule(r){1-1}\cmidrule(l){2-4}
FedCI    & 7.60       & 7.71       & 7.47       \\
CausalFI & 0.31       & 0.30       & 0.28       \\ \bottomrule
\end{tabular}
\end{table}

\subsection{Summary}

We introduced CausalFI as a solution to the problem of federated causal inference arising from decentralized and incomplete data. It addresses the challenge of missing data under the Missing at Random (MAR) and Missing Completely at Random (MCAR) assumptions. Our approach involves retrieving the conditional distribution of missing confounders based on the observed confounders obtained from decentralized data sources. Crucially, our method enables the estimation of heterogeneous causal effects while maintaining privacy, without the necessity of sharing raw data.

 \section{Conclusion}

In this article, we introduced a framework for federated learning of causal effects. The proposed framework decomposed the objective function into multiple components, each associated with a data source. It enables the training of causal models without sharing raw data among the sources. We identified challenges of estimating causal effects in federated settings and introduced three instances of the proposed framework to address these challenges. We first proposed FedCI, a Bayesian causal inference paradigm via a reformulation of multi-output GPs to learn causal effects while keeping data at their local sites. An inference method involving the decomposition of ELBO was presented, allowing the model to be trained in a federated setting. Although FedCI could estimate causal effects in federated settings, it assumed that the data distribution among the sources was the same. To address dissimilar data distribution among the sources, we proposed CausalRFF, which allowed each component data group to inherit different distributions and required no prior knowledge of data discrepancies among the sources. CausalRFF utilized Random Fourier Features that naturally induced the decomposition of the loss function into individual components. We also proved statistical guarantees which showed how multiple data sources were effectively incorporated into our causal model. Both FedCI and CausalRFF were not designed to work with missing data, which might be important in real-life applications. We introduced CausalFI to address the challenge of federated causal inference from incomplete and decentralized data. It tackled the issue of missing data under MAR and MCAR assumptions. CausalFI recovers the conditional distribution of missing confounders given the observed confounders from the decentralized data sources. Importantly, it estimates heterogeneous causal effects while ensuring privacy but not requiring the sharing of raw data.

This article also opened several promising directions for future research, as outlined below. The inherent use of GPs in FedCI would incur computational time of the inverse covariance matrix in each source of cubic time complexity. Hence, a possible future research direction is to reformulate this in terms of sparse Gaussian process models \citep{Hensman13}. Another interesting future research direction is to extend the proposed framework for missing not at random data. Such a scenario might require further assumptions for the causal effects to be identifiable. In addition, it is important to test for the missing data mechanism. At present, the only method is to apply existing tests locally on each source and then aggregate them with a simple voting approach. Developing a federated test for the missing data mechanism represents another avenue for future research. 
This article presents an important step towards a privacy-preserving causal learning model. Another promising area for future exploration involves integrating the framework with differential privacy to bolster privacy assurances. A potential implementation of differential privacy within the proposed framework is to apply gradient clipping and introduce Gaussian noise to Algorithm~\ref{algo:federated-training}, drawing inspiration from the foundational DP-SGD algorithm proposed by \citet{abadi2016deep}. Nevertheless, the straightforward application of DP-SGD might severely impact performance due to the addition of Gaussian noise to the gradients and the limitations imposed by batch size. Consequently, causal model performance suffers, and convergence is slow. Addressing these challenges and adapting Algorithm~\ref{algo:federated-training} for DP-SGD is an interesting research direction.

\newpage

\appendix

\section{The Preprocessing Procedure}
\label{sec:appendix-preprocessing}
The assumptions were described briefly in Section~\ref{sec:assumptions} of the main text. Here we present the preprocessing procedures to remove duplicated individuals.

\begin{figure}\centering
    \includegraphics[width=0.7\textwidth]{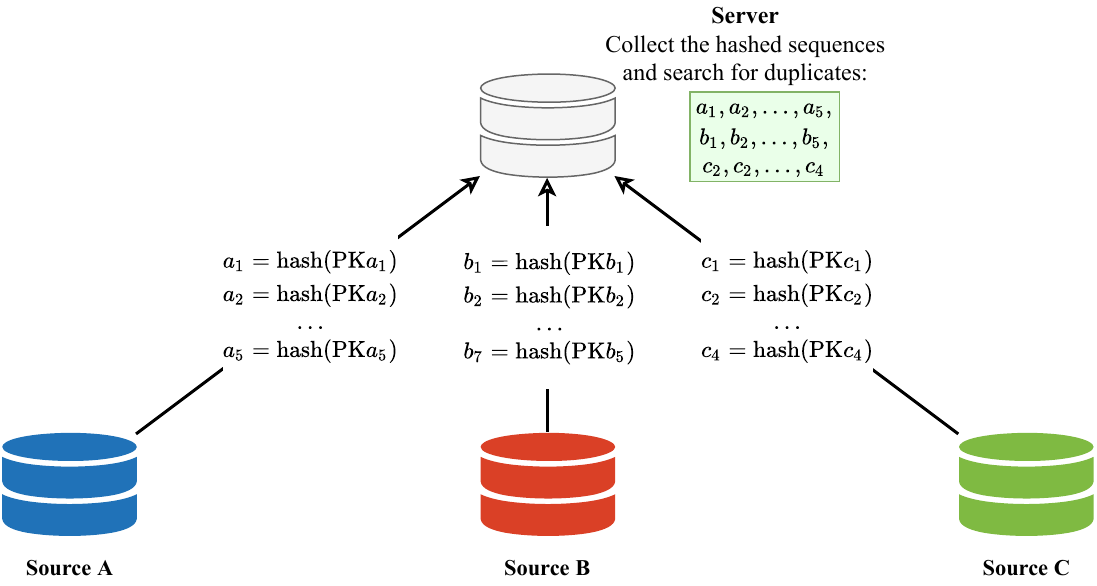}
    \caption{The secure preprocessing procedures to identify duplicated individuals among multiple sources. $\text{PK}a_i$ ($i=1,\!...,5$), $\text{PK}b_i$ ($i=1,\!...,7$), $\text{PK}c_i$ ($i=1,\!...,4$) are the primary keys of each individual in each source. $a_i$ ($i=1,\!...,5$), $b_i$ ($i=1,\!...,7$), $c_i$ ($i=1,\!...,4$) are the hashed sequences of these individuals.}
    \label{fig:preprocessing}
\end{figure}

The preprocessing procedure are summarized as follows. Firstly, each source would use a one-way hash function (such as MD4, MD5, SHA or SHA256) to encrypt each individuals' primary key and then send the hashed sequences to a server. By doing this, the individuals' data are secured. Note that the one-way hash function is agreed among the sources so that they would use the same function. Then, the server collects all hashed sequences from all sources and perform a matching algorithm to see if there exists repeated individuals among different sources. For each repeated individual, the server randomly choose to keep it on a small number (predefined) of sources and inform the other sources to exclude this individual from the training process. The whole procedure is to ensure that an individual does not exists in a huge number of sources, thus prevent learning a biased model. We summarize the procedure in Figure~\ref{fig:preprocessing}.

Assumption~\ref{assumption:unique-ident} and the preprocessing procedure are required for data that are highly repeated in different sources only. For data that are not likely to have a high number of repetitions such as patients from different hospitals of different countries, the above assumption and the preprocessing procedure are not required. Note that the existing methods also need Assumption~\ref{assumption:unique-ident} since they need to combine data and remove repeated individuals.

In this work, we assume that all of the assumptions described in this section are satisfied, and the preprocessing procedure was performed if it is necessary.

\section{The Federated Evidence Lower Bound}
\label{sec:appendix-elbo}

Naively applying variational inference would lead to a non-decomposable ELBO. The proposed ELBO can be decomposed into multiple components, thus enabling federated optimization. We give a full derivation as follows:
\begin{align*}
        \log \p(\mathbf{y}_{\textrm{obs}}\,|\,\mathbf{X},\mathbf{w}) 
        &= \log \int p(\mathbf{y}_{\textrm{obs}},\mathbf{g}, \Psi, \Sigma\,|\,\mathbf{X},\mathbf{w}) d\mathbf{g}d\Psi d\Sigma\\
        &= \log \int p(\mathbf{y}_{\textrm{obs}}\,|\,\mathbf{g}, \Psi, \Sigma,\mathbf{X},\mathbf{w})p(\mathbf{g}, \Psi, \Sigma|\mathbf{X},\mathbf{w}) d\mathbf{g}d\Psi d\Sigma.
\end{align*}

\begin{figure}[!ht]
\centering
		\includegraphics[width=0.6\textwidth]{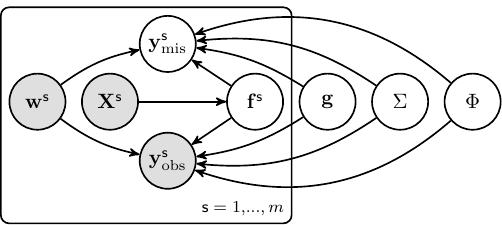}

    \caption{
        Graphical model that summarizes the proposed framework with treatment $\mathbf{w}^\mathsf{s}$, covariate $\mathbf{X}^\mathsf{s}$, and the two potential outcomes $\mathbf{y}_\textrm{mis}^\mathsf{s}$ and $\mathbf{y}_\textrm{obs}^\mathsf{s}$. The quantity $\mathbf{f}^\mathsf{s}$ is idiosyncratic to the sources and $\mathbf{g}$ contains shared characteristics across all the sources. $\Sigma$ and $\Psi$ are shared parameters. Note that this is not a causal graph.
    } \label{fig:the-model}
\end{figure}

From Figure~\ref{fig:the-model}, we see that $\mathbf{g}, \Psi, \Sigma \ind \mathbf{X}^\mathsf{s},\mathbf{w}^\mathsf{s}$ (for all $\mathsf{s}=1,2,\dots,m$), i.e, $\mathbf{g}, \Psi, \Sigma$ are independent with $\mathbf{X}^\mathsf{s},\mathbf{w}^\mathsf{s}$ when $\mathbf{y}_{\text{obs}}^\mathsf{s}, \mathbf{y}_{\text{mis}}^\mathsf{s}$ are not given. Thus, $p(\mathbf{g}, \Psi, \Sigma|\mathbf{X},\mathbf{w}) = p(\mathbf{g}, \Psi, \Sigma)$.

In addition, from Figure~\ref{fig:the-model}, 
we also have 
\begin{align*}
    p(\mathbf{y}_{\textrm{obs}}\,|\,\mathbf{g}, \Psi, \Sigma,\mathbf{X},\mathbf{w}) = \prod_{\mathsf{s}=1}^m   p(\mathbf{y}_{\textrm{obs}}^\mathsf{s}\,|\,\mathbf{g}, \Psi, \Sigma,\mathbf{X}^\mathsf{s},\mathbf{w}^\mathsf{s}).
\end{align*}
Thus,
\begin{align*}
        \log p(\mathbf{y}_{\textrm{obs}}\,|\,\mathbf{X},\mathbf{w}) 
&= \log \int q(\mathbf{g}, \Psi, \Sigma)\prod_{\mathsf{s}=1}^m   p(\mathbf{y}_{\textrm{obs}}^\mathsf{s}\,|\,\mathbf{g}, \Psi, \Sigma,\mathbf{X}^\mathsf{s},\mathbf{w}^\mathsf{s})\times\frac{p(\mathbf{g}, \Psi, \Sigma)}{q(\mathbf{g}, \Psi, \Sigma)} d\mathbf{g}d\Psi d\Sigma\\
        &\ge \int q(\mathbf{g}, \Psi, \Sigma)\log\Bigg(\prod_{\mathsf{s}=1}^m   p(\mathbf{y}_{\textrm{obs}}^\mathsf{s}\,|\,\mathbf{g}, \Psi, \Sigma,\mathbf{X}^\mathsf{s},\mathbf{w}^\mathsf{s}) \times\frac{p(\mathbf{g}, \Psi, \Sigma)}{q(\mathbf{g}, \Psi, \Sigma)}\Bigg) d\mathbf{g}d\Psi d\Sigma\\ 
        &= \sum_{\mathsf{s}=1}^m \e_q[\log  p(\mathbf{y}_{\textrm{obs}}^\mathsf{s}\,|\,\mathbf{g}, \Psi, \Sigma,\mathbf{X}^\mathsf{s},\mathbf{w}^\mathsf{s})] - \mathbb{D}_{\text{KL}}[q(\mathbf{g}, \Psi, \Sigma)\|p(\mathbf{g}, \Psi, \Sigma)]\\
        &= \sum_{\mathsf{s}=1}^m\Big( \e_q[\log  p(\mathbf{y}_{\textrm{obs}}^\mathsf{s}\,|\,\mathbf{g}, \Psi, \Sigma,\mathbf{X}^\mathsf{s},\mathbf{w}^\mathsf{s})] - \frac{1}{m}\mathbb{D}_{\text{KL}}[q(\mathbf{g}, \Psi, \Sigma)\|p(\mathbf{g}, \Psi, \Sigma)]\Big)\\
        &= \sum_{\mathsf{s}=1}^m J^\mathsf{s},
\end{align*}
where
\begin{align*}
    J^\mathsf{s} &\vcentcolon=  \e_q[\log  p(\mathbf{y}_{\textrm{obs}}^\mathsf{s}\,|\,\mathbf{g}, \Psi, \Sigma,\mathbf{X}^\mathsf{s},\mathbf{w}^\mathsf{s})] - \frac{1}{m}\mathbb{D}_{\text{KL}}[q(\mathbf{g}, \Psi, \Sigma)\|p(\mathbf{g}, \Psi, \Sigma)].
\end{align*}
Hence, we can divide the ELBO into multiple components, which leads to federated training of the model. Without the proposed model, the ELBO cannot be decomposed into multiple components and hence cannot be trained in a federated setting.

\section{Proof of Lemma~1}
\label{sec:appendix-proof-lem-1}

\begin{proof}
We denote $\xi_0^{\mathsf{s}} \sim \mathsf{N}(\mathbf{0}, \mathbf{I}_{n_{\mathsf{s}}})$ and $\xi_1^{\mathsf{s}} \sim \mathsf{N}(\mathbf{0}, \mathbf{I}_{n_{\mathsf{s}}})$. Then, from the model definition (Eq.~(5) in the main text), we have
\begin{align*}
\setlength\arraycolsep{0.0pt} \begin{bmatrix}
y_1^{\mathsf{s}}(0)&\dots&y_{n_{\mathsf{s}}}^{\mathsf{s}}(0)\\
y_1^{\mathsf{s}}(1)&\dots&y_{n_{\mathsf{s}}}^{\mathsf{s}}(1)
\end{bmatrix} &=\Psi^{\frac{1}{2}} \begin{bmatrix}
f_1^{\mathsf{s}}(0)+g^{\mathsf{s}}(0)&\dots&f_{n_{\mathsf{s}}}^{\mathsf{s}}(0)+g^{\mathsf{s}}(0)\\
f_1^{\mathsf{s}}(1)+g^{\mathsf{s}}(1)&\dots&f_{n_{\mathsf{s}}}^{\mathsf{s}}(1)+g^{\mathsf{s}}(1)
\end{bmatrix} +\Sigma^{\frac{1}{2}}\begin{bmatrix}
\varepsilon_1^{\mathsf{s}}(0)&\dots&\varepsilon_{n_{\mathsf{s}}}^{\mathsf{s}}(0)\\
\varepsilon_1^{\mathsf{s}}(1)&\dots&\varepsilon_{n_{\mathsf{s}}}^{\mathsf{s}}(1)
\end{bmatrix}.
\end{align*}
The above equation is equivalent to the following
\begin{align*}
&\mathbf{Y}^{\mathsf{s}} =  \setlength\arraycolsep{5pt} \begin{bmatrix}
\bm{\mu}_0&\bm{\mu}_1
\end{bmatrix}
(\Psi^{\frac{1}{2}})^\top +\setlength\arraycolsep{5pt} \begin{bmatrix}
\varepsilon_0^{\mathsf{s}}&\varepsilon_1^{\mathsf{s}}
\end{bmatrix}(\Sigma^{\frac{1}{2}})^\top,
\end{align*}
where 
\begin{align*}
    \bm{\mu}_0 &= \mu_0(\mathbf{X}^{\mathsf{s}}) + \mathbf{g}_0^{\mathsf{s}} + (\mathbf{K}^{\mathsf{s}})^{\frac{1}{2}}\xi_0^{\mathsf{s}}, &\bm{\mu}_1 &= \mu_1(\mathbf{X}^{\mathsf{s}}) + \mathbf{g}_1^{\mathsf{s}} + (\mathbf{K}^{\mathsf{s}})^{\frac{1}{2}}\xi_1^{\mathsf{s}}.
\end{align*}
Further expanding the right hand side, we have
\begin{align*}
&\mathbf{Y}^{\mathsf{s}} \!=\! \setlength\arraycolsep{5pt}\begin{bmatrix}\mu_0(\mathbf{X}^{\mathsf{s}})\!+\!\mathbf{g}_0^{\mathsf{s}}&\mu_1(\mathbf{X}^{\mathsf{s}})+\mathbf{g}_1^{\mathsf{s}}
\end{bmatrix}(\Psi^{\frac{1}{2}})^\top + (\mathbf{K}^{\mathsf{s}})^{\frac{1}{2}}\setlength\arraycolsep{5pt}\begin{bmatrix}\xi_0^{\mathsf{s}}&\xi_1^{\mathsf{s}}
\end{bmatrix}(\Psi^{\frac{1}{2}})^\top +\setlength\arraycolsep{5pt}\begin{bmatrix}
\varepsilon_0^{\mathsf{s}}&\varepsilon_1^{\mathsf{s}}
\end{bmatrix}(\Sigma^{\frac{1}{2}})^\top\\
&\text{vec}(\mathbf{Y}^{\mathsf{s}}) =  \left(\Psi^{\frac{1}{2}} \otimes \mathbf{I}_{n_{\mathsf{s}}}\right)\begin{bmatrix}\mu_0(\mathbf{X}^{\mathsf{s}})+\mathbf{g}_0^{\mathsf{s}}\\\mu_1(\mathbf{X}^{\mathsf{s}})+\mathbf{g}_1^{\mathsf{s}}
\end{bmatrix}  +\left(\Psi^{\frac{1}{2}} \otimes (\mathbf{K}^{\mathsf{s}})^{\frac{1}{2}}\right)\begin{bmatrix}\xi_0^{\mathsf{s}}\\\xi_1^{\mathsf{s}}
\end{bmatrix} + (\Sigma^{\frac{1}{2}} \otimes \mathbf{I}_{n_{\mathsf{s}}})\begin{bmatrix}
\varepsilon_0^{\mathsf{s}}\\\varepsilon_1^{\mathsf{s}}
\end{bmatrix},
\end{align*}
where $\text{vec}(\cdot)$ denotes the vectorization of a matrix, which converts a matrix into a column vector.

For the second term on the right hand side of the above equation, note that $\xi_0^{\mathsf{s}} \sim \mathsf{N}(\mathbf{0}, \mathbf{I}_{n_{\mathsf{s}}})$ and $\xi_1^{\mathsf{s}} \sim \mathsf{N}(\mathbf{0}, \mathbf{I}_{n_{\mathsf{s}}})$, so we have the following
\begin{align*}
&\begin{bmatrix}\xi_0^{\mathsf{s}}\\\xi_1^{\mathsf{s}}
\end{bmatrix} \sim \mathsf{N}(\mathbf{0}, \mathbf{I}_{2n_{\mathsf{s}}})\\
&\left(\Psi^{\frac{1}{2}} \otimes (\mathbf{K}^{\mathsf{s}})^{\frac{1}{2}}\right)\begin{bmatrix}\xi_0^{\mathsf{s}}\\\xi_1^{\mathsf{s}}\end{bmatrix} \sim \mathsf{N}\left(\mathbf{0}, \left(\Psi^{\frac{1}{2}} \otimes (\mathbf{K}^{\mathsf{s}})^{\frac{1}{2}}\right)\mathbf{I}_{2N}\left(\!\Psi^{\frac{1}{2}} \otimes (\mathbf{K}^{\mathsf{s}})^{\frac{1}{2}}\right)^\top\right)\\
&\left(\Psi^{\frac{1}{2}} \otimes (\mathbf{K}^{\mathsf{s}})^{\frac{1}{2}}\right)\begin{bmatrix}\xi_0^{\mathsf{s}}\\\xi_1^{\mathsf{s}}
\end{bmatrix} \sim \mathsf{N}\left(\mathbf{0}, \Psi\otimes \mathbf{K}^{\mathsf{s}}\right).
\end{align*}
For the last term, note that $\varepsilon_0^{\mathsf{s}} \sim \mathsf{N}(0, \mathbf{I}_{n_{\mathsf{s}}}), \varepsilon_1^{\mathsf{s}} \sim \mathsf{N}(0, \mathbf{I}_{n_{\mathsf{s}}})$, thus
\begin{align*}
&\begin{bmatrix}\varepsilon_0^{\mathsf{s}}\\\varepsilon_1^{\mathsf{s}}
\end{bmatrix} \sim \mathsf{N}(\mathbf{0}, \mathbf{I}_{2n_{\mathsf{s}}})\\
&\left(\Sigma^{\frac{1}{2}} \otimes \mathbf{I}_{n_{\mathsf{s}}}\right)\begin{bmatrix}\varepsilon_0^{\mathsf{s}}\\\varepsilon_1^{\mathsf{s}}
\end{bmatrix} \sim \mathsf{N}\left(\mathbf{0}, \left(\Sigma^{\frac{1}{2}} \otimes \mathbf{I}_{n_{\mathsf{s}}}\right)\mathbf{I}_{2n}\left(\Sigma^{\frac{1}{2}} \otimes \mathbf{I}_{n_{\mathsf{s}}}\right)^\top\right)\\
&\left(\Sigma^{\frac{1}{2}} \otimes \mathbf{I}_{n_{\mathsf{s}}}\right)\begin{bmatrix}\varepsilon_0^{\mathsf{s}}\\\varepsilon_1^{\mathsf{s}}
\end{bmatrix} \sim \mathsf{N}\left(\mathbf{0}, \Sigma\otimes \mathbf{I}_{n_{\mathsf{s}}}\right).
\end{align*}
Consequently, 
\vspace{-6pt}
\begin{align*}
&\text{vec}(\mathbf{Y}^{\mathsf{s}}) \big| \Psi, \Sigma, \mathbf{X}^{\mathsf{s}}, \mathbf{w}^{\mathsf{s}}, \mathbf{g}^{\mathsf{s}} \sim \mathsf{N}\left( \left(\Psi^{\frac{1}{2}} \otimes \mathbf{I}_{n_{\mathsf{s}}}\right)\begin{bmatrix}\mu_0(\mathbf{X}^{\mathsf{s}})+ \mathbf{g}_0^{\mathsf{s}}\\\mu_1(\mathbf{X}^{\mathsf{s}}) + \mathbf{g}_1^{\mathsf{s}}
\end{bmatrix} , \Psi \otimes \mathbf{K}^{\mathsf{s}} + \Sigma \otimes \mathbf{I}_{n_{\mathsf{s}}}\right),
\end{align*}
which implies that
\begin{align*}
&\begin{bmatrix}
\mathbf{y}^{\mathsf{s}}(0)\\
\mathbf{y}^{\mathsf{s}}(1)
\end{bmatrix}\Big|\Psi, \Sigma, \mathbf{X}^{\mathsf{s}}, \mathbf{w}^{\mathsf{s}}, \mathbf{g}^{\mathsf{s}} \sim \mathsf{N}\left( \left(\Psi^{\frac{1}{2}} \otimes \mathbf{I}_{n_{\mathsf{s}}}\right)\begin{bmatrix}\mu_0(\mathbf{X}^{\mathsf{s}})+ \mathbf{g}_0^{\mathsf{s}}\\\mu_1(\mathbf{X}^{\mathsf{s}})+ \mathbf{g}_1^{\mathsf{s}}
\end{bmatrix}, \Psi \otimes \mathbf{K}^{\mathsf{s}} + \Sigma \otimes \mathbf{I}_{n_{\mathsf{s}}}\right).
\end{align*}
This completes the proof.
\end{proof}

\section{Proof of Lemma~2}

\label{sec:appendix-proof-lem-2}

\begin{proof}
Following the proof of Lemma~2, we note that if the observed treatment $w_i^{\mathsf{s}} = 0$, then the mean of $p(y^{\mathsf{s}}_{i,\textrm{obs}} | \mathbf{X}^{\mathsf{s}}, \mathbf{w}^{\mathsf{s}}, \Psi, \Sigma, \mathbf{g}^{\mathsf{s}})$ equals to the mean of $p(y_i^{\mathsf{s}}(0) | \Psi, \Sigma, \mathbf{X}^{\mathsf{s}}, \mathbf{w}^{\mathsf{s}}, \mathbf{g}^{\mathsf{s}})$ and the mean of $p(y^{\mathsf{s}}_{i,\textrm{mis}} | \mathbf{X}^{\mathsf{s}}, \mathbf{w}^{\mathsf{s}}, \Psi, \Sigma, \mathbf{g}^{\mathsf{s}})$ equals to the mean of $p(y_i^{\mathsf{s}}(1) | \Psi, \Sigma, \mathbf{X}^{\mathsf{s}}, \mathbf{w}^{\mathsf{s}}, \mathbf{g}^{\mathsf{s}})$. If the observed treatment $w_i^{\mathsf{s}} = 1$, then the mean of $p(y^{\mathsf{s}}_{i,\textrm{obs}} | \mathbf{X}^{\mathsf{s}}, \mathbf{w}^{\mathsf{s}}, \Psi, \Sigma, \mathbf{g}^{\mathsf{s}})$ equals to the mean of $p(y_i^{\mathsf{s}}(1) | \Psi, \Sigma, \mathbf{X}^{\mathsf{s}}, \mathbf{w}^{\mathsf{s}}, \mathbf{g}^{\mathsf{s}})$ and the mean of $p(y^{\mathsf{s}}_{i,\textrm{mis}} | \mathbf{X}^{\mathsf{s}}, \mathbf{w}^{\mathsf{s}}, \Psi, \Sigma, \mathbf{g}^{\mathsf{s}})$ equals to the mean of $p(y_i^{\mathsf{s}}(0) | \Psi, \Sigma, \mathbf{X}^{\mathsf{s}}, \mathbf{w}^{\mathsf{s}}, \mathbf{g}^{\mathsf{s}})$. Hence, we have
\begin{align*}
\mu_{\textrm{obs}}(\mathbf{X}^\mathsf{s}) &= (\mathbf{1} - \mathbf{w}^\mathsf{s})\odot\mathbf{m}_0 + \mathbf{w}^\mathsf{s} \odot\mathbf{m}_1,\\
\mu_{\textrm{mis}}(\mathbf{X}^\mathsf{s}) &= \mathbf{w}^\mathsf{s} \odot\mathbf{m}_0 + (\mathbf{1} - \mathbf{w}^\mathsf{s}) \odot\mathbf{m}_1,
\end{align*}
Similarly, for the covariance matrix, each element in $\mathbf{K}_{\textrm{obs}}$, $\mathbf{K}_{\textrm{mis}}$, and   $\mathbf{K}_{\textrm{om}}$ also depends on whether $w_i^{\mathsf{s}} = 0$ or $w_i^{\mathsf{s}} = 1$. So each element in these matrices is computed by the following kernel function
\begin{align*}
k_{\textrm{obs}}(x_i, x_j) &= \big[(1-w_i)(1-w_j)\psi_{11} + w_iw_j\psi_{22} + (1-w_i)w_j\psi_{12} + w_i(1-w_j)\psi_{21}\big] \mathsf{k}(x_i, x_j)\\[-0cm]
&\,\,\,\,\, + \big[(1-w_i)\sigma_{11} + w_i\sigma_{22}\big] \mathds{1}_{i=j},\\
k_{\textrm{mis}}(x_i, x_j) &= \big[w_iw_j\psi_{11} + (1-w_i)(1-w_j)\psi_{22} + (1-w_i)w_j\psi_{21} + w_i(1-w_j)\psi_{12}\big] \mathsf{k}(x_i, x_j)\\
& \,\,\,\,\,+ \big[w_i\sigma_{11} + (1-w_i)\sigma_{22}\big] \mathds{1}_{i=j},\\
k_{\textrm{om}}(x_i, x_j) &= \big[(1-w_i)(1-w_j)\psi_{21} + w_iw_j\psi_{12} + (1-w_i)w_j\psi_{22} + w_i(1-w_j)\psi_{11}\big] \mathsf{k}(x_i, x_j) \\
&\,\,\,\,\,+ \big[(1-w_i)\sigma_{21} + w_i\sigma_{12}\big] \mathds{1}_{i=j},
\end{align*}
where $\psi_{ab}$ and $\sigma_{ab}$ are the $(a,b)$--th elements of $\Psi$ and $\Sigma$, respectively.

This completes the proof.
\end{proof}

\section{Identification of CausalRFF}
\label{ap:proof-iden}
The causal effects are unidentifiable  if the confounders are unobserved. However, \citet{louizos2017causal} showed that if the joint distribution $\p_\mathsf{s}(x^\mathsf{s},y^\mathsf{s},w^\mathsf{s},z^\mathsf{s})$ can be recovered, then the causal effects are identifiable. In the following, we show how they are identifiable.

\begin{proof} The proof is adapted from \citet[Theorem~1,][]{louizos2017causal}. We need to show that the distribution $\p_\mathsf{s}(y^\mathsf{s}|\doo(W=\bm{w}^\mathsf{s}),x^\mathsf{s})$ is identifiable from observational data. We have
\begin{align*}
    \p_\mathsf{s}(y^\mathsf{s}|\doo(W=\bm{w}^\mathsf{s}),x^\mathsf{s}) &= \int \p_\mathsf{s}(y^\mathsf{s}|\doo(W=\bm{w}^\mathsf{s}),x^\mathsf{s}, z^\mathsf{s})\p_\mathsf{s}(z^\mathsf{s}|\doo(W=\bm{w}^\mathsf{s}),x^\mathsf{s}) dz^\mathsf{s}\\
    &= \int \p_\mathsf{s}(y^\mathsf{s}|\bm{w}^\mathsf{s},x^\mathsf{s}, z^\mathsf{s})\p_\mathsf{s}(z^\mathsf{s}|x^\mathsf{s}) dz^\mathsf{s}.
\end{align*}
where the last equality is obtained by applying the $do$-calculus. The last expression, $\int \p_\mathsf{s}(y^\mathsf{s}|\bm{w}^\mathsf{s},x^\mathsf{s}, z^\mathsf{s})\p_\mathsf{s}(z^\mathsf{s}|x^\mathsf{s}) dz^\mathsf{s}$,  can be identified by the joint distribution $\p_\mathsf{s}(x^\mathsf{s},y^\mathsf{s},w^\mathsf{s},z^\mathsf{s})$. In our work, $\p_\mathsf{s}(x^\mathsf{s},y^\mathsf{s},w^\mathsf{s},z^\mathsf{s})$ is recovered by its factorization with the distributions $
    \p_\mathsf{s}(w^\mathsf{s}|x^\mathsf{s})$, $\p_\mathsf{s}(y^\mathsf{s}|x^\mathsf{s},w^\mathsf{s})$, $\p_\mathsf{s}(z^\mathsf{s}|x^\mathsf{s}$, $y^\mathsf{s},w^\mathsf{s}), \p_\mathsf{s}(y^\mathsf{s}|w^\mathsf{s},z^\mathsf{s})$, and $\p(z^\mathsf{s})$. Adaptively learning these distributions in a federated setting is the main task of our work. This completes the proof.
\end{proof}

\section{Computing CATE, local ATE, and global ATE for CausalRFF}
This section gives details on how to compute CATE, local ATE and global ATE after training the model.
\subsection{Computing the CATE and local ATE}
After training the model, each source \textit{can} compute the CATE and the local ATE on for its own source and use it for itself.
\begin{align*}
E[y_i^\mathsf{s}|\doo(w_i^\mathsf{s}\!=\!w), x_i^\mathsf{s}] &=  \int E[y_i^\mathsf{s}|w_i^\mathsf{s}\!=\!w,z_i^\mathsf{s}]\p(z_i^\mathsf{s}|x_i^\mathsf{s})dz_i^\mathsf{s} \simeq \frac{1}{N}\sum_{l=1}^N f_y(w_i^\mathsf{s}\!=\!w,z_i^\mathsf{s}[l])
\end{align*}
where $f_y(w_i^\mathsf{s}\!=\!w,z_i^\mathsf{s}[l])$ is the mean function of $\p_\mathsf{s}(y_i^\mathsf{s}|w_i^\mathsf{s},z_i^\mathsf{s})$ and $\{{z}_i^\mathsf{s}[l]\}_{l=1}^N \overset{i.i.d.}{\sim}\p_\mathsf{s}(z_i^\mathsf{s}|x_i^\mathsf{s})$.

The problem is to draw $\{z_i^\mathsf{s}[l]\}_{l=1}^N$ from $\p_\mathsf{s}(z_i^\mathsf{s}|x_i^\mathsf{s})$. We observe that
\begin{align*}
    \p_\mathsf{s}(z_i^\mathsf{s}|x_i^\mathsf{s}) = \sum_{w_i^\mathsf{s}\in\{0,1\}}\int \p_\mathsf{s}(z_i^\mathsf{s}|x_i^\mathsf{s}, y_i^\mathsf{s},w_i^\mathsf{s})\p_\mathsf{s}(y_i^\mathsf{s}|x_i^\mathsf{s},w_i^\mathsf{s})\p_\mathsf{s}(w_i^\mathsf{s}|x_i^\mathsf{s})\,dy_i^\mathsf{s}.
\end{align*}
Hence, to draw samples, we proceed in the following steps:
\begin{itemize}
    \item[(1)] Draw a sample of $w_i^\mathsf{s}$ from $\p_\mathsf{s}(w_i^\mathsf{s}|x_i^\mathsf{s})$.
    \item[(2)] Substitute the above sample of $w_i^\mathsf{s}$ to $\p_\mathsf{s}(y_i^\mathsf{s}|x_i^\mathsf{s},w_i^\mathsf{s})$.
    \item[(3)] Draw a sample of $y_i^\mathsf{s}$ from $\p_\mathsf{s}(y_i^\mathsf{s}|x_i^\mathsf{s},w_i^\mathsf{s})$.
    \item[(4)] Substitute the above sample of $y_i^\mathsf{s}$ to $\p_\mathsf{s}(z_i^\mathsf{s}|x_i^\mathsf{s}, y_i^\mathsf{s},w_i^\mathsf{s})$.
    \item[(5)] Draw a sample of $z_i^\mathsf{s}$ from $\p_\mathsf{s}(z_i^\mathsf{s}|x_i^\mathsf{s}, y_i^\mathsf{s},w_i^\mathsf{s})$.
\end{itemize}
The density function of $\p_\mathsf{s}(y_i^\mathsf{s}|x_i^\mathsf{s},w_i^\mathsf{s})$ and $\p_\mathsf{s}(w_i^\mathsf{s}|x_i^\mathsf{s})$ are available after training the model. 
As described in the main text, there are two options to draw from $\p_\mathsf{s}(z_i^\mathsf{s}|x_i^\mathsf{s}, y_i^\mathsf{s},w_i^\mathsf{s})$. The first option is to draw from $\q(x_i^\mathsf{s})$ sine it approximates $\p_\mathsf{s}(z_i^\mathsf{s}|x_i^\mathsf{s}, y_i^\mathsf{s},w_i^\mathsf{s})$. The second option is to use Metropolis-Hastings algorithm with independent sampler \citep{liu1996metropolized}. For the second option, we have that
\begin{align*}
    \p_\mathsf{s}(z_i^\mathsf{s}|x_i^\mathsf{s}, y_i^\mathsf{s},w_i^\mathsf{s}) \propto \p_\mathsf{s}(y_i^\mathsf{s}|z_i^\mathsf{s},w_i^\mathsf{s})\p_\mathsf{s}(w_i^\mathsf{s}|z_i^\mathsf{s})\p_\mathsf{s}(x_i^\mathsf{s}|z_i^\mathsf{s})\p(z_i^\mathsf{s}).
\end{align*}
Hence, it can be used to compute the acceptance probability of interest. Note that the second option would give more exact samples since it further filters the samples based on the exact acceptance probability.

The above would help estimate the CATE given $x_i^\mathsf{s}$. The local ATE is the average of CATE of individuals in a source $\mathsf{s}$. These quantities can be estimated in a local source's machine. We show how to compute the global ATE in the next section.

\subsection{Computing the global ATE from local ATE of each Source}
To compute a global ATE, the server would collect all the local ATE in each source and then compute their weighted average. For example, suppose that we have three sources whose local ATE values are $7.0$, $8.5$, and $6.8$. \textcolor{black}{These local ATEs are averaged over 10, 5, and 12 individuals, in that order}. Then, the global ATE is given as follows:
\begin{align*}
    \text{global ATE} = \frac{10\times 7.0 + 8\times 8.5 + 12\times 6.8}{10 + 8 + 12} = 7.32.
\end{align*}
Since each source only shares their local ATE and the number of individuals, it does not leak any sensitive information about the individuals.

\section{Derivation of the loss functions}
\label{ap:derivation-loss}
In this section, we present the loss functions and the form of functions that modulate the desired distributions.
\subsection{Learning distributions involving latent confounder}
\label{sec:learn-dist-latent-confounder}

The ELBO of the log marginal likelihood has the following expression
\begin{align*}
    \log&\p(\mathbf{x}, \mathbf{y},\mathbf{w}) = \log\int \p(\mathbf{x}, \mathbf{y},\mathbf{w},\mathbf{z})d\mathbf{z} \\
    &\ge \int \q(\mathbf{z})\log\frac{\p(\mathbf{x}, \mathbf{y},\mathbf{w},\mathbf{z})}{\q(\mathbf{z})}d\mathbf{z}\\
    &=\sum_{\mathsf{s}\in\bm{\mathcal{S}}}\sum_{i=1}^{n_\mathsf{s}}\Big(\e_q\big[\log\p_\mathsf{s}(y_i^\mathsf{s}|w_i^\mathsf{s},z_i^\mathsf{s}) + \log\p_\mathsf{s}(w_i^\mathsf{s}|z_i^\mathsf{s}) + \log\p_\mathsf{s}(x_i^\mathsf{s}|z_i^\mathsf{s})\big] - \text{KL}[\q(z_i^\mathsf{s})\|\p(z_i^\mathsf{s})]\Big) =\vcentcolon \mathcal{L}.
\end{align*}
Using the complete dataset $\tilde{\mathsf{D}}^\mathsf{s} = \bigcup_{l=1}^{M}\big\{( w_i^\mathsf{s}, y_i^\mathsf{s}, x_i^\mathsf{s}, z_i^\mathsf{s}[l])\big\}_{i=1}^{n_\mathsf{s}}, \forall\mathsf{s} \in \bm{\mathcal{S}}$, we minimize the following loss function ${J}$:
\begin{align*}
    J = \widehat{\mathcal{L}} + \sum_{c\in\mathcal{A}} R(f_c), \qquad \mathcal{A} = \{y_0, y_1, q_0, q_q, x, w\},
\end{align*}
where $\widehat{\mathcal{L}}$ is the empirical loss function obtained from the negative of $\mathcal{L}$. 
In the following, we find the form of $f_c$ based on the representer theorem.

We further define $f_x = [f_{x,1},\!...,f_{x,d_x}]$, where $f_{x,d}$ is a function taking $z_i^\mathsf{s}$ as input and mapping it to a real value in $\mathbb{R}$. Similarly,  $f_{q_0} = [f_{q_0,1},\!...,f_{q_0,d_z}]$ and $f_{q_1} = [f_{q_1,1},\!...,f_{q_1,d_z}]$.

		Let $\mathcal{H}_c$ $(c \in \mathcal{A})$ be a reproducing Kernel Hilbert space (RKHS) and $\kappa_c(\cdot,\cdot)$ be kernel function associated with $\mathcal{H}_c$. We define $\mathcal{B}_c$ as follows:
		\begin{align*}
		&\mathcal{B}_{y_0} = \texttt{span}\big\{\kappa_{y0}(\cdot,z_i^\mathsf{s}[l]), \text{ where } \mathsf{s} \in \bm{\mathcal{S}}; i=1,\!...,n_\mathsf{s};l=1,\!...,M\big\},\\
		&\mathcal{B}_{y_1} = \texttt{span}\big\{\kappa_{y1}(\cdot,z_i^\mathsf{s}[l]), \text{ where }\mathsf{s} \in \bm{\mathcal{S}}; i=1,\!...,n_\mathsf{s};l=1,\!...,M\big\},\\
		&\mathcal{B}_x = \texttt{span}\left\{\kappa_x(\cdot,z_i^\mathsf{s}[l]), \text{ where }\mathsf{s} \in \bm{\mathcal{S}};i=1,\!...,n_\mathsf{s};l=1,\!...,M\right\},\\
		&\mathcal{B}_w = \texttt{span}\left\{\kappa_w(\cdot,z_i^\mathsf{s}[l]), \text{ where }\mathsf{s} \in \bm{\mathcal{S}};i=1,\!...,n_\mathsf{s};l=1,\!...,M\right\},\\
		&\mathcal{B}_{q_0} = \texttt{span}\left\{\kappa_{q0}(\cdot,[x_i^\mathsf{s},y_i^\mathsf{s}]), \text{ where }\mathsf{s} \in \bm{\mathcal{S}};i=1,\!...,n_\mathsf{s}\right\},\\
		&\mathcal{B}_{q_1} = \texttt{span}\left\{\kappa_{q1}(\cdot,[x_i^\mathsf{s},y_i^\mathsf{s}]), \text{ where }\mathsf{s} \in \bm{\mathcal{S}};i=1,\!...,n_\mathsf{s}\right\}.
		\end{align*} 
		We posit the following regularizers:
        \begin{align*}
            R(f_{y0}) = \mathsf{reg\_factor}_{y0}\times\|f_{y0}\|_{\mathcal{H}_{y0}}^2,\quad R(f_x) = \sum_{d=1}^{d_x}\mathsf{reg\_factor}_{x,d}\times\|f_{x,d}\|_{\mathcal{H}_x}^2\quad (d=1,\!...,d_x).
        \end{align*}
        The regularizers $R(f_{y1})$ and $R(f_w)$ are similar to that of $R(f_{y0})$, and $R(f_{q0})$, $R(f_{q1})$ are similar to that of $R(f_x)$.		
		
		We see that $\mathcal{B}_c$ is a subspace of $\mathcal{H}_c$. We project $f_{y0}$, $f_{y1}$, $f_w$, $f_{x,d}$ ($d=1,\!...,d_x$), $f_{q0,d}$  ($d=1,\!...,d_z$) and $f_{q1,d}$  ($d=1,\!...,d_z$) onto the subspaces $\mathcal{B}_{y0}$, $\mathcal{B}_{y1}$, $\mathcal{B}_w$, $\mathcal{B}_x$, $\mathcal{B}_{q0}$ and $\mathcal{B}_{q1}$, respectively, and obtain $f_{y0}'$, $f_{y1}'$, $f_w'$, $f_{x,d}'$, $f_{q0,d}'$ and $f_{q1,d}'$. Next, we also project them onto the perpendicular spaces of $\mathcal{B}_{(\cdot)}$ to obtain $f_{y_0}^{\bot}$, $f_{y_1}^{\bot}$, $f_w^{\bot}$, $f_{x,d}^{\bot}$, $f_{q_0,d}^{\bot}$ and $f_{q_1,d}^{\bot}$. 
		
		Note that $f_{(\cdot)} = f_{(\cdot)}'+ f_{(\cdot)}^{\bot}$. Hence, $\|f_{(\cdot)}\|_{\mathcal{H}_{(\cdot)}}^2 = \|f_{(\cdot)}'\|_{\mathcal{H}_{(\cdot)}}^2 + \|f_{(\cdot)}^{\bot}\|_{\mathcal{H}_{(\cdot)}}^2 \ge \|f_{(\cdot)}'\|_{\mathcal{H}_{(\cdot)}}^2$, which implies\\that $\mathsf{reg\_factor}_{(\cdot)}\times\|f_{(\cdot)}\|_{\mathcal{H}_{(\cdot)}}^2$ is minimized if $f_{(\cdot)}$ is in its subspace $\mathcal{B}_{(\cdot)}$.\hfill~~~~~\textbf{(I)}
		\\[0.2cm]
		In addition, due to the reproducing property, we have 
\begin{align*}
		f_{y_0}(z_i^\mathsf{s}[l]) &= \big\langle f_{y_0},\kappa_{y_0}(\cdot, z_i^\mathsf{s}[l]) \big\rangle_{\mathcal{H}_y} 
= \big\langle f_{y_0}',\kappa_{y_0}(\cdot, z_i^\mathsf{s}[l]) \big\rangle_{\mathcal{H}_y} + \big\langle f_{y_0}^{\bot},\kappa_{y_0}(\cdot, z_i^\mathsf{s}[l]) \big\rangle_{\mathcal{H}_y}
= f_{y_0}'(z_i^\mathsf{s}[l]).
		\end{align*}
		Similarly, we also have $f_{y1}( z_i^\mathsf{d}[l])=f_{y1}'( z_i^\mathsf{d}[l])$, $f_w(z_i^\mathsf{d}[l]) = f_w'(z_i^\mathsf{d}[l])$, $f_{x,d}(z_i^l) = f_{x,d}'(z_i^\mathsf{d}[l])$,  $f_{q0,d}(y_i^\mathsf{d}, x_i^\mathsf{d}) = f_{q0,d}'(y_i^\mathsf{d}, x_i^\mathsf{d})$ and $f_{q_1,d}(y_i^\mathsf{d}, x_i^\mathsf{d}) = f_{q_1,d}'(y_i^\mathsf{d}, x_i^\mathsf{d})$. Hence,
		\\~~~~~~~~~~~$\widehat{\mathcal{L}}(f_{y0}, f_{y1}, f_{q0}, f_{q1}, f_x, f_w) = \widehat{\mathcal{L}}(f_{y0}', f_{y1}', f_{q0}', f_{q1}', f_x', f_w').$ \hfill\textbf{(II)}
		\\\textbf{(I)} and \textbf{(II)} imply that $f_{y0}, f_{y1}, f_{q0,d}, f_{q1,d}, f_{x,d}, f_w$ are the weighted sum of elements in their corresponding subspace. Hence,
\begin{align*}
    \Aboxed{f_c(\bm{u}^\mathsf{s}) = \sum_{\mathsf{v}\in\bm{\mathcal{S}}}\sum_{j=1}^{n_\mathsf{v}\times M}\kappa(\bm{u}^\mathsf{s}, \bm{u}_j^\mathsf{v})\bm{\alpha}_j^\mathsf{v}.}
\end{align*}
Using this form with the adaptive kernel and Random Fourier Feature described in the main text (Section~4.1), we obtain the desired model.
\subsection{Learning auxiliary distributions}
The derivation of $J_w$, $J_y$ and the form of functions modulated the auxiliary distributions are similar to those of $J$ as detailed in Section~\ref{sec:learn-dist-latent-confounder}. The difference is that the empirical loss functions are obtained from the negative log-likelihood instead of the ELBO.

\section{Spectral distribution of some popular kernels}
Table~\ref{tab:spec-dens} (adopted from \citet{milton2019spatial}) presents some popular kernels and their associated spectral density $s(\bm{\omega})$. Those density functions are needed to draw samples of $\bm{\omega}$ for Random Fourier Features presented in Section~4 of the main text. In our experiments, we used Gaussian kernel.
\begin{table}[!ht]
\caption{Some popular kernels and their associated spectral density. Note that $K_\nu(\cdot)$ denotes the modified Bessel function of the second kind, $\Gamma(\cdot)$ is the gamma function.
}
\label{tab:spec-dens}
\centering
\setlength{\tabcolsep}{5pt}
\scriptsize
\begin{tabular}{@{}lcc@{}}
\toprule
\textbf{Kernel}               & \makecell[c]{\textbf{Kernel function}, $\displaystyle k(x_1 -x_2)$} & \makecell[c]{\textbf{Spectral density}, $\displaystyle s(\bm{\omega})$} \\ \midrule
Gaussian           & $\displaystyle\exp\left(-\frac{\|x_1-x_2\|_2^2}{2\ell^2}\right)$       &  $\displaystyle\left(\frac{2\pi}{\ell^2}\right)^{\frac{-d}{2}}\exp\left(-\frac{\ell^2\|\bm{\omega}\|_2^2}{2}\right)$                \\
Laplacian & $\displaystyle\exp\Big(-\ell\|x_1-x_2\|_1\Big)$       & $\displaystyle\left(\frac{2}{\pi}\right)^{\frac{d}{2}}\prod_{i=1}^d\frac{\ell}{\ell^2 + \omega_i^2}$                  \\ 
Matérn & $\displaystyle\frac {2^{1-\nu }}{\Gamma (\nu )}\Bigg ({\sqrt {2\nu }}{\frac {\|x_1-x_2\|_2}{\ell }}\Bigg )^{\nu }K_{\nu }{\Bigg (}{\sqrt {2\nu }}{\frac {\|x_1-x_2\|_2}{\ell }}{\Bigg )}
$       & $\displaystyle {\frac {2^{d}\pi ^{\frac {d}{2}}\Gamma (\nu +{\frac {d}{2}})(2\nu )^{\nu }}{\Gamma (\nu )\ell ^{2\nu }}}\left({\frac {2\nu }{\ell^2}}+4\pi ^{2}\|\bm{\omega}\|_2^2\right)^{-\left(\nu +{\frac {d}{2}}\right)}$                  \\ 
\bottomrule
\end{tabular}
\end{table}

\section{Proof of Lemma~\ref{lem:minimax-bounds1}}
\label{ap:proof-lem-1}

Let $\bm{\mathcal{S}}_{\setminus\mathsf{s}}:= \bm{\mathcal{S}}\setminus\{\mathsf{s}\}$. The model is summarized as follows:
\begin{align*}
    \p(z_i^\mathsf{s}) &= \mathsf{N}(0,\sigma_z^2\mathbf{I}_{d_z}),\\
    \p(w_i^\mathsf{s}|z_i^\mathsf{s}) &= \mathsf{Bern}\Big(\varphi\Big(\Big(\theta_w^\mathsf{s} + \sum_{\mathsf{v}\in\bm{\mathcal{S}}_{\setminus\mathsf{s}}}\lambda^{\mathsf{s},\mathsf{v}}\theta_w^\mathsf{v}\Big)^\top \phi(z_i^\mathsf{s})\Big)\Big),\\
    \p(y_i^\mathsf{s}|w_i^\mathsf{s}, z_i^\mathsf{s}) &= \mathsf{N}\Big(\Big(w_i^\mathsf{s}\Big(\theta_{y1}^\mathsf{s} + \sum_{\mathsf{v}\in\bm{\mathcal{S}}_{\setminus\mathsf{s}}}\lambda^{\mathsf{s},\mathsf{v}}\theta_{y1}^\mathsf{v}\Big) + (1-w_i^\mathsf{s})\Big(\theta_{y0}^\mathsf{s} + \sum_{\mathsf{v}\in\bm{\mathcal{S}}_{\setminus\mathsf{s}}}\lambda^{\mathsf{s},\mathsf{v}}\theta_{y0}^\mathsf{v}\Big)\Big)^\top \phi(z_i^\mathsf{s}), \sigma_y^2\Big),\\
    \p(x_i^\mathsf{s}|z_i^\mathsf{s}) &= \mathsf{N}\Big(\Big(\theta_x^\mathsf{s} + \sum_{\mathsf{v}\in\bm{\mathcal{S}}_{\setminus\mathsf{s}}}\lambda^{\mathsf{s},\mathsf{v}}\theta_x^\mathsf{v}\Big)^\top \phi(z_i^\mathsf{s}), \sigma_x^2\mathbf{I}_{d_x}\Big),
\end{align*}
where $z_i^{(\cdot)} \in \mathbb{R}^{d_z}$, $y_i^{(\cdot)} \in \mathbb{R}$, $w_i^{(\cdot)} \in \{0,1\}$, $x_i^{(\cdot)} \in \mathbb{R}^{d_x}$, $\lambda > 0$.

Let $\bm{\uptheta} = \{\theta_w^\mathsf{s},\theta_{y0}^\mathsf{s},\theta_{y1}^\mathsf{s}, \theta_x^\mathsf{s}\}_{\mathsf{s}\in\bm{\mathcal{S}}}$. 
Let $\mathcal{V}_w$, $\mathcal{V}_{y0}$, $\mathcal{V}_{y1}$, $\mathcal{V}_x$ be $1/(2\sqrt{m})$-packing of the unit $\|\cdot\|_2$-balls with cardinality at least $(2\sqrt{m})^{2B}$, $(2\sqrt{m})^{2B}$, $(2\sqrt{m})^{2B}$, $(2\sqrt{m})^{2Bd_x}$, respectively. Let $\mathcal{V}^\mathsf{s} =\delta (\mathcal{V}_w\times \mathcal{V}_{y0} \times \mathcal{V}_{y1} \times \mathcal{V}_x)$ and $\mathcal{V} = \mathcal{V}^{\mathsf{s}_1} \times \mathcal{V}^{\mathsf{s}_2} \times\!...\times\mathcal{V}^{\mathsf{s}_m}$. We see that
\begin{align*}
    |\mathcal{V}|\ge (2\sqrt{m})^{2mB(d_x+3)}.
\end{align*}
In the following, we derive the minimax bound:

\begin{proof}
We have that 
\begin{align*}
    \|\bm{\uptheta}_1 - \bm{\uptheta}_2\|_2 &= \sqrt{\sum_{\mathsf{s}\in\bm{\mathcal{S}}}\sum_{c\in\bm{\mathcal{A}}}\|(\theta_c^\mathsf{s})_1 - (\theta_c^\mathsf{s})_2\|_2^2} \ge \sqrt{\sum_{\mathsf{s}\in\bm{\mathcal{S}}}4\left(\frac{\delta}{2\sqrt{m}}\right)^2} = \delta.
\end{align*}
The marginal distribution
\begin{align*}
    \p_{\bm{\uptheta}}(w,y,x) &= \int \p_{\bm{\uptheta}}(w,y,x, z)dz = \int\p_{\bm{\uptheta}}(y|w,z)\p_{\bm{\uptheta}}(w|z)\p_{\bm{\uptheta}}(x|z)\p(z)dz.
\end{align*}
Moreover, we have that
\begin{align*}
    &D_{\textrm{KL}}(\p^n_{\bm{\uptheta}_1}\,\|\,\p^n_{\bm{\uptheta}_2}) = \sum_{\mathsf{s}\in\bm{\mathcal{S}}}D_{\textrm{KL}}(\p^{n_\mathsf{s}}_{\bm{\uptheta}_1}\,\|\,\p^{n_\mathsf{s}}_{\bm{\uptheta}_2}).
\end{align*}

We divide the proof into three parts \textbf{(I)}, \textbf{(II)}, and \textbf{(III)}:

\textbf{(I) The upper bound of $D_{\textrm{KL}}(\p^{n_\mathsf{s}}_{\theta_1}\,\|\,\p^{n_\mathsf{s}}_{\theta_2})$}

Since the data is independent, we have that
\begin{align*}
     &D_{\textrm{KL}}(\p^{n_\mathsf{s}}_{{\bm{\uptheta}}_1}\,\|\,\p^{n_\mathsf{s}}_{{\bm{\uptheta}}_2}) = n_\mathsf{s}D_{\textrm{KL}}(\p^1_{{\bm{\uptheta}}_1}\,\|\,\p^1_{{\bm{\uptheta}}_2})\\
    &\le\! n_\mathsf{s}\int \!D_{\textrm{KL}}\left(\p_{{\bm{\uptheta}}_1}(y|w,z)\p_{{\bm{\uptheta}}_1}(w|z)\p_{{\bm{\uptheta}}_1}(x|z)\Big\|\p_{{\bm{\uptheta}}_2}(y|w,z')\p_{{\bm{\uptheta}}_2}(w|z')\p_{{\bm{\uptheta}}_2}(x|z')\right)\p(z)\p(z')dzdz'\\
&= n_\mathsf{s}\int \bigg[\p_{{\bm{\uptheta}}_1}(w=0|z) D_{\textnormal{KL}}\big[\p_{{\bm{\uptheta}}_1}(y|w=0,z) \big\|\p_{{\bm{\uptheta}}_2}(y|w=0,z')\big] \\
    &\qquad\qquad\qquad+ \p_{{\bm{\uptheta}}_1}(w=1|z) D_{\textnormal{KL}}\big[\p_{{\bm{\uptheta}}_1}(y|w=1,z) \big\|\p_{{\bm{\uptheta}}_2}(y|w=1,z')\big]\\
     &\qquad\qquad\qquad+ D_{\textnormal{KL}}\big[\p_{{\bm{\uptheta}}_1}(w|z)\big\|\p_{{\bm{\uptheta}}_2}(w|z')\big] + D_{\textnormal{KL}}\big[\p_{{\bm{\uptheta}}_1}(x|z)\big\|\p_{{\bm{\uptheta}}_2}(x|z')\big]\bigg]\p(z)\p(z')dzdz'.
\end{align*}
In the following, we find the upper bound of each component.

$\diamond$ \emph{\underline{Upper bound of the first and second component}}
\begin{align*}
    \p_{\bm{\uptheta}_1}(w=0|z) &D_{\textnormal{KL}}\big[\p_{\bm{\uptheta}_1}(y|w=0,z) \big\|\p_{\bm{\uptheta}_2}(y|w=0,z')\big]\\
    &\le\frac{1}{2\sigma_y^2}\Big(\Big((\theta_{y0}^\mathsf{s})_1 + \sum_{\mathsf{v}\in\bm{\mathcal{S}}_{\setminus\mathsf{s}}}\lambda^{\mathsf{s},\mathsf{v}}(\theta_{y0}^\mathsf{v})_1\Big)^\top\phi(z) - \Big((\theta_{y0}^\mathsf{s})_2 + \sum_{\mathsf{v}\in\bm{\mathcal{S}}_{\setminus\mathsf{s}}}\lambda^{\mathsf{s},\mathsf{v}}(\theta_{y0}^\mathsf{v})_2\Big)^\top\phi(\mathsf{z'})\Big)^2\\
&\le \frac{8B^2\delta^2(1+\sum_{\mathsf{v}\in\bm{\mathcal{S}}_{\setminus\mathsf{s}}}\lambda^{\mathsf{s},\mathsf{v}})^2}{\sigma_y^2}.
\end{align*}
Similarly, we also have
\begin{align*}
    &\p_{\bm{\uptheta}_1}(w=1|z) D_{\textnormal{KL}}\big[\p_{\bm{\uptheta}_1}(y|w=1,z) \big\|\p_{\bm{\uptheta}_2}(y|w=1,z')\big] \le \frac{8B^2\delta^2(1+\sum_{\mathsf{v}\in\bm{\mathcal{S}}_{\setminus\mathsf{s}}}\lambda^{\mathsf{s},\mathsf{v}})^2}{\sigma_y^2}.
\end{align*}
$\diamond$ \emph{\underline{Upper bound of the third component}}
\begin{align*}
    &D_{\textnormal{KL}}\big[\p_{\bm{\uptheta}_1}(w|z)\big\|\p_{\bm{\uptheta}_2}(w|z')\big]\\
    &=\varphi\Big(\Big((\theta_w^\mathsf{s})_1 + \sum_{\mathsf{v}\in\bm{\mathcal{S}}_{\setminus\mathsf{s}}}\lambda^{\mathsf{s},\mathsf{v}}(\theta_w^\mathsf{v})_1\Big)^\top \phi(z)\Big)\log\frac{\varphi\Big(\Big((\theta_w^\mathsf{s})_1 + \sum_{\mathsf{v}\in\bm{\mathcal{S}}_{\setminus\mathsf{s}}}\lambda^{\mathsf{s},\mathsf{v}}(\theta_w^\mathsf{v})_1\Big)^\top \phi(z)\Big)}{\varphi\Big(\Big((\theta_w^\mathsf{s})_2 + \sum_{\mathsf{v}\in\bm{\mathcal{S}}_{\setminus\mathsf{s}}}\lambda^{\mathsf{s},\mathsf{v}}(\theta_w^\mathsf{v})_2\Big)^\top \phi(z')\Big)} \\
    &\quad+ \varphi\Big(-\Big((\theta_w^\mathsf{s})_1 + \sum_{\mathsf{v}\in\bm{\mathcal{S}}_{\setminus\mathsf{s}}}\lambda^{\mathsf{s},\mathsf{v}}(\theta_w^\mathsf{v})_1\Big)^\top \phi(z)\Big)\log\frac{\varphi\Big(-\Big((\theta_w^\mathsf{s})_1 + \sum_{\mathsf{v}\in\bm{\mathcal{S}}_{\setminus\mathsf{s}}}\lambda^{\mathsf{s},\mathsf{v}}(\theta_w^\mathsf{v})_1\Big)^\top \phi(z)\Big)}{\varphi\Big(-\Big((\theta_w^\mathsf{s})_2 + \sum_{\mathsf{v}\in\bm{\mathcal{S}}_{\setminus\mathsf{s}}}\lambda^{\mathsf{s},\mathsf{v}}(\theta_w^\mathsf{v})_2\Big)^\top \phi(z')\Big)}.
\end{align*}
For the first component,
\begin{align*}
    \varphi\Big(\Big(&(\theta_w^\mathsf{s})_1 + \sum_{\mathsf{v}\in\bm{\mathcal{S}}_{\setminus\mathsf{s}}}\lambda^{\mathsf{s},\mathsf{v}}(\theta_w^\mathsf{v})_1\Big)^\top \phi(\mathbf{z})\Big)\log\frac{\varphi\Big(\Big((\theta_w^\mathsf{s})_1 + \sum_{\mathsf{v}\in\bm{\mathcal{S}}_{\setminus\mathsf{s}}}\lambda^{\mathsf{s},\mathsf{v}}(\theta_w^\mathsf{v})_1\Big)^\top \phi(\mathbf{z})\Big)}{\varphi\Big(\Big((\theta_w^\mathsf{s})_2 + \sum_{\mathsf{v}\in\bm{\mathcal{S}}_{\setminus\mathsf{s}}}\lambda^{\mathsf{s},\mathsf{v}}(\theta_w^\mathsf{v})_2\Big)^\top \phi(\mathbf{z}')\Big)}\\
&\le  \Big|\log \Big(1+ e^{-\big((\theta_w^\mathsf{s})_2 + \sum_{\mathsf{v}\in\bm{\mathcal{S}}_{\setminus\mathsf{s}}}\lambda^{\mathsf{s},\mathsf{v}}(\theta_w^\mathsf{v})_2\big)^\top \phi(\mathbf{z})}\Big) - \log\Big(1+e^{-\big((\theta_w^\mathsf{s})_1 + \sum_{\mathsf{v}\in\bm{\mathcal{S}}_{\setminus\mathsf{s}}}\lambda^{\mathsf{s},\mathsf{v}}(\theta_w^\mathsf{v})_1\big)^\top \phi(\mathbf{z}')}\Big)\Big|\\
&\le \Big\|(\theta_w^\mathsf{s})_1 + \sum_{\mathsf{v}\in\bm{\mathcal{S}}_{\setminus\mathsf{s}}}\lambda^{\mathsf{s},\mathsf{v}}(\theta_w^\mathsf{v})_1\Big\|_2\| \phi(\mathbf{z})\|_2 + \Big\|(\theta_w^\mathsf{s})_2 + \sum_{\mathsf{v}\in\bm{\mathcal{S}}_{\setminus\mathsf{s}}}\lambda^{\mathsf{s},\mathsf{v}}(\theta_w^\mathsf{v})_2\Big\|_2 \|\phi(\mathbf{z}')\|_2\\
&\le \Big(\delta + \sum_{\mathsf{v}\in\bm{\mathcal{S}}_{\setminus\mathsf{s}}}\lambda^{\mathsf{s},\mathsf{v}}\delta\Big)\| \phi(\mathbf{z})\|_2 + \Big(\delta + \sum_{\mathsf{v}\in\bm{\mathcal{S}}_{\setminus\mathsf{s}}}\lambda^{\mathsf{s},\mathsf{v}}\delta\Big) \|\phi(\mathbf{z}')\|_2\\
&\le 4B\delta\Big(1 + \sum_{\mathsf{v}\in\bm{\mathcal{S}}_{\setminus\mathsf{s}}}\lambda^{\mathsf{s},\mathsf{v}}\Big).
\end{align*}
Similarly, we also have
\begin{align*}
    \varphi\Big(-\Big((\theta_w^\mathsf{s})_1 &+ \sum_{\mathsf{v}\in\bm{\mathcal{S}}_{\setminus\mathsf{s}}}\lambda^{\mathsf{s},\mathsf{v}}(\theta_w^\mathsf{v})_1\Big)^\top \phi(z)\Big)\log\frac{\varphi\Big(-\Big((\theta_w^\mathsf{s})_1 + \sum_{\mathsf{v}\in\bm{\mathcal{S}}_{\setminus\mathsf{s}}}\lambda^{\mathsf{s},\mathsf{v}}(\theta_w^\mathsf{v})_1\Big)^\top \phi(z)\Big)}{\varphi\Big(-\Big((\theta_w^\mathsf{s})_2 + \sum_{\mathsf{v}\in\bm{\mathcal{S}}_{\setminus\mathsf{s}}}\lambda^{\mathsf{s},\mathsf{v}}(\theta_w^\mathsf{v})_2\Big)^\top \phi(z')\Big)} \\
    &\qquad\le 4B\delta\Big(1 + \sum_{\mathsf{v}\in\bm{\mathcal{S}}_{\setminus\mathsf{s}}}\lambda^{\mathsf{s},\mathsf{v}}\Big).
\end{align*}
Thus,
\begin{align*}
    D_{\textnormal{KL}}\big[\p_{\bm{\uptheta}_1}(w|z)\big\|\p_{\bm{\uptheta}_2}(w|z')\big] \le 8B\delta\Big(1 + \sum_{\mathsf{v}\in\bm{\mathcal{S}}_{\setminus\mathsf{s}}}\lambda^{\mathsf{s},\mathsf{v}}\Big).
\end{align*}
$\diamond$ \emph{\underline{Upper bound of the fourth component}}
\begin{align*}
    D_{\textnormal{KL}}&\big[\p_{\bm{\uptheta}_1}(x|z)\big\|\p_{\bm{\uptheta}_2}(x|z')\big]\\
    &=\frac{1}{2\sigma_x^2}\Big\|\Big((\theta_x^\mathsf{s})_1 + \sum_{\mathsf{v}\in\bm{\mathcal{S}}_{\setminus\mathsf{s}}}\lambda^{\mathsf{s},\mathsf{v}}(\theta_x^\mathsf{v})_1\Big)^\top\phi(z) - \Big((\theta_x^\mathsf{s})_2 + \sum_{\mathsf{v}\in\bm{\mathcal{S}}_{\setminus\mathsf{s}}}\lambda^{\mathsf{s},\mathsf{v}}(\theta_x^\mathsf{v})_2\Big)^\top\phi(z')\Big\|_2^2\\
    &\le\frac{1}{2\sigma_x^2}\Big(\Big\|\Big((\theta_x^\mathsf{s})_1 + \sum_{\mathsf{v}\in\bm{\mathcal{S}}_{\setminus\mathsf{s}}}\lambda^{\mathsf{s},\mathsf{v}}(\theta_x^\mathsf{v})_1\Big)^\top\phi(z)\Big\|_2  + \Big\|\Big((\theta_x^\mathsf{s})_2 + \sum_{\mathsf{v}\in\bm{\mathcal{S}}_{\setminus\mathsf{s}}}\lambda^{\mathsf{s},\mathsf{v}}(\theta_x^\mathsf{v})_2\Big)^\top\phi(z')\Big\|_2\Big)^2\\
&\le\frac{8B^2\delta^2\Big(1+\sum_{\mathsf{v}\in\bm{\mathcal{S}}_{\setminus\mathsf{s}}}\lambda^{\mathsf{s},\mathsf{v}}\Big)^2}{\sigma_x^2}.
\end{align*}
\textbf{(II) Combining the results}

From the above upper bound of each of the components, we obtain
\begin{align*}
    D_{\textrm{KL}}(\p^{n_\mathsf{s}}_{\bm{\uptheta}_1}\,\|\,\p^{n_\mathsf{s}}_{\bm{\uptheta}_2})&\le n_\mathsf{s}\int \bigg[\frac{16B^2\delta^2(1+\sum_{\mathsf{v}\in\bm{\mathcal{S}}_{\setminus\mathsf{s}}}\lambda^{\mathsf{s},\mathsf{v}})^2}{\sigma_y^2} + 8B\delta\Big(1 + \sum_{\mathsf{v}\in\bm{\mathcal{S}}_{\setminus\mathsf{s}}}\lambda^{\mathsf{s},\mathsf{v}}\Big) \\
    &\qquad+ \frac{8B^2\delta^2(1+\sum_{\mathsf{v}\in\bm{\mathcal{S}}_{\setminus\mathsf{s}}}\lambda^{\mathsf{s},\mathsf{v}})^2}{\sigma_x^2}\bigg]\p(z)\p(z')dzdz'\\
    &= n_\mathsf{s} \bigg[\bigg(\frac{1}{\sigma_y^2} + \frac{1}{2\sigma_x^2}\bigg)16B^2\delta^2\Big(1+\sum_{\mathsf{v}\in\bm{\mathcal{S}}_{\setminus\mathsf{s}}}\lambda^{\mathsf{s},\mathsf{v}}\Big)^2 + 8B\delta\Big(1+\sum_{\mathsf{v}\in\bm{\mathcal{S}}_{\setminus\mathsf{s}}}\lambda^{\mathsf{s},\mathsf{v}}\Big)\bigg].
\end{align*}

\noindent\textbf{(III) The minimax lower bound}

We have that
\begin{align*}
    D_{\textrm{KL}}(\p^n_{\bm{\uptheta}_1}\,\|\,\p^n_{\bm{\uptheta}_2}) &= \sum_{\mathsf{s}\in\bm{\mathcal{S}}}D_{\textrm{KL}}(\p^{n_\mathsf{s}}_{\bm{\uptheta}_1}\,\|\,\p^{n_\mathsf{s}}_{\bm{\uptheta}_2})\\
    &\le \sum_{\mathsf{s}\in\bm{\mathcal{S}}}n_\mathsf{s} \bigg[\bigg(\frac{1}{\sigma_y^2} + \frac{1}{2\sigma_x^2}\bigg)16B^2\delta^2\Big(1+\sum_{\mathsf{v}\in\bm{\mathcal{S}}_{\setminus\mathsf{s}}}\lambda^{\mathsf{s},\mathsf{v}}\Big)^2 + 8B\delta\Big(1+\sum_{\mathsf{v}\in\bm{\mathcal{S}}_{\setminus\mathsf{s}}}\lambda^{\mathsf{s},\mathsf{v}}\Big)\bigg].
\end{align*}
Consequently,
\begin{align*}
    &\inf_{\hat{\bm{\uptheta}}_n} \sup_{P\in\mathcal{P}} \mathbb{E}_P \left[ \|\hat{\bm{\uptheta}}_n\!-\!\bm{\uptheta}(P)\|_2 \right] \\
    &\ge\!\! \frac{\delta}{2}\!\left(\!1\!-\! \frac{\sum_{\mathsf{s}\in\bm{\mathcal{S}}}n_\mathsf{s} \bigg[\!\bigg(\frac{1}{\sigma_y^2} \!+\! \frac{1}{2\sigma_x^2}\bigg)16B^2\delta^2\Big(1\!+\!\sum_{\mathsf{v}\in\bm{\mathcal{S}}_{\setminus\mathsf{s}}}\lambda^{\mathsf{s},\mathsf{v}}\Big)^2 \!+\! 8B\delta\Big(1\!+\!\sum_{\mathsf{v}\in\bm{\mathcal{S}}_{\setminus\mathsf{s}}}\lambda^{\mathsf{s},\mathsf{v}}\Big)\!\bigg] \!+\! \log 2}{\log|\mathcal{V}|}\!\right)\\
    &\ge\!\! \frac{\delta}{2}\!\left(\!1\!-\! \frac{\sum_{\mathsf{s}\in\bm{\mathcal{S}}}n_\mathsf{s} \bigg[\!\bigg(\frac{1}{\sigma_y^2} \!+\! \frac{1}{2\sigma_x^2}\bigg)16B^2\delta^2\Big(1\!+\!\sum_{\mathsf{v}\in\bm{\mathcal{S}}_{\setminus\mathsf{s}}}\lambda^{\mathsf{s},\mathsf{v}}\Big)^2 \!+\! 8B\delta\Big(1\!+\!\sum_{\mathsf{v}\in\bm{\mathcal{S}}_{\setminus\mathsf{s}}}\lambda^{\mathsf{s},\mathsf{v}}\Big)\!\bigg] \!+\! \log 2}{2mB(d_x+3)\log(2\sqrt{m})}\!\right)\!\!.
\end{align*}
We choose $\delta=\frac{\sqrt{mB(d_x+3)}\log(2\sqrt{m})}{4B\sum_{\mathsf{s}\in\bm{\mathcal{S}}}n_\mathsf{s}\big(1+\sum_{\mathsf{v}\in\bm{\mathcal{S}}_{\setminus\mathsf{s}}}\lambda^{\mathsf{s},\mathsf{v}}\big)^2}$, then
\begin{align*}
   &1- \frac{\sum_{\mathsf{s}\in\bm{\mathcal{S}}}n_\mathsf{s} \bigg[\bigg(\frac{1}{\sigma_y^2} + \frac{1}{2\sigma_x^2}\bigg)16B^2\delta^2\Big(1+\sum_{\mathsf{v}\in\bm{\mathcal{S}}_{\setminus\mathsf{s}}}\lambda^{\mathsf{s},\mathsf{v}}\Big)^2 + 8B\delta\Big(1+\sum_{\mathsf{v}\in\bm{\mathcal{S}}_{\setminus\mathsf{s}}}\lambda^{\mathsf{s},\mathsf{v}}\Big)\bigg] + \log 2}{2mB(d_x+3)\log(2\sqrt{m})} \\
&\ge 1- \bigg(\frac{1}{\sigma_y^2} + \frac{1}{2\sigma_x^2}\bigg)\frac{\log(2\sqrt{m})}{2\sum_{\mathsf{s}\in\bm{\mathcal{S}}}n_\mathsf{s}\Big(1+\sum_{\mathsf{v}\in\bm{\mathcal{S}}_{\setminus\mathsf{s}}}\lambda^{\mathsf{s},\mathsf{v}}\Big)^2} - \frac{1}{\sqrt{mB(d_x+3)}} - \frac{1}{2mB(d_x+3)}\\
    &\ge 1- \bigg(\frac{1}{\sigma_y^2} + \frac{1}{2\sigma_x^2}\bigg)\frac{\log(2\sqrt{m})}{2\sum_{\mathsf{s}\in\bm{\mathcal{S}}}n_\mathsf{s}\Big(1+\sum_{\mathsf{v}\in\bm{\mathcal{S}}_{\setminus\mathsf{s}}}\lambda^{\mathsf{s},\mathsf{v}}\Big)^2} - \frac{1}{2} - \frac{1}{8}.
\end{align*}
If $\sum_{\mathsf{s}\in\bm{\mathcal{S}}}n_\mathsf{s}\Big(1+\sum_{\mathsf{v}\in\bm{\mathcal{S}}_{\setminus\mathsf{s}}}\lambda^{\mathsf{s},\mathsf{v}}\Big)^2 \ge 2\bigg(\frac{1}{\sigma_y^2} + \frac{1}{2\sigma_x^2}\bigg)\log(2\sqrt{m})$, then
\begin{align*}
    \inf_{\hat{\bm{\uptheta}}_n} \sup_{P\in\mathcal{P}} \mathbb{E}_P \left[ \|\hat{\bm{\uptheta}}_n\!-\!\bm{\uptheta}(P)\|_2 \right] &\ge \frac{1}{2}\times\frac{\sqrt{mB(d_x+3)}\log(2\sqrt{m})}{4B\sum_{\mathsf{s}\in\bm{\mathcal{S}}}n_\mathsf{s}\big(1+\sum_{\mathsf{v}\in\bm{\mathcal{S}}_{\setminus\mathsf{s}}}\lambda^{\mathsf{s},\mathsf{v}}\big)^2}\times \left(1- \frac{1}{4} - \frac{1}{2} - \frac{1}{8}\right)\\
    &= \frac{\sqrt{m(d_x+3)}\log(2\sqrt{m})}{64\sqrt{B}\sum_{\mathsf{s}\in\bm{\mathcal{S}}}n_\mathsf{s}\big(1+\sum_{\mathsf{v}\in\bm{\mathcal{S}}_{\setminus\mathsf{s}}}\lambda^{\mathsf{s},\mathsf{v}}\big)^2}.
\end{align*}

This completes the proof.
\end{proof}

\section{Proof of Lemma~\ref{lem:mimimax-bounds2}}
\label{ap:proof-lem-2}

The proof of Lemma~\ref{lem:mimimax-bounds2} is divided into two parts \textbf{(i)} and \textbf{(ii)}. We compute them separately:
\subsection{Proof of Part (i)}
\label{sec:proof-lem1-part-i}

We summarize the model as follows
\begin{align*}
    w^\mathsf{s} &\sim \mathsf{Bern}\Big(\varphi\Big(\Big(\psi^\mathsf{s} + \sum_{\mathsf{v}\in\bm{\mathcal{S}}_{\setminus\mathsf{s}}}\gamma^{\mathsf{s},\mathsf{v}}\psi^\mathsf{v}\Big)^\top\phi(x^\mathsf{s})\Big)\Big).
\end{align*}
Let $\bm{\uppsi} = \{\psi^\mathsf{s}\}_{\mathsf{s}\in\bm{\mathcal{S}}}$. 
Let $\mathcal{V}_\mathsf{s}$ be $1/(2\sqrt{m})$-packing of the unit $\|\cdot\|_2$-balls with cardinality at least $(2\sqrt{m})^{2B}$. We now choose a set $\mathcal{V} = \delta(\mathcal{V}_{\mathsf{s}_1}\times\mathcal{V}_{\mathsf{s}_2}\times\!...\times\mathcal{V}_{\mathsf{s}_m})$. We see that
\begin{align*}
    |\mathcal{V}|\ge (2\sqrt{m})^{2mB}.
\end{align*}

\begin{proof}
We have that 
\begin{align*}
    \|\bm{\uppsi}_1 - \bm{\uppsi}_2\|_2 = \sqrt{\sum_{\mathsf{s}\in\bm{\mathcal{S}}}\|\psi_1^\mathsf{s} - \psi_2^\mathsf{s}\|_2^2} 
\ge \delta/2.
\end{align*}
Moreover,
\begin{align*}
    &D_{\textrm{KL}}(\p^n_{\bm{\uppsi}_1}\,\|\,\p^n_{\bm{\uppsi}_2}) = \sum_{\mathsf{s}\in\bm{\mathcal{S}}}D_{\textrm{KL}}(\p^{n_\mathsf{s}}_{\bm{\uppsi}_1}\,\|\,\p^{n_\mathsf{s}}_{\bm{\uppsi}_2}).
\end{align*}
We first find upper bound of $D_{\textrm{KL}}(\p^{n_\mathsf{s}}_{\bm{\uppsi}_1}\,\|\,\p^{n_\mathsf{s}}_{\bm{\uppsi}_2})$. Since the data is independent, we have that
\begin{align*}
     D_{\textrm{KL}}&(\p^{n_\mathsf{s}}_{\bm{\uppsi}_1}\,\|\,\p^{n_\mathsf{s}}_{\bm{\uppsi}_2}) = n_\mathsf{s}D_{\textrm{KL}}(\p^1_{\bm{\uppsi}_1}\,\|\,\p^1_{\bm{\uppsi}_2})\\
     &= n_\mathsf{s}\Bigg[\varphi\Big(\Big(\psi_1^\mathsf{s} + \sum_{\mathsf{v}\in\bm{\mathcal{S}}_{\setminus\mathsf{s}}}\gamma^{\mathsf{s},\mathsf{v}}\psi_1^\mathsf{v}\Big)^\top\phi(x^{\mathsf{s}})\Big) \log \frac{\varphi\Big(\Big(\psi_1^\mathsf{s} + \sum_{\mathsf{v}\in\bm{\mathcal{S}}_{\setminus\mathsf{s}}}\gamma^{\mathsf{s},\mathsf{v}}\psi_1^\mathsf{v}\Big)^\top\phi(x^{\mathsf{s}})\Big)}{\varphi\Big(\Big(\psi_2^\mathsf{s} + \sum_{\mathsf{v}\in\bm{\mathcal{S}}_{\setminus\mathsf{s}}}\gamma^{\mathsf{s},\mathsf{v}}\psi_2^\mathsf{v}\Big)^\top\phi(x^{\mathsf{s}})\Big)} \\
     &\qquad+ \varphi\Big(-\Big(\psi_1^\mathsf{s} + \sum_{\mathsf{v}\in\bm{\mathcal{S}}_{\setminus\mathsf{s}}}\gamma^{\mathsf{s},\mathsf{v}}\psi_1^\mathsf{v}\Big)^\top\phi(x^{\mathsf{s}})\Big) \log \frac{\varphi\Big(-\Big(\psi_1^\mathsf{s} + \sum_{\mathsf{v}\in\bm{\mathcal{S}}_{\setminus\mathsf{s}}}\gamma^{\mathsf{s},\mathsf{v}}\psi_1^\mathsf{v}\Big)^\top\phi(x^{\mathsf{s}})\Big)}{\varphi\Big(-\Big(\psi_2^\mathsf{s} + \sum_{\mathsf{v}\in\bm{\mathcal{S}}_{\setminus\mathsf{s}}}\gamma^{\mathsf{s},\mathsf{v}}\psi_2^\mathsf{v}\Big)^\top\phi(x^{\mathsf{s}})\Big)}\Bigg].
\end{align*}
The first component:
\begin{align*}
    \varphi\Big(\Big(\psi_1^\mathsf{s} + &\sum_{\mathsf{v}\in\bm{\mathcal{S}}_{\setminus\mathsf{s}}}\gamma^{\mathsf{s},\mathsf{v}}\psi_1^\mathsf{v}\Big)^\top\phi(\mathbf{x}^{\mathsf{s}})\Big) \log \frac{\varphi\Big(\Big(\psi_1^\mathsf{s} + \sum_{\mathsf{v}\in\bm{\mathcal{S}}_{\setminus\mathsf{s}}}\gamma^{\mathsf{s},\mathsf{v}}\psi_1^\mathsf{v}\Big)^\top\phi(\mathbf{x}^{\mathsf{s}})\Big)}{\varphi\Big(\Big(\psi_2^\mathsf{s} + \sum_{\mathsf{v}\in\bm{\mathcal{S}}_{\setminus\mathsf{s}}}\gamma^{\mathsf{s},\mathsf{v}}\psi_2^\mathsf{v}\Big)^\top\phi(\mathbf{x}^{\mathsf{s}})\Big)}\\
&\le\left|\log \Big(1+ e^{-\big(\psi_2^\mathsf{s} + \sum_{\mathsf{v}\in\bm{\mathcal{S}}_{\setminus\mathsf{s}}}\gamma^{\mathsf{s},\mathsf{v}}\psi_2^\mathsf{v}\big)^\top\phi(\mathbf{x}^{\mathsf{s}})}\Big) - \log \Big(1+ e^{-\big(\psi_1^\mathsf{s} + \sum_{\mathsf{v}\in\bm{\mathcal{S}}_{\setminus\mathsf{s}}}\gamma^{\mathsf{s},\mathsf{v}}\psi_1^\mathsf{v}\big)^\top\phi(\mathbf{x}^{\mathsf{s}})}\Big) \right|\\
    &\overset{(\star)}{\le} \Big| \Big(\psi_2^\mathsf{s} + \sum_{\mathsf{v}\in\bm{\mathcal{S}}_{\setminus\mathsf{s}}}\gamma^{\mathsf{s},\mathsf{v}}\psi_2^\mathsf{v}\Big)^\top\phi(\mathbf{x}^{\mathsf{s}}) -  \Big(\psi_1^\mathsf{s} + \sum_{\mathsf{v}\in\bm{\mathcal{S}}_{\setminus\mathsf{s}}}\gamma^{\mathsf{s},\mathsf{v}}\psi_1^\mathsf{v}\Big)^\top\phi(\mathbf{x}^{\mathsf{s}})\Big|\\
&\le 4B\delta\Big(1 + \sum_{\mathsf{v}\in\bm{\mathcal{S}}_{\setminus\mathsf{s}}}\gamma^{\mathsf{s},\mathsf{v}}\Big),
\end{align*}
where ($\star$) follows from the fact that $x\mapsto\log(1 + e^x)$ is Lipschitz. In particular,
\begin{align*}
    \big|\log(1+e^{x_1}) - \log(1+e^{x_2})\big| = \left|\int_{x_1}^{x_2} \frac{e^x}{1+e^x} dx\right|
    \le\left|\int_{x_1}^{x_2} 1 dx\right| = \big|x_1-x_2\big|.
\end{align*}

Similarly, for the second component, we also have
\begin{align*}
    \varphi\Big(-\Big(\psi_1^\mathsf{s} + \sum_{\mathsf{v}\in\bm{\mathcal{S}}_{\setminus\mathsf{s}}}\gamma^{\mathsf{s},\mathsf{v}}\psi_1^\mathsf{v}\Big)^\top\phi(x^{\mathsf{s}})\Big) &\log \frac{\varphi\Big(-\Big(\psi_1^\mathsf{s} + \sum_{\mathsf{v}\in\bm{\mathcal{S}}_{\setminus\mathsf{s}}}\gamma^{\mathsf{s},\mathsf{v}}\psi_1^\mathsf{v}\Big)^\top\phi(x^{\mathsf{s}})\Big)}{\varphi\Big(-\Big(\psi_2^\mathsf{s} + \sum_{\mathsf{v}\in\bm{\mathcal{S}}_{\setminus\mathsf{s}}}\gamma^{\mathsf{s},\mathsf{v}}\psi_2^\mathsf{v}\Big)^\top\phi(x^{\mathsf{s}})\Big)} \\
    &\le 4B\delta\Big(1 + \sum_{\mathsf{v}\in\bm{\mathcal{S}}_{\setminus\mathsf{s}}}\gamma^{\mathsf{s},\mathsf{v}}\Big).
\end{align*}
Thus,
\begin{align*}
    D_{\textrm{KL}}(\p^{n_\mathsf{s}}_{\bm{\uppsi}_1}\,\|\,\p^{n_\mathsf{s}}_{\bm{\uppsi}_2}) \le 8B\delta\Big(1 + \sum_{\mathsf{v}\in\bm{\mathcal{S}}_{\setminus\mathsf{s}}}\gamma^{\mathsf{s},\mathsf{v}}\Big)n_\mathsf{s}.
\end{align*}
Consequently,
\begin{align*}
    D_{\textrm{KL}}(\p^n_{\bm{\uppsi}_1}\,\|\,\p^n_{\bm{\uppsi}_2}) \le 8B\delta\sum_{\mathsf{s}\in\bm{\mathcal{S}}}n_\mathsf{s}\Big(1 + \sum_{\mathsf{v}\in\bm{\mathcal{S}}_{\setminus\mathsf{s}}}\gamma^{\mathsf{s},\mathsf{v}}\Big).
\end{align*}
So, we have that
\begin{align*}
    \inf_{\hat{\bm{\uppsi}}_n} \sup_{P\in\mathcal{P}} \mathbb{E}_P \left[ \|\hat{\bm{\uppsi}}_n-\bm{\uppsi}(P)\|_2 \right] &\ge \frac{\delta}{4}\left(1- \frac{8B\delta\sum_{\mathsf{s}\in\bm{\mathcal{S}}}n_\mathsf{s}\Big(1 + \sum_{\mathsf{v}\in\bm{\mathcal{S}}_{\setminus\mathsf{s}}}\gamma^{\mathsf{s},\mathsf{v}}\Big) + \log 2}{\log|\mathcal{V}|}\right)\\
    &\ge \frac{\delta}{4}\left(1- \frac{8B\delta\sum_{\mathsf{s}\in\bm{\mathcal{S}}}n_\mathsf{s}\Big(1 + \sum_{\mathsf{v}\in\bm{\mathcal{S}}_{\setminus\mathsf{s}}}\gamma^{\mathsf{s},\mathsf{v}}\Big) + \log 2}{2mB\log(2\sqrt{m})}\right).
\end{align*}
We choose $\delta=\frac{m\log(2\sqrt{m})}{16\sum_{\mathsf{s}\in\bm{\mathcal{S}}}n_\mathsf{s}\big(1 + \sum_{\mathsf{v}\in\bm{\mathcal{S}}_{\setminus\mathsf{s}}}\gamma^{\mathsf{s},\mathsf{v}}\big)}$, then
\begin{align*}
   1- \frac{8B\delta\sum_{\mathsf{s}\in\bm{\mathcal{S}}}n_\mathsf{s}\Big(1 + \sum_{\mathsf{v}\in\bm{\mathcal{S}}_{\setminus\mathsf{s}}}\gamma^{\mathsf{s},\mathsf{v}}\Big) + \log 2}{2mB\log(2\sqrt{m})} 
\ge \frac{1}{4}.
\end{align*}
Thus,
\begin{align*}
    \inf_{\hat{\bm{\uppsi}}_n} \sup_{P\in\mathcal{P}} \mathbb{E}_P \left[ \|\hat{\bm{\uppsi}}_n-\bm{\uppsi}(P)\|_2 \right] &\ge \frac{1}{4} \times \frac{mB\log(2\sqrt{m})}{16B\sum_{\mathsf{s}\in\bm{\mathcal{S}}}n_\mathsf{s}\big(1 + \sum_{\mathsf{v}\in\bm{\mathcal{S}}_{\setminus\mathsf{s}}}\gamma^{\mathsf{s},\mathsf{v}}\big)} \times \frac{1}{4}\\
    &= \frac{m\log(2\sqrt{m})}{256\sum_{\mathsf{s}\in\bm{\mathcal{S}}}n_\mathsf{s}\big(1 + \sum_{\mathsf{v}\in\bm{\mathcal{S}}_{\setminus\mathsf{s}}}\gamma^{\mathsf{s},\mathsf{v}}\big)}.
\end{align*}
This completes the proof of part (i).
\end{proof}

\subsection{Proof of Part (ii)}
\label{sec:proof-lem1-part-ii}
\begin{proof}
We summarize the model as follows
\begin{align*}
    y^\mathsf{s} &= \Big((1-w^\mathsf{s})\big(\beta_0^\mathsf{s} + \sum_{\mathsf{v}\in\bm{\mathcal{S}}_{\setminus\mathsf{s}}}\eta^{\mathsf{s},\mathsf{v}}\beta_0^\mathsf{v}\big) + w^\mathsf{s}\big(\beta_1^\mathsf{s} + \sum_{\mathsf{v}\in\bm{\mathcal{S}}_{\setminus\mathsf{s}}}\eta^{\mathsf{s},\mathsf{v}}\beta_1^\mathsf{v}\big)\Big)^\top\phi(\mathbf{x}^\mathsf{s})  + \epsilon_\mathsf{s}, \qquad \epsilon_\mathsf{s}\sim\mathsf{N}(0,\sigma^2).
\end{align*}
Let $\bm{\upbeta} = \{\beta_0^\mathsf{s}, \beta_1^\mathsf{s}\}_{\mathsf{s}\in\bm{\mathcal{S}}}$. 
Let $\mathcal{V}_{0\mathsf{s}}$ and $\mathcal{V}_{1\mathsf{s}}$ be $1/(2\sqrt{m})$-packing of the unit $\|\cdot\|_2$-balls with cardinality at least $(2\sqrt{m})^{2B}$. Let $\mathcal{V}_{\mathsf{s}} = \mathcal{V}_{0\mathsf{s}}\times\mathcal{V}_{1\mathsf{s}}$. We now choose a set $\mathcal{V} = \delta(\mathcal{V}_{\mathsf{s}_1}\times \mathcal{V}_{\mathsf{s}_2}\times\!...\times\mathcal{V}_{\mathsf{s}_m})$. We see that
\begin{align*}
    |\mathcal{V}|\ge (2\sqrt{m})^{4mB}.
\end{align*}
We have that 
\begin{align*}
    \|\bm{\upbeta}_1 - \bm{\upbeta}_2\|_2 &= \sqrt{\sum_{\mathsf{s}\in\bm{\mathcal{S}}}\Big(\|(\beta_0^\mathsf{s})_1 - (\beta_0^\mathsf{s})_2\|_2^2 + \|(\beta_1^\mathsf{s})_1 - (\beta_1^\mathsf{s})_2\|_2^2\Big)}
\ge \delta/\sqrt{2}.
\end{align*}
Moreover,
\begin{align*}
    D_{\textrm{KL}}(p_{\bm{\upbeta}_1}^n\,\|\,p_{\bm{\upbeta}_2}^n) = \sum_{\mathsf{s}\in\bm{\mathcal{S}}}D_{\textrm{KL}}(p_{\bm{\upbeta}_1}^{n_\mathsf{s}}\,\|\,p_{\bm{\upbeta}_2}^{n_\mathsf{s}}) = \sum_{\mathsf{s}\in\bm{\mathcal{S}}}n_\mathsf{s}D_{\textrm{KL}}(p_{\bm{\upbeta}_1}^1\,\|\,p_{\bm{\upbeta}_2}^1).
\end{align*}
In addition,
\begin{align*}
    D_{\textrm{KL}}&(p_{\bm{\upbeta}_1}^1\,\|\,p_{\bm{\upbeta}_2}^1) \\
    &= \frac{1}{2\sigma^2}\Bigg(\Big((1-w^\mathsf{s})\big((\beta_0^\mathsf{s})_1 + \sum_{\mathsf{v}\in\bm{\mathcal{S}}_{\setminus\mathsf{s}}}\eta^{\mathsf{s},\mathsf{v}}(\beta_0^\mathsf{v})_1\big) + w^\mathsf{s}\big((\beta_1^\mathsf{s})_1 + \sum_{\mathsf{v}\in\bm{\mathcal{S}}_{\setminus\mathsf{s}}}\eta^{\mathsf{s},\mathsf{v}}(\beta_1^\mathsf{v})_1\big)\Big)^\top\phi(\mathbf{x}^\mathsf{s}) \\
    &\qquad\quad - \Big((1-w^\mathsf{s})\big((\beta_0^\mathsf{s})_2 + \sum_{\mathsf{v}\in\bm{\mathcal{S}}_{\setminus\mathsf{s}}}\eta^{\mathsf{s},\mathsf{v}}(\beta_0^\mathsf{v})_2\big) + w^\mathsf{s}\big((\beta_1^\mathsf{s})_2 + \sum_{\mathsf{v}\in\bm{\mathcal{S}}_{\setminus\mathsf{s}}}\eta^{\mathsf{s},\mathsf{v}}(\beta_1^\mathsf{v})_2\big)\Big)^\top\phi(\mathbf{x}^\mathsf{s})\Bigg)^2\\
&\le \frac{1}{2\sigma^2}\Bigg(\Big((1-w^\mathsf{s})\big(2\delta + \sum_{\mathsf{v}\in\bm{\mathcal{S}}_{\setminus\mathsf{s}}}\eta^{\mathsf{s},\mathsf{v}}2\delta\big) + w^\mathsf{s}\big(2\delta + \sum_{\mathsf{v}\in\bm{\mathcal{S}}_{\setminus\mathsf{s}}}\eta^{\mathsf{s},\mathsf{v}}2\delta\big)\Big)\|\phi(\mathbf{x}^\mathsf{s})\|_2\Bigg)^2 \\
&\le \frac{8B^2\delta^2}{\sigma^2}\Big(1 + \sum_{\mathsf{v}\in\bm{\mathcal{S}}_{\setminus\mathsf{s}}}\eta^{\mathsf{s},\mathsf{v}}\Big)^2,
\end{align*}
Thus,
\begin{align*}
    D_{\textrm{KL}}(p_{\bm{\upbeta}_1}^n\,\|\,p_{\bm{\upbeta}_2}^n) \le \frac{8B^2\delta^2}{\sigma^2}\sum_{\mathsf{s}\in\bm{\mathcal{S}}}n_\mathsf{s}\Big(1 + \sum_{\mathsf{v}\in\bm{\mathcal{S}}_{\setminus\mathsf{s}}}\eta^{\mathsf{s},\mathsf{v}}\Big)^2.
\end{align*}
Consequently,
\begin{align*}
    \inf_{\hat{\bm{\upbeta}}_n} \sup_{P\in\mathcal{P}} \mathbb{E}_P \left[ \|\hat{\bm{\upbeta}}_n-\bm{\upbeta}(P)\|_2 \right] &\ge \frac{\delta}{2\sqrt{2}}\left(1- \frac{\frac{8B^2\delta^2}{\sigma^2}\sum_{\mathsf{s}\in\bm{\mathcal{S}}}n_\mathsf{s}\Big(1 + \sum_{\mathsf{v}\in\bm{\mathcal{S}}_{\setminus\mathsf{s}}}\eta^{\mathsf{s},\mathsf{v}}\Big)^2 + \log 2}{\log|\mathcal{V}|}\right)\\
    &\ge \frac{\delta}{2\sqrt{2}}\left(1- \frac{\frac{8B^2\delta^2}{\sigma^2}\sum_{\mathsf{s}\in\bm{\mathcal{S}}}n_\mathsf{s}\Big(1 + \sum_{\mathsf{v}\in\bm{\mathcal{S}}_{\setminus\mathsf{s}}}\eta^{\mathsf{s},\mathsf{v}}\Big)^2 + \log 2}{4mB\log(2\sqrt{m})}\right).
\end{align*}
We choose $\delta^2=\frac{mB\log(2\sqrt{m})}{4\frac{B^2}{\sigma^2}\sum_{\mathsf{s}\in\bm{\mathcal{S}}}n_\mathsf{s}\Big(1 + \sum_{\mathsf{v}\in\bm{\mathcal{S}}_{\setminus\mathsf{s}}}\eta^{\mathsf{s},\mathsf{v}}\Big)^2}$, then
\begin{align*}
   1- \frac{\frac{8B^2\delta^2}{\sigma^2}\sum_{\mathsf{s}\in\bm{\mathcal{S}}}n_\mathsf{s}\Big(1 + \sum_{\mathsf{v}\in\bm{\mathcal{S}}_{\setminus\mathsf{s}}}\eta^{\mathsf{s},\mathsf{v}}\Big)^2 + \log 2}{4mB\log(2\sqrt{m})} = 1- \frac{2mB\log(2\sqrt{m})+ \log 2}{4mB\log(2\sqrt{m})} \ge \frac{1}{4}.
\end{align*}
Thus,
\begin{align*}
    \inf_{\hat{\bm{\upbeta}}_n} \sup_{P\in\mathcal{P}} \mathbb{E}_P \left[ \|\hat{\bm{\upbeta}}_n-\bm{\upbeta}(P)\|_2 \right] &\ge \frac{1}{2\sqrt{2}}\sqrt{\frac{4mB\log(2\sqrt{m})}{2\frac{8B^2}{\sigma^2}\sum_{\mathsf{s}\in\bm{\mathcal{S}}}n_\mathsf{s}\Big(1 + \sum_{\mathsf{v}\in\bm{\mathcal{S}}_{\setminus\mathsf{s}}}\eta^{\mathsf{s},\mathsf{v}}\Big)^2}} \times \frac{1}{4}\\
    &= \frac{\sigma}{16\sqrt{2}}\sqrt{\frac{m\log(2\sqrt{m})}{B\sum_{\mathsf{s}\in\bm{\mathcal{S}}}n_\mathsf{s}\Big(1 + \sum_{\mathsf{v}\in\bm{\mathcal{S}}_{\setminus\mathsf{s}}}\eta^{\mathsf{s},\mathsf{v}}\Big)^2}}.
\end{align*}
This completes the proof of part (ii).
\end{proof}

\section{Further cases of the minimax lower bounds}

In Lemma~\ref{lem:minimax-bounds1} and \ref{lem:mimimax-bounds2}, we have presented the minimax lower bounds when $y_i^\mathsf{s}\in\mathbb{R}$ and $x_i^\mathsf{s}\in\mathbb{R}^{d_x}$. Here, we briefly describe the other cases.

\subsection{Further cases of Lemma~\ref{lem:minimax-bounds1}}

In this section, we further detail the lower bound for binary outcomes and binary proxy variables. In this case, we need to re-derive the upper bound of 
\begin{align*} \p_{\bm{\uptheta}_1}(w=j|z) D_{\textnormal{KL}}\big[\p_{\bm{\uptheta}_1}(y|w=j,z) \big\|\p_{\bm{\uptheta}_2}(y|w=j,z')\big] \quad \text{ and }\quad D_{\textnormal{KL}}\big[\p_{\bm{\uptheta}_1}(x|z)\big\|\p_{\bm{\uptheta}_2}(x|z')\big],
\end{align*}
where $j = 1,2$. Using similar derivations as before for the quantity $D_{\textnormal{KL}}\big[\p_{\bm{\uptheta}_1}(w|z)\big\|\p_{\bm{\uptheta}_2}(w|z')\big]$, we have that
\begin{align*}
    \p_{\bm{\uptheta}_1}(w=j|z) D_{\textnormal{KL}}\big[\p_{\bm{\uptheta}_1}(y|w=j,z) \big\|\p_{\bm{\uptheta}_2}(y|w=j,z')\big] \le 8B\delta\Big(1 + \sum_{\mathsf{v}\in\bm{\mathcal{S}}_{\setminus\mathsf{s}}}\lambda^{\mathsf{s},\mathsf{v}}\Big),
\end{align*}
and 
\begin{align*}
    D_{\textnormal{KL}}\big[\p_{\bm{\uptheta}_1}(x|z)\big\|\p_{\bm{\uptheta}_2}(x|z')\big] \le d_x 8B\delta\Big(1 + \sum_{\mathsf{v}\in\bm{\mathcal{S}}_{\setminus\mathsf{s}}}\lambda^{\mathsf{s},\mathsf{v}}\Big).
\end{align*}
Combining the results, we have
\begin{align*}
    D_{\textrm{KL}}(\p^n_{\bm{\uptheta}_1}\,\|\,\p^n_{\bm{\uptheta}_2}) &= \sum_{\mathsf{s}\in\bm{\mathcal{S}}}D_{\textrm{KL}}(\p^{n_\mathsf{s}}_{\bm{\uptheta}_1}\,\|\,\p^{n_\mathsf{s}}_{\bm{\uptheta}_2})\le \sum_{\mathsf{s}\in\bm{\mathcal{S}}}n_\mathsf{s} 8(d_x+3)B\delta\Big(1 + \sum_{\mathsf{v}\in\bm{\mathcal{S}}_{\setminus\mathsf{s}}}\lambda^{\mathsf{s},\mathsf{v}}\Big).
\end{align*}
Consequently, we have that
\begin{align*}
    &\inf_{\hat{\bm{\uptheta}}_n} \sup_{P\in\mathcal{P}} \mathbb{E}_P \left[ \|\hat{\bm{\uptheta}}_n\!-\!\bm{\uptheta}(P)\|_2 \right] \ge \frac{\delta}{2}\!\left(\!1\!-\! \frac{\sum_{\mathsf{s}\in\bm{\mathcal{S}}}n_\mathsf{s} 8(d_x+3)B\delta\Big(1 + \sum_{\mathsf{v}\in\bm{\mathcal{S}}_{\setminus\mathsf{s}}}\lambda^{\mathsf{s},\mathsf{v}}\Big) \!+\! \log 2}{2mB(d_x+3)\log(2\sqrt{m})}\!\right)\!\!.
\end{align*}
We choose $\delta=\frac{m\log(2\sqrt{m})}{8\sum_{\mathsf{s}\in\bm{\mathcal{S}}}n_\mathsf{s}\Big(1 + \sum_{\mathsf{v}\in\bm{\mathcal{S}}_{\setminus\mathsf{s}}}\lambda^{\mathsf{s},\mathsf{v}}\Big)}$, then
\begin{align*}
&1- \frac{\sum_{\mathsf{s}\in\bm{\mathcal{S}}}n_\mathsf{s} 8(d_x+3)B\delta\Big(1 + \sum_{\mathsf{v}\in\bm{\mathcal{S}}_{\setminus\mathsf{s}}}\lambda^{\mathsf{s},\mathsf{v}}\Big) \!+\! \log 2}{2mB(d_x+3)\log(2\sqrt{m})} \ge \frac{3}{8}.
\end{align*}
Thus,
\begin{align*}
    \Aboxed{\inf_{\hat{\bm{\uptheta}}_n} \sup_{P\in\mathcal{P}} \mathbb{E}_P \left[ \|\hat{\bm{\uptheta}}_n\!-\!\bm{\uptheta}(P)\|_2 \right] \ge \frac{3mB\log(2\sqrt{m})}{128\sum_{\mathsf{s}\in\bm{\mathcal{S}}}n_\mathsf{s}B\Big(1 + \sum_{\mathsf{v}\in\bm{\mathcal{S}}_{\setminus\mathsf{s}}}\lambda^{\mathsf{s},\mathsf{v}}\Big)}.}
\end{align*}

\textbf{Remark.} Note that the derivation in this Section and in Section~\emph{\ref{sec:proof-lem1-part-i}} give us enough tools to compute the minimax lower bounds for any further case, i.e., any combination of the outcomes and proxy variables (binary or continuous). The key is to initially find the upper bound of $D_{\textrm{KL}}(\p^n_{\bm{\uptheta}_1}\,\|\,\p^n_{\bm{\uptheta}_2})$ based on the constructed packing. Then, using Fano's method to obtain the minimax lower bounds.
\subsection{Further cases of Lemma~\ref{lem:mimimax-bounds2}}
Note that the lower bound of Lemma~\ref{lem:mimimax-bounds2}, part~\textbf{(i)} has only one case since we only focus on binary treatment, and it is presented in the main text. For part~\textbf{(ii)}, consider $y_i^\mathsf{s}\in \{0,1\}$, then the model of the outcomes would follow a Bernoulli distribution. Reusing the scheme in Section~\ref{sec:proof-lem1-part-ii}, we need to find the new upper bound of $D_{\textrm{KL}}(p_{\bm{\upbeta}_1}^n\,\|\,p_{\bm{\upbeta}_2}^n)$. In particular,
\begin{align*}
    D_{\textrm{KL}}(p_{\bm{\upbeta}_1}^n\,\|\,p_{\bm{\upbeta}_2}^n) &= \sum_{\mathsf{s}\in\bm{\mathcal{S}}}n_\mathsf{s}\bigg[\varphi(v_1) \log \frac{\varphi(v_1)}{\varphi(v_2)} + \varphi(-v_1) \log \frac{\varphi(-v_1}{\varphi(-v_2)}\bigg],
\end{align*}
where $ v_j = \Big((1-w^\mathsf{s})\big((\beta_0^\mathsf{s})_j + \sum_{\mathsf{v}\in\bm{\mathcal{S}}_{\setminus\mathsf{s}}}\eta^{\mathsf{s},\mathsf{v}}(\beta_0^\mathsf{v})_j\big) + w^\mathsf{s}\big((\beta_1^\mathsf{s})_j + \sum_{\mathsf{v}\in\bm{\mathcal{S}}_{\setminus\mathsf{s}}}\eta^{\mathsf{s},\mathsf{v}}(\beta_1^\mathsf{v})_j\big)\Big)^\top\phi(\mathbf{x}^\mathsf{s})$.
We have that
\begin{align*}
    \varphi(v_1) \log \frac{\varphi(v_1)}{\varphi(v_2)}
&\le \bigg\|(1-w^\mathsf{s})\big((\beta_0^\mathsf{s})_1 - (\beta_0^\mathsf{s})_2 + \sum_{\mathsf{v}\in\bm{\mathcal{S}}_{\setminus\mathsf{s}}}\eta^{\mathsf{s},\mathsf{v}}[(\beta_0^\mathsf{v})_1 - (\beta_0^\mathsf{v})_2]\big) \\
    &\qquad\qquad+ w^\mathsf{s}\big((\beta_1^\mathsf{s})_1 - (\beta_1^\mathsf{s})_2 + \sum_{\mathsf{v}\in\bm{\mathcal{S}}_{\setminus\mathsf{s}}}\eta^{\mathsf{s},\mathsf{v}}[(\beta_1^\mathsf{v})_1 - (\beta_1^\mathsf{v})_2]\big)\bigg\|_2\|\phi(\mathbf{x}^\mathsf{s})\|_2 \\
&\le 4B\delta\Big(1 + \sum_{\mathsf{v}\in\bm{\mathcal{S}}_{\setminus\mathsf{s}}}\gamma^{\mathsf{s},\mathsf{v}}\Big),
\end{align*}
Similarly, $\varphi(-v_1) \log \frac{\varphi(-v_1}{\varphi(-v_2)} \le 4B\delta\Big(1 + \sum_{\mathsf{v}\in\bm{\mathcal{S}}_{\setminus\mathsf{s}}}\gamma^{\mathsf{s},\mathsf{v}}\Big)$.
Hence,
\begin{align*}
    D_{\textrm{KL}}(p_{\bm{\upbeta}_1}^n\,\|\,p_{\bm{\upbeta}_2}^n) \le 8B\delta\sum_{\mathsf{s}\in\bm{\mathcal{S}}}n_\mathsf{s}\Big(1 + \sum_{\mathsf{v}\in\bm{\mathcal{S}}_{\setminus\mathsf{s}}}\eta^{\mathsf{s},\mathsf{v}}\Big).
\end{align*}
Using similar technique in Section~\ref{sec:proof-lem1-part-ii}, we obtain
\begin{align*}
    \Aboxed{\inf_{\hat{\bm{\upbeta}}_n} \sup_{P\in\mathcal{P}} \mathbb{E}_P \left[ \|\hat{\bm{\upbeta}}_n-\bm{\upbeta}(P)\|_2 \right] &\ge \frac{m\log(2\sqrt{m})}{32\sqrt{2}\sum_{\mathsf{s}\in\bm{\mathcal{S}}}n_\mathsf{s}\Big(1 + \sum_{\mathsf{v}\in\bm{\mathcal{S}}_{\setminus\mathsf{s}}}\eta^{\mathsf{s},\mathsf{v}}\Big)}.}
\end{align*}
We observe that the lower bound is similar to that of Lemma~\ref{lem:mimimax-bounds2}, part~\textbf{(i)} since they are both lower bounds of a binary response variable. The constant in this bound is larger ($1/(32\sqrt{2})$) than that of Lemma~\ref{lem:mimimax-bounds2}, part~\textbf{(i)} ($1/256$). This is expected since there are more parameters in this model, i.e., $\{\beta_0^\mathsf{s},\beta_1^\mathsf{s}\}_{\mathsf{s}\in\bm{\mathcal{S}}}$, as compared to the model in Lemma~\ref{lem:mimimax-bounds2}, part~\textbf{(i)} ($\{\psi^\mathsf{s}\}_{\mathsf{s}\in\bm{\mathcal{S}}}$).

\section{Derivation of \texorpdfstring{$\psi_w$}{psi-w} and \texorpdfstring{$\psi_w(x,u_r)$}{psi-w(x,ur}}
\label{ap:causalfi-psi}
\textbf{For $\psi_w(x,u_r)$:} We have that 
\begin{align*}
    \psi_w(x,u_r)&=E[Y(w)|X=x,U_r=u_r].
\end{align*}
By the law of total expectation, we have
\begin{align*}
\psi_w(x,u_r)&=E_{U_{\tilde{r}}}\big[E[Y(w)|X=x,U_r=u_r,U_{\tilde{r}}]\big].
\end{align*}
Under the strong ignorability assumption (Assumption~1 in the main text), we have that $Y(w)\ind W|X,U_r,U_{\tilde{r}}$. Hence, we can rewrite 
\begin{align*}
\psi_w(x,u_r)&=E_{U_{\tilde{r}}}\big[E[Y(w)|X=x,U_r=u_r,U_{\tilde{r}}, W=w]\big]\\
&=E_{U_{\tilde{r}}}\big[E[Y|W=w,X=x,U_r=u_r,U_{\tilde{r}}]\big],
\end{align*}
where the last equality follows from the consistency assumption.

\noindent\textbf{For $\psi_w$:} We have that $\psi_w = E[Y(w)]$.

\noindent By the law of total expectation, we have
\begin{align*}
\psi_w &=E_{X,U_r,U_{\tilde{r}}}\big[E[Y(w)|X,U_r,U_{\tilde{r}}]\big]\\
&=E_{X,U_r}\big[E[Y(w)|X,U_r]\big]\\
&= E_{X,U_r}\big[\psi_w(X,U_r)\big].
\end{align*}
\section{The Loss Functions and Model Structures}
\label{ap:causalfi-loss}
This section present details of the loss functions and model structures to parameterize the distributions.
\subsection{The Loss Functions}
The loss functions for learning $p(u|y,w,x)$, $\hat{p}(u_{\tilde{r}}|x,u_r)$, and $\hat{p}(y|w,x,u)$ have similar form. 

\noindent For $p(u|y,w,x)$, we have
\begin{align*}
    \log p(\mathbf{u}_r|\mathbf{y}, \mathbf{w}, \mathbf{x}) &\ge  \int \q_\phi(\theta)\log\frac{p(\mathbf{u}_r,\theta|\mathbf{y}, \mathbf{w}, \mathbf{x})}{\q_\phi(\theta)} d\theta\\
    &= E_{\theta}\big[\log p(\mathbf{u}_r|\mathbf{y},\mathbf{w},\mathbf{x};\theta)\big] - D_{\text{KL}}[q_\phi(\theta)\|p(\theta)]\\
    &= \sum_{\mathsf{s}=1}^m\!\Big(E_{\theta}\big[\log p(\mathbf{u}_r^\mathsf{s}|\mathbf{y}^\mathsf{s},\mathbf{w}^\mathsf{s},\mathbf{x}^\mathsf{s};\theta)\big] \!-\! \frac{1}{m}D_{\text{KL}}[q_\phi(\theta)\|p(\theta)]\Big) =\vcentcolon\sum_{\mathsf{s}=1}^m L^{\mathsf{s}}.
\end{align*}
We learn $p(u|y,w,x)$ by maximizing the above evidence lower bound (ELBO) in a federated setting. We then use it to generate $\mathbf{u}$ to learn $\hat{p}(u_{\tilde{r}}|x,u_r)$ and $\hat{p}(y|w,x,u)$.

\noindent The loss function for $\hat{p}(u_{\tilde{r}}|x,u_r)$ is similar. We have that
\begin{align*}
    \log \hat{p}(\mathbf{u}_{\tilde{r}}|\mathbf{x}, \mathbf{u}_r) &\ge \sum_{\mathsf{s}=1}^m\!\Big(E_{\theta_u}\big[\log p(\mathbf{u}_{\tilde{r}}^\mathsf{s}|\mathbf{x}^\mathsf{s}, \mathbf{u}_r^\mathsf{s};\theta_u)\big] -\frac{1}{m}D_{\text{KL}}[q_{\hat{\phi}}(\theta_u)\|p(\theta_u)]\Big) =\vcentcolon\sum_{\mathsf{s}=1}^m L_u^{\mathsf{s}}.
\end{align*}
For $\hat{p}(y|w,x,u)$, we have
\begin{align*}
    \log \hat{p}(\mathbf{y}|\mathbf{w}, \mathbf{x},\mathbf{u}) &\ge \sum_{\mathsf{s}=1}^m\!\Big(E_{\theta_u}\big[\log p(\mathbf{y}^\mathsf{s}|\mathbf{w}^\mathsf{s}, \mathbf{x}^\mathsf{s},\mathbf{u}^\mathsf{s};\theta_y)\big] - \frac{1}{m}D_{\text{KL}}[q_{\hat{\phi}}(\theta_y)\|p(\theta_y)]\Big)\\
    &=\vcentcolon\sum_{\mathsf{s}=1}^m L_y^{\mathsf{s}}.
\end{align*}

\begin{figure}
    \centering
    \includegraphics[width=0.45\textwidth]{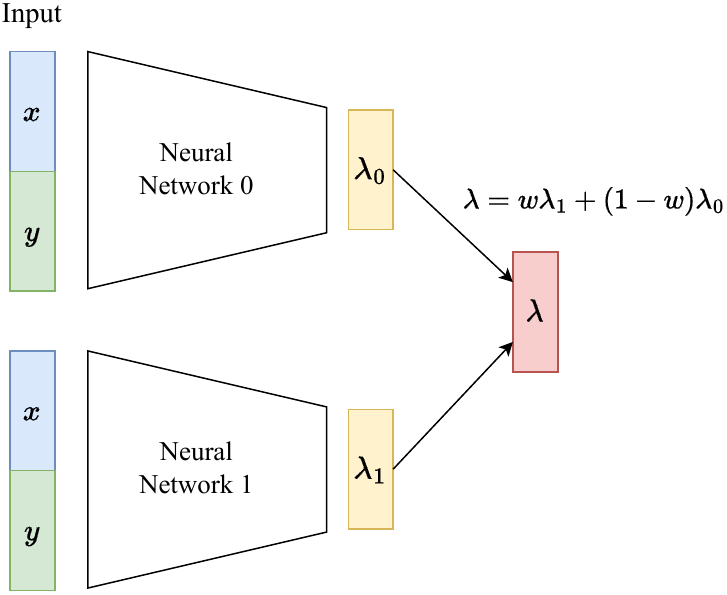}
    \caption{The model structure that  parameterizes for  $p(\mathbf{u}_r^\mathsf{s}|\mathbf{y}^\mathsf{s},\mathbf{w}^\mathsf{s},\mathbf{x}^\mathsf{s};\theta)$.}
    \label{fig:neural-nets-1}
\end{figure}
\begin{figure}
    \centering
    \includegraphics[width=0.45\textwidth]{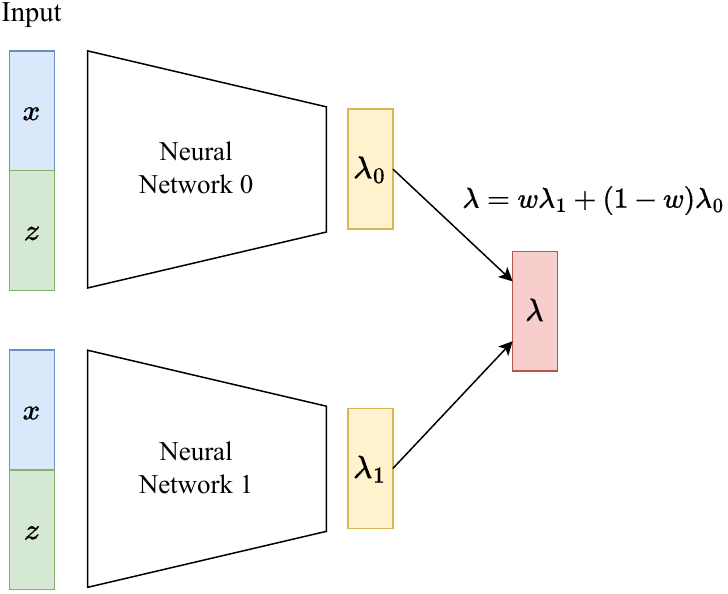}
    \caption{The model structure that  parameterizes for  $p(\mathbf{y}^\mathsf{s}|\mathbf{w}^\mathsf{s}, \mathbf{x}^\mathsf{s},\mathbf{u}^\mathsf{s};\theta_y)$.}
    \label{fig:neural-nets-2}
\end{figure}
\begin{figure}
    \centering
    \includegraphics[width=0.27\textwidth]{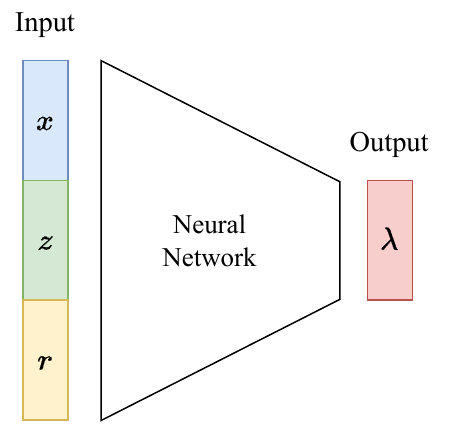}
    \caption{The model structure that  parameterizes for  $p(\mathbf{u}_{\tilde{r}}^\mathsf{s}|\mathbf{x}^\mathsf{s}, \mathbf{u}_r^\mathsf{s};\theta_u)$.}
    \label{fig:neural-nets}
\end{figure}

\subsection{Model Structures}
\textbf{For $p(\mathbf{u}_r^\mathsf{s}|\mathbf{y}^\mathsf{s},\mathbf{w}^\mathsf{s},\mathbf{x}^\mathsf{s};\theta)$:} The input to this model are $x,y,w$. As explained in the main text, we model this distribution with 2 neural networks, each associated with the case of $w=1$ or $w=0$. We illustrate the model structure in Figure~\ref{fig:neural-nets-1}. Herein, $\lambda$ is an output vector to model parameters of $p(\mathbf{u}_r^\mathsf{s}|\mathbf{y}^\mathsf{s},\mathbf{w}^\mathsf{s},\mathbf{x}^\mathsf{s};\theta)$, depending on which distributions are used (Gaussian, Bernoulli, Categorical, Poisson, etc.) to model $u$.

\noindent\textbf{For $p(\mathbf{y}^\mathsf{s}|\mathbf{w}^\mathsf{s}, \mathbf{x}^\mathsf{s},\mathbf{u}^\mathsf{s};\theta_y)$:} The input to this model are $x,u,w$. Similarly, we also model this distribution with 2 neural networks, each associated with the case of $w=1$ or $w=0$. We illustrate the model structure in Figure~\ref{fig:neural-nets-2}. Similarly, $\lambda$ is an output vector to model parameters of $p(\mathbf{u}_r^\mathsf{s}|\mathbf{y}^\mathsf{s},\mathbf{w}^\mathsf{s},\mathbf{x}^\mathsf{s};\theta)$, depending on which distribution is used for the outcome $y$ (Gaussian, Bernoulli, Categorical, Poisson, etc.).

\noindent \textbf{For $p(\mathbf{u}_{\tilde{r}}^\mathsf{s}|\mathbf{x}^\mathsf{s}, \mathbf{z}_r^\mathsf{s};\theta_u)$:}  since the inputs are $x$ and $u_r$, and the number of dimensions of $u_r$ is variable, we need to indicate this in our model. Hence, we construct a neural network with the input is a concatenation of $x$, $u$, and $r$, where the missing values is replaced by 0 and we used $r$ to indicate where the missing values are.

We illustrate the structure in Figure~\ref{fig:neural-nets}. Herein, $\lambda$ is an output vector to model parameters of $p(\mathbf{u}_{\tilde{r}}^\mathsf{s}|\mathbf{x}^\mathsf{s}, \mathbf{u}_r^\mathsf{s};\theta_u)$. The number of dimensions of $\lambda$ equals to that of the incomplete confounder $z$ and it is fixed. We take values in $\lambda$ at which the missing indicators are 0s (missing locations in $u$)  to model parameters for $p(\mathbf{u}_{\tilde{r}}^\mathsf{s}|\mathbf{x}^\mathsf{s}, \mathbf{z}_r^\mathsf{s};\theta_u)$. In our experiments, we used a fully connected neural network with 3 hidden layers, each of 20 hidden nodes.

\vskip 0.2in
\bibliography{ref}

\end{document}